\documentclass[10pt,twocolumn,letterpaper]{article}

\usepackage{cvpr}              

\definecolor{cvprblue}{rgb}{0.21,0.49,0.74}
\usepackage[pagebackref,breaklinks,colorlinks,allcolors=cvprblue]{hyperref}
\usepackage{pifont}
\newcommand{\xmark}{\ding{55}}
\newcommand{\cmark}{\ding{51}}
\usepackage{multirow}
\usepackage{caption}
\usepackage{subcaption}

\def\paperID{4469} 
\def\confName{CVPR}
\def\confYear{2025}

\title{End-to-End HOI Reconstruction Transformer with Graph-based Encoding}

\author{
    Zhenrong Wang$^{1}$\hskip1.0em Qi Zheng$^{1}$\footnotemark[2]\hskip1.0em Sihan Ma$^{2}$\hskip1.0emMaosheng Ye$^{3}$\hskip1.0em Yibing Zhan$^{4}$\hskip1.0em Dongjiang Li$^{4}$ \\
   $^{1}$Shenzhen University\hskip1.0em $^{2}$University of Sydney\hskip1.0em
   $^{3}$DeepRoute.AI\hskip1.0em
   $^{4}$JD Explore Academy \\
   {\tt\small 2022280175@email.szu.edu.cn, qiz@szu.edu.cn} \\
   \textcolor{red}{\href{https://hoi-tg.github.io/}{\textcolor{red}{\tt\small https://hoi-tg.github.io/}}}
}

\begin{document}
\maketitle
\renewcommand{\thefootnote}{\fnsymbol{footnote}}
\footnotetext[2]{Corresponding author.}
\begin{abstract}

With the diversification of human-object interaction (HOI) applications and the success of capturing human meshes, HOI reconstruction has gained widespread attention. Existing mainstream HOI reconstruction methods often rely on explicitly modeling interactions between humans and objects. However, such a way leads to a natural conflict between 3D mesh reconstruction, which emphasizes global structure, and fine-grained contact reconstruction, which focuses on local details. To address the limitations of explicit modeling, we propose the End-to-End HOI Reconstruction Transformer with Graph-based Encoding (HOI-TG). It implicitly learns the interaction between humans and objects by leveraging self-attention mechanisms. Within the transformer architecture, we devise graph residual blocks to aggregate the topology among vertices of different spatial structures. This dual focus effectively balances global and local representations. Without bells and whistles, HOI-TG achieves state-of-the-art performance on BEHAVE and InterCap datasets. Particularly on the challenging InterCap dataset, our method improves the reconstruction results for human and object meshes by 8.9\% and 8.6\%, respectively.

\end{abstract}

\section{Introduction}
\label{sec:intro}

\begin{figure}[htbp]
    \centering
    \begin{subfigure}{0.45\textwidth}
        \includegraphics[width=\linewidth]{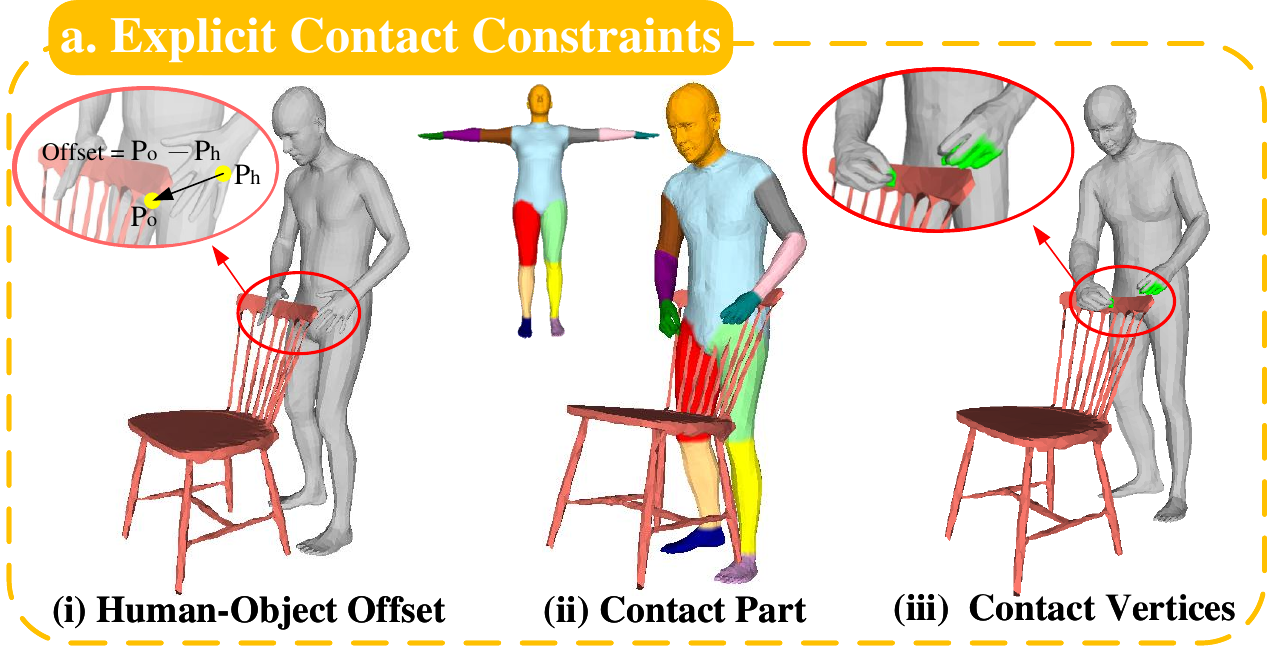}
        \phantomsubcaption
        \label{fig:motivation1}
    \end{subfigure}
    \vfill
    \vspace{-0.3cm}
    \begin{subfigure}{0.45\textwidth}
        \includegraphics[width=\linewidth]{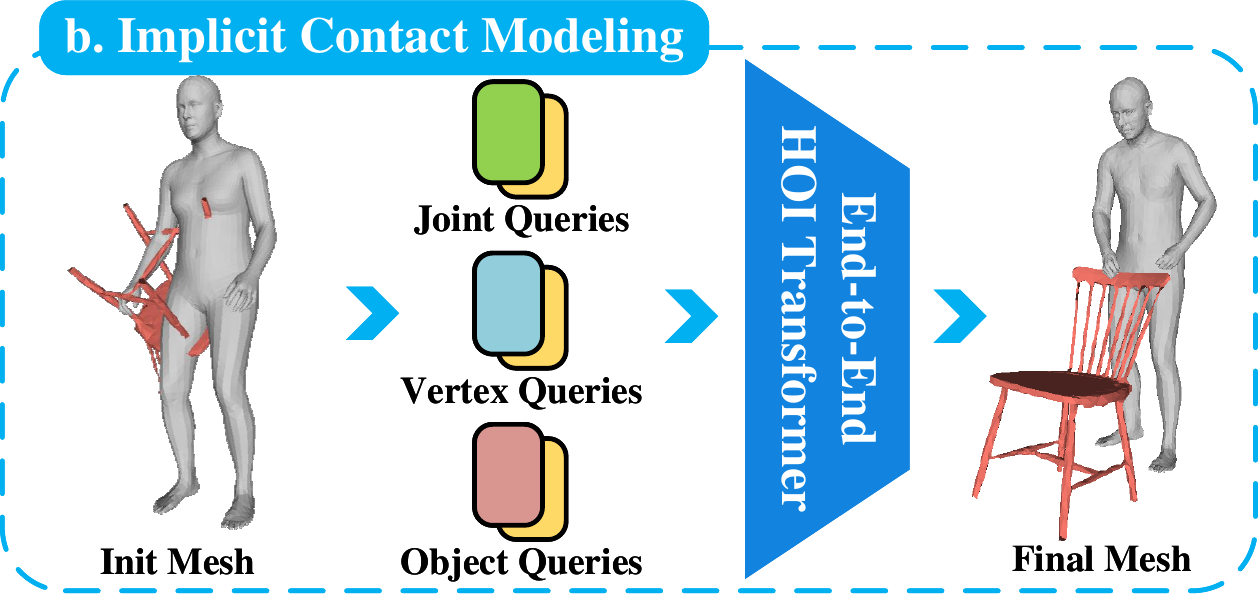}
        \phantomsubcaption
        \label{fig:motivation2}
    \end{subfigure}
    \caption{Comparison between existing explicit contact constraints for HOI reconstruction and our implicit contact modeling. 
    }
    \label{fig:motivation}
\end{figure}

Significant breakthroughs have recently advanced in reconstructing 3D human meshes from a single image~\cite{hand4whole,hmr1,hmr2,hmr3}. Such an advancement stimulates the study of joint reconstruction of 3D human and object meshes, which has numerous applications in areas such as augmented reality (AR)~\cite{AR}, virtual reality (VR)~\cite{VR}, and robotic manipulation~\cite{robotics}. Unlike reconstructing human meshes from a single image, Human-Object Interaction (HOI) reconstruction involves capturing the relative posture and interactions between humans and objects. Therefore, it presents considerable challenges, primarily because of the intricate articulated interactions and occlusions.


Recently, several studies \cite{chore,joint,stackflow} have explored leveraging interaction information for jointly modeling human and object reconstruction. These works generally incorporate interaction representations to impose hard constraints on the models. As shown in \cref{fig:motivation1}, (i) StackFLOW \cite{stackflow} utilizes the human-object offsets between anchors, densely sampled from the surfaces of the human and object meshes, to represent their spatial relationships. (ii) CHORE \cite{chore} predicts a part correspondence field to identify which body parts are in contact with object points. (iii) CONTHO \cite{joint} estimates vertex-level human-object contact maps to prevent the learning of erroneous correlations between humans and objects.

Although leveraging contact constraints on reconstruction models seems suitable for the HOI task, a natural conflict appears. On the one hand, mesh reconstruction of humans and objects prioritizes an overall relative position between them. On the other hand, explicit interaction constraints—such as human-object offsets and contact maps—focus on local relationships. Achieving a good balance between these two types of constraints is not easy. For instance, StackFLOW \cite{stackflow} heavily relies on a time-consuming post-optimization process during inference to improve the overall reconstruction quality. 
To address this issue, as shown in \cref{fig:motivation2}, we propose a straightforward framework that implicitly incorporates interaction-inclusive mesh reconstruction within a transformer architecture without additional explicit interaction constraints.

Given the remarkable success of transformers in human mesh recovery \cite{end2end,hmr3,graphormer,hmr2}, devising such a transformer-based framework for HOI reconstruction seems intuitive. However, three main challenges exist: 1) Using the same features across all 3D points would limit the ability to differentiate between various 3D modalities. 
2) Learning interaction poses directly from static templates is much more intricate than that in human mesh recovery. 3) Simply employing self-attention would confuse the local boundaries between the human and the object, which blurs the distinction between the reconstruction of the vertices from these two relatively independent targets.

To address the first two challenges, we draw inspiration from previous works \cite{hu2024learning,joint} and provide transformers with more precise feature inputs. Specifically, our approach involves: 1) Applying grid sampling to the initial vertices, which allows us to assign distinct features to each vertex, thereby improving the discriminative power of the 3D models. 2) Utilizing the initial mesh rather than a template mesh to mitigate the difficulties associated with learning complex interactions within the model. For the third challenge, we propose integrating specific graph convolutional structures into the transformer architecture for both human and object components. Such structures have been proven successful in human mesh reconstruction~\cite{graphormer}. In our case, the integration facilitates better fusion of local information, thereby improving the ability to differentiate between the reconstructions of the human and the object.

Therefore, we propose an End-to-End Human-Object Interaction Reconstruction Transformer with Graph-based Encoding, referred to as HOI-TG, to tackle the above challenges. First, we generate initial mesh vertices for the human and the object. These initial 3D coordinates are concatenated with the grid-sampled features to create a robust input, including the queries of joint, vertex, and object as the input to our transformer. The enhanced input improves the model's capacity to differentiate between 2D and 3D representations while simplifying the learning process for complex interactions via the attention mechanism. 
We reconstruct the coordinates using the output from a linear layer, derive the final human vertices based on a predefined upsampling matrix, and predict the object's pose—translation and rotation—through rigid body transformation using template-based methods. Owing to the transformer architecture, our model can effectively focus on global interactions between humans and objects. To further improve the ability to integrate local features of both entities, we devise and integrate the Human Graph Residual Block and the Object Graph Residual Block within the transformer module. 


Compared with previous state-of-the-art results, our HOI-TG achieved 1) improvements of \textbf{8.0\%} (human) and \textbf{5.0\%} (object) in 3D reconstruction accuracy and enhanced precision by \textbf{3.4\%} and recall by \textbf{5.8\%} for reconstructed contact accuracy on the BEHAVE dataset~\cite{behave}; 2) improvements of \textbf{8.9\%} (human) and \textbf{8.6\%} (object) in 3D mesh reconstruction, along with \textbf{3.9\%} (precision) and \textbf{4.1\%} (recall) for contact accuracy on the InterCap dataset~\cite{intercap}. Experiments underscore the effectiveness of our HOI-TG framework in accurately modeling interactions and complex structures. Regarding reconstruction visualization, we observe more accurate relative positions between humans and objects with reduced instances of model penetration.


In summary, we make the following contributions.
\begin{itemize}
    \item We introduce a novel end-to-end HOI reconstruction transformer framework to model human and object meshes in interaction from a single image.
    \item We design the Human Graph Residual Block and the Object Graph Residual Block for the transformer encoder to enhance the distinction between two meshes and strengthen the capability of local feature fusion. 
    \item HOI-TG achieves new state-of-the-art performance on the BEHAVE and the InterCap datasets, 
    which highlights the effectiveness of the proposed framework.
\end{itemize}
\begin{figure*}[t]
  \centering
  \includegraphics[width=1.0\linewidth]{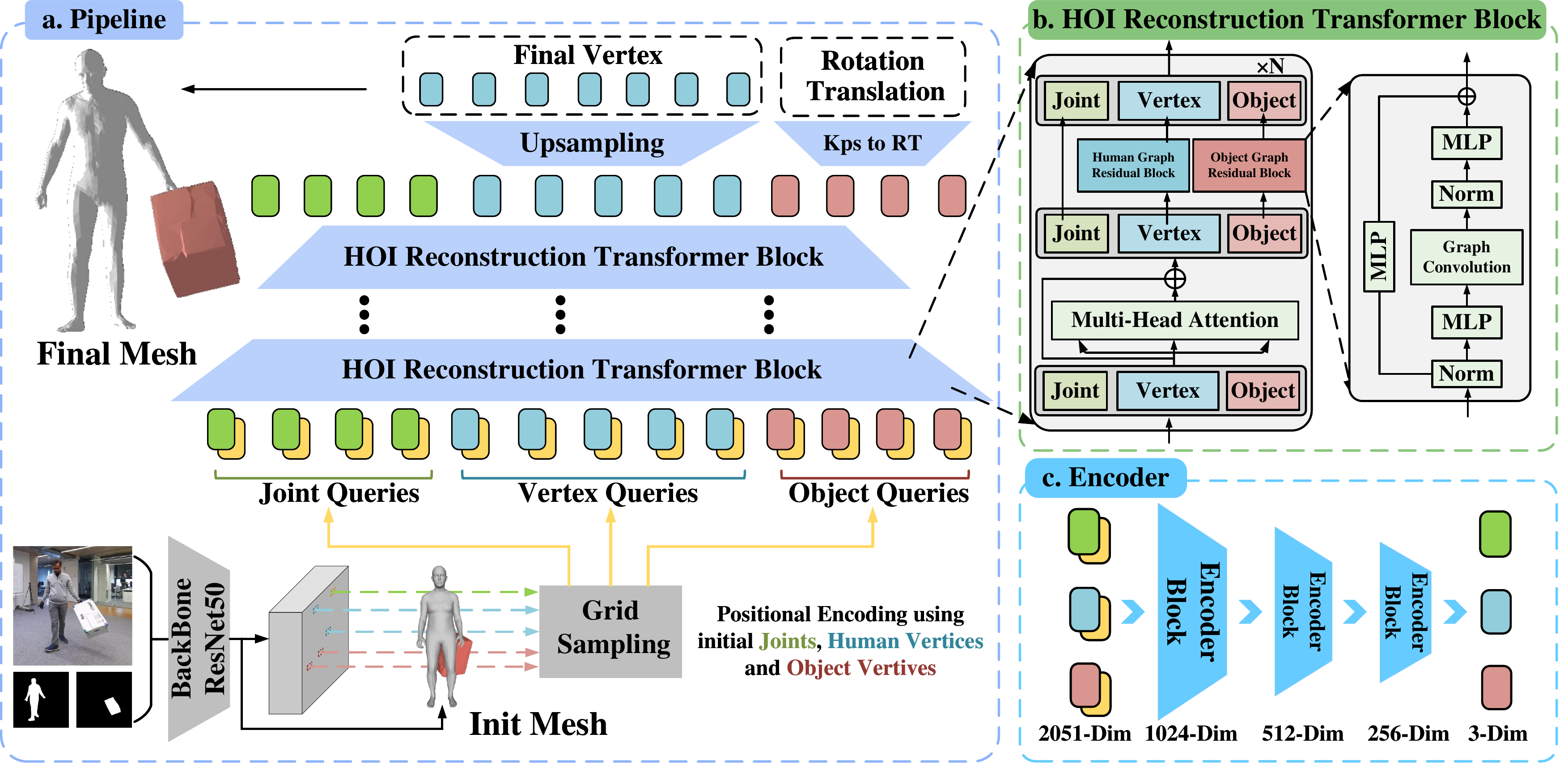}
  \caption{\textbf{Overview of our HOI-TG.} a. Pipeline draws the process of HOI reconstruction. Given the input image and human \& object segmentations, we extract the image feature and generate an initial human mesh using the ResNet50 backbone. Then, we prepare joint queries, vertex queries, and object queries by concatenating grid sampling features and per-vertex 3D coordinates. Based on the queries, HOI reconstruction transformer blocks reconstruct human joints \& vertices, and object mesh. Final HOI meshes are calculated by upsampling and rigid transformation. b. HOI Reconstruction Transformer Block contains a Human Graph Residual Block and an Object Graph Residual Block for separate encoding for humans and objects. c. Encoder shows the change of hidden dimensions throughout HOI-TG.
  }
  \label{fig:pipeline}
\end{figure*}

\section{Related Works}

\textbf{3D human reconstruction.} 3D human reconstruction from a monocular camera has been an active research topic for a long time. It is difficult due to complex pose variations, occlusions, and limited 3D training data. Prior studies~\cite{param1,param2,param3,param4,param5,param6,pixie,hand4whole} propose adopting the pre-trained parametric human models (\eg, SMPL \cite{smpl}, MANO \cite{MANO}). ExPose \cite{expose} directly predicts the 3D parameters of the body, face, and hands in the SMPL-X format, employing body-driven attention to localize the face and hand regions in the original image. PIXIE \cite{pixie} utilizes SMPL-X's unified shape space for all body parts, allowing every expert to contribute to the overall model. Hand4Whole \cite{hand4whole} presents a comprehensive framework for 3D human mesh estimation, reconstructing the body, hands, and face by leveraging features from 3D pose-guided grid sampling.

On the other hand, instead of adopting a parametric human model, non-parametric approaches regress vertices directly from an image~\cite{graphcmr,end2end,pose2mesh,i2l,graphormer,transformer1,DeFormer,joint,transformer3,transformer4}. Among the previous studies, graph convolutional neuralnetwork (GCN) is one of the most popular options as it can model the local interactions between neighboring vertices based on a given adjacency matrix~\cite{graphcmr,graphormer}. However, it is less efficient to capture global interactions between the human vertices and body joints. To overcome this limitation, transformer-based methods use a self-attention mechanism to attend vertices and body joints freely in the mesh and thereby encode the non-local relationship of a human mesh. 
For instance, METRO \cite{end2end} leverages an encoder-based transformer to model non-local intersections among mesh vertices and joints. Graphormer \cite{graphormer} integrating GCN into the transformer architecture to improve mesh vertices local modeling. Deformer \cite{DeFormer} proposes using new body-mesh-driven attention modules to extract multi-scale visual features, enhancing the efficiency of 3D mesh recovery.

Inspired by the success of transformers-based methods in human mesh recovery, we propose a novel transformer framework suitable for HOI reconstruction. 

\noindent\textbf{3D human and object reconstruction.}
Many early studies on human-scene contact estimation have focused on representing contact through 2D projections \cite{2Dcontact}, 3D joint-level interactions \cite{3D1,3D2}, or 3D patch-level contacts \cite{3Dp1}. More recently, many researchers have begun to explore the estimation of dense vertex-level contact maps that are defined on the human body surface~\cite{ver_contact1,ver_contact2,ver_contact3,ver_contact4,joint}. Advances in human-scene contact estimation facilitate the study of human-object interaction reconstruction.
HODome \cite{hdome} introduces a layer-wise neural network to reconstructe humans and objects from multi-view. HolisticMesh \cite{HolisticMesh} and PHOSA \cite{PHOSA} present optimization frameworks that are rooted in physics-based and contact-based approaches, respectively. CHORE \cite{chore} compute a neural reconstruction of human and object represented implicitly with two unsigned distance fields. StackFLOW \cite{stackflow} infers the posterior distribution of human-object spatial relations from the image, and optimizes the human body pose and object 6D pose. CONTHO \cite{joint} proposes contact-based refinement to prevent the learning of erroneous correlations between humans and objects. 

The above methods impose explicit constraints for HOI, which causes a conflict between local contact prediction and global mesh reconstruction. Thus, we propose employing a transformer architecture that implicitly learns character interaction relationships.

\section{Method}
Given an RGB image depicting an HOI event and the 3D mesh template of the corresponding object, the objective of HOI reconstruction is to reconstruct the human mesh and determine the object's rotation and translation relative to the template.
We propose a novel transformer-based framework, namely the HOI Reconstruction Transformer with Graph-based Encoding (HOI-TG). To effectively model complex interactions between humans and objects, HOI-TG combines the strengths of transformer architectures and GCN. To ensure the pipeline functions as intended, we first prepare joint queries, vertex queries, and object queries as inputs by grid sampling. Then, the Multi-Layer Transformer conducts HOI reconstruction. \cref{fig:pipeline}\textcolor{cvprblue}{a} shows the overall pipeline of HOI-TG.

\subsection{Preparation of queries} \label{sec:input}
Transferring a transformer-based human reconstruction method, such as METRO \cite{end2end}, directly to this task failed to yield satisfactory results. We identified two primary reasons for the failure. First, using the same 2D features at all 3D points diminishes the differentiation among various 3D modalities.
Second, learning interaction poses from a static template is very challenging. Drawing inspiration from CONTHO \cite{joint}, we develop an initialization strategy to provide the transformer with more effective input.

The input tensor $\mathbf{I}_{\textrm{input}} \in \mathbb{R}^{5 \times H \times W}$ comprises an RGB image $\mathbf{I} \in \mathbb{R}^{3 \times H \times W}$, a segmentation mask $\mathbf{S}_h \in \mathbb{R}^{1 \times H \times W}$ of the human, and a segmentation mask $\mathbf{S}_o \in \mathbb{R}^{1 \times H \times W}$ of the object. $H$ and $W$ represent the height and width of the image, respectively.
A ResNet50 backbone network \cite{hand4whole} pre-trained for human mesh reconstruction is employed to extract features $\mathbf{F} \in \mathbb{R}^{2048 \times H/32 \times W/32}$ from $\mathbf{I}_{\textrm{input}}$. Meanwhile, the backbone also produces a rough estimation of body parameters $\theta_{\textrm{body}} \in \mathbb{R}^{76}$ and hand parameters $\theta_{\textrm{hand}} \in \mathbb{R}^{90}$. The predicted parameters are subsequently forwarded to the SMPLH \cite{MANO} model, resulting in the human mesh $\mathbf{M}_h^{\textrm{init}*} \in \mathbb{R}^{6890 \times 3}$ and joints $\mathbf{M}_j^{\textrm{init}} \in \mathbb{R}^{73 \times 3}$. We also predict camera parameters from $\mathbf{F}$ for subsequent grid sampling and the 2D projection of joint loss calculations.
To reduce computational cost and redundancy in the original mesh, we downsample $\mathbf{M}_h^{\textrm{init}*} \in \mathbb{R}^{6890 \times 3}$ to obtain $\mathbf{M}_h^{\textrm{init}} \in \mathbb{R}^{431 \times 3}$, following \cite{end2end,graphormer,reduce}. 

For the object mesh $\mathbf{M}_o^{\textrm{init}}$, we predict the init rotation $\mathbf{R}_{\textrm{init}}$ and translation $\mathbf{T}_{\textrm{init}}$ by adding a linear layer after the pre-trained feature extractor. Since the linear layer is randomly initialized, the initial stage does not incorporate interaction information between humans and objects. The initial results are represented by the ``Init mesh" in \cref{fig:pipeline}\textcolor{cvprblue}{a}. This stage aims to provide more 3D recognizable input features for the subsequent transformer encoder. 

The 3D queries features $\mathbf{Q}_j$, $\mathbf{Q}_{hv}$, and $\mathbf{Q}_{ov}$ consist of grid sampling features and per-vertex 3D coordinates as positional encoding. The grid sampling features are obtained by projecting the 3D vertices ($\mathbf{M}_j^{\textrm{init}}$, $\mathbf{M}_h^{\textrm{init}}$, and  $\mathbf{M}_o^{\textrm{init}}$) into a 2D (\textit{x,y}) coordinate system using the predicted camera parameters, followed by extracting the corresponding coordinate features from $\mathbf{F}$. We then form the 3D queries features ($\mathbf{Q}_j$, $\mathbf{Q}_{hv}$ and $\mathbf{Q}_{ov}$) by concatenating the grid sampling features with the per-vertex 3D coordinates from the initial meshes ($\mathbf{M}_j^{\textrm{init}}$, $\mathbf{M}_h^{\textrm{init}}$, and $\mathbf{M}_o^{\textrm{init}}$). Consequently, the final 3D features for human joints, human vertices, and objects vertices are represented as $\mathbf{Q}_j \in \mathbb{R}^{73 \times (2048 + 3)}$, $\mathbf{Q}_{hv} \in \mathbb{R}^{431 \times (2048 + 3)}$ and $\mathbf{Q}_{ov} \in \mathbb{R}^{64 \times (2048 + 3)}$.

\subsection{Transformer encoder for HOI reconstruction}
Traditional transformer architectures have fixed hidden layers, which poses challenges for directly predicting 3D vertices from high-dimensional features. To address this issue, as shown in \cref{fig:pipeline}\textcolor{cvprblue}{c}, we follow \cite{end2end,graphormer,reduce} and adopt a decreasing hidden layer strategy to reconstruct 3D coordinates. Specifically, we use linear layers to reduce the dimensionality at each encoder layer. As illustrated in \cref{fig:pipeline}\textcolor{cvprblue}{a}, our HOI-TG reconstructs human joints, 3D human and object meshes, denoted as $\mathbf{M}_j$, $\mathbf{M}_h$ and $\mathbf{M}_o$, from the 3D queries features $\mathbf{Q}_{j}$, $\mathbf{Q}_{hv}$, and $\mathbf{Q}_{ov}$.


Given the 3D queries features ($\mathbf{Q}_j$, $\mathbf{Q}_{hv}$, and $\mathbf{Q}_{ov}$), we concatenate all 3D queries features to obtain $\mathbf{Q} \in \mathbb{R}^{(73 + 431 + 64) \times 2051}$. The multi-layer transformer-based encoder is employed to capture the attention relationships among humans, among objects, and between humans and objects.
As shown in \cref{fig:pipeline}\textcolor{cvprblue}{c}, our encoder is composed of three HOI Reconstruction Transformer Blocks. The three encoder blocks maintain the same number of tokens but have different hidden dimensions. In this work, we follow the methodology outlined in \cite{end2end}, where the hidden dimensions of the three encoders are set to 1024, 512, and 256, respectively. 
We denote the encoded features of the joint, vertex, and object as $\mathbf{Q}^{'}_j$, $\mathbf{Q}^{'}_{hv}$, and $\mathbf{Q}^{'}_{ov}$, respectively.

Ultimately, we predict the 3D human joints $\mathbf{M}_j$, human mesh vertices $\mathbf{M}_h$, and object mesh vertices $\mathbf{M}_o$ from $\mathbf{Q}^{'}_j$, $\mathbf{Q}^{'}_{hv}$, and $\mathbf{Q}^{'}_{ov}$. For computational efficiency and redundancy reduction, we downsample the human vertices during the input stage and then upsample them back to the original scale in the prediction stage.
We utilize the predefined upsampling and downsampling matrices \cite{sample} for these operations.
For the prediction of $\mathbf{R}$ and $\mathbf{T}$, we calculate the rigid transform between the predicted object vertex $\mathbf{M}_o$ and the object template.


\begin{figure}[t]
    \centering
    \begin{subfigure}{.2\textwidth}
        \includegraphics[width=\linewidth]{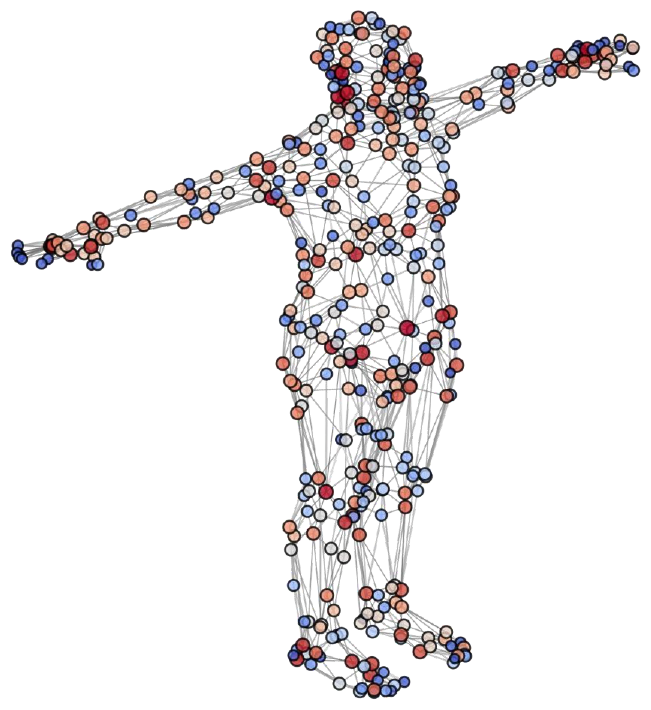}
        \caption{Human graph}
        \label{fig:hgraph}
    \end{subfigure}
    \hspace{0.4cm}
    \begin{subfigure}{.2\textwidth}
        \includegraphics[width=\linewidth]{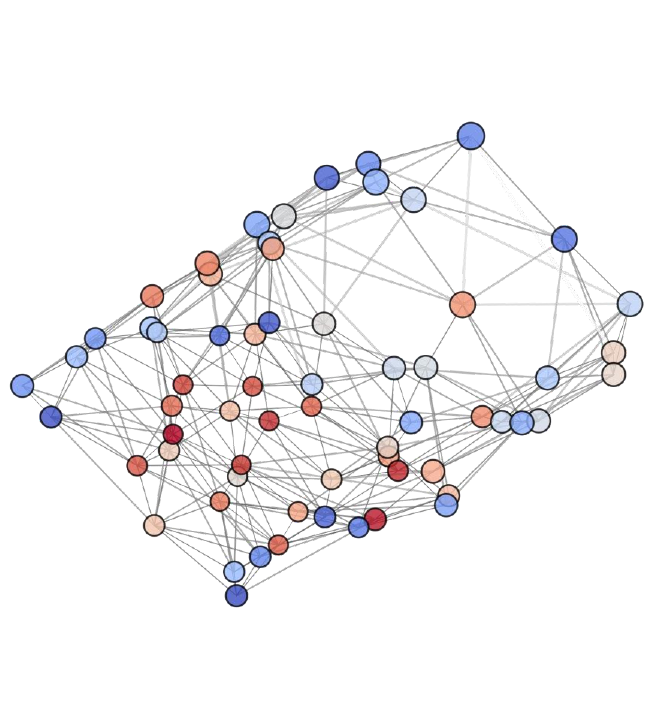}
        \caption{Object graph}
        \label{fig:ograph}
    \end{subfigure}
    \caption{\textbf{Graph adjacency of the human and a specific object.} 
    The adjacency contains connectivity and distance information. Warmer colors indicate vertices with higher centrality. 
    }
    \label{fig:graph}
\end{figure}

\subsection{Graph-based encoding for human and object}\label{GHO}
Using self-attention to learn the global interaction relationships between humans and objects may disrupt the existing topological relationships. To address this issue, while employing the self-attention mechanism to capture global human interaction dynamics, we incorporate corresponding graph convolutional structures for both humans and objects to enhance the modeling of their respective topological structures.

As shown in \cref{fig:pipeline}\textcolor{cvprblue}{b} 
, we set four layers for each encoder block. For each layer, given the 3D query features, the input features $\mathbf{Q}$ are processed sequentially through the multi-head attention module and the Graph Residual Blocks. We design the Human Residual Block for the human vertices and the Object Residual Block for the object vertices. 

\noindent\textbf{Human Graph Residual Block.}  
Given the human mesh features $\mathbf{Q}^{\textrm{mid}}_{hv} \in \mathbb{R}^{n \times d}$ generated by a multi-head attention module, we enhance local vertex interactions using graph convolution:
\begin{equation}
  \mathbf{Q}^{\textrm{mid}'}_{hv} = \mathbf{GraphConv}(\bar{\mathbf{A}}, \mathbf{Q}^{\textrm{mid}}_{hv}; \mathbf{W}_{G}) = \sigma(\bar{\mathbf{A}} \mathbf{Q}^{\textrm{mid}}_{hv} \mathbf{W}_{G}).
  \label{graph}
\end{equation}
Here, $\bar{\mathbf{A}}$ denotes the adjacency matrix of the graph as shown in \cref{fig:hgraph}, and $\mathbf{W}_{G}$ represents the trainable parameters. The function $\sigma(\cdot)$ is the GeLU activation function. We adhere to the design principles outlined in \cite{graphormer,graphcmr} to construct this block.

\noindent\textbf{Object Graph Residual Block.}  
The Object Graph Residual Block is analogous to the Human Graph Residual Block. However, due to different topological structures, the predefined $\bar{\mathbf{A}}$ in the graph convolution varies. We have defined different graph adjacency matrices for various object templates, where a specific example shows in \cref{fig:ograph}. We utilize the K-Nearest Neighbors Graph algorithm to extract the graph structure corresponding to distinct object templates. More details about the implementation can be found in the supplementary material.

\setlength{\tabcolsep}{1.5pt}
\begin{table*}[h]
  \begin{subtable}[h]{0.65\textwidth}
  \centering
  \begin{tabular}{@{}lcccc|cccc@{}}
    \toprule
           & \multicolumn{4}{c|}{BEHAVE} & \multicolumn{4}{c}{InterCap} \\ \midrule
    Methods & \small{CD$_\textrm{human}$$\downarrow$}   & \small{CD$_\textrm{object}$$\downarrow$}   & \small{Contact$_\textrm{p}$$\uparrow$}    & \small{Contact$_\textrm{r}$$\uparrow$} & \small{CD$_\textrm{human}$$\downarrow$}   & \small{CD$_\textrm{object}$$\downarrow$}   & \small{Contact$_\textrm{p}$$\uparrow$}    & \small{Contact$_\textrm{r}$$\uparrow$} \\ \midrule
    METRO~\cite{end2end} & 46.82 & 59.13 & 0.084 & 0.520 & 42.83 & 82.12 & 0.049 & 0.641 \\ 
    Graphormer~\cite{graphormer} & 42.40 & 36.60 & 0.047 & 0.348 & 38.82 & 46.18 & 0.031 & 0.332  \\ \hline\midrule
    PHOSA \cite{PHOSA}  & 12.17 & 26.62 & 0.393 & 0.266 & 11.20 & 20.57 & 0.228 & 0.159 \\
    CHORE \cite{chore}  & 5.58  & 10.66 & 0.587 & 0.472 & 7.01  & 12.81 & 0.339 & 0.253 \\
    CONTHO \cite{joint} & 4.99  & 8.42  & 0.628 & 0.496 & 5.96  & 9.50  & 0.661 & 0.432 \\ 
    HOI-TG (Ours)   & \textbf{4.59} & \textbf{8.00} & \textbf{0.662} & \textbf{0.554} & \textbf{5.43} & \textbf{8.68} & \textbf{0.700} & \textbf{0.473} \\ \bottomrule
  \end{tabular}
  \end{subtable}
  \hspace{0.2cm}
  \begin{subfigure}[h]{0.4\textwidth}
  \centering
      \includegraphics[height=0.6\textwidth]{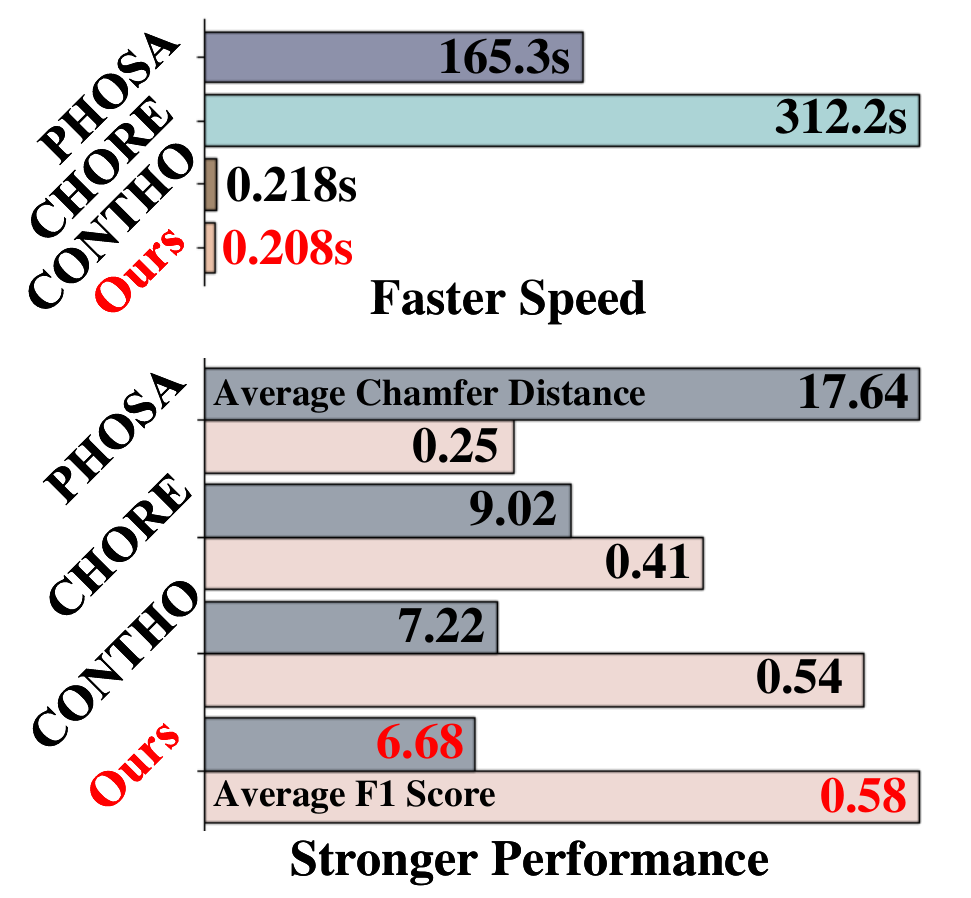}
  \end{subfigure}
  \caption{Comparison of various methods on the BEHAVE~\cite{behave} and InterCap~\cite{intercap} datasets. 
  As a reference, we report the results of directly transferring the transformer-based human mesh recovery methods METRO~\cite{end2end} and Graphormer~\cite{graphormer} to the HOI reconstruction task. 
  The best results are marked as \textbf{bold}. We also compare the running time, average Chamfer distance, and average F1 score (${=}2{\times}p{\times}r/(p{+}r)$).
  }
  \label{tab:comparison}
\end{table*}





\setlength{\tabcolsep}{0pt}
\begin{figure*}
    \centering
    \begin{tabular}{cccccc}
    \includegraphics[width=0.1\textwidth]{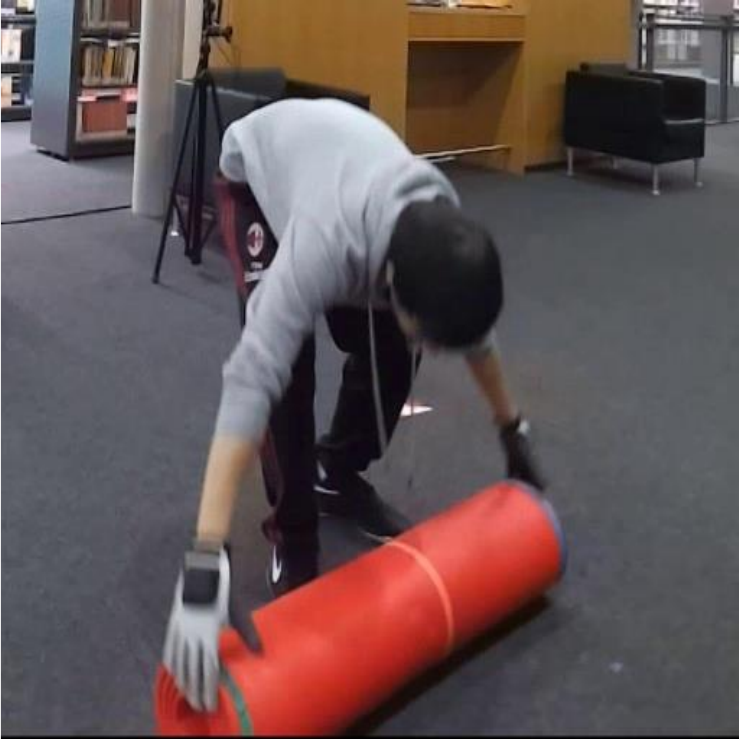}     & \includegraphics[width=0.2\textwidth]{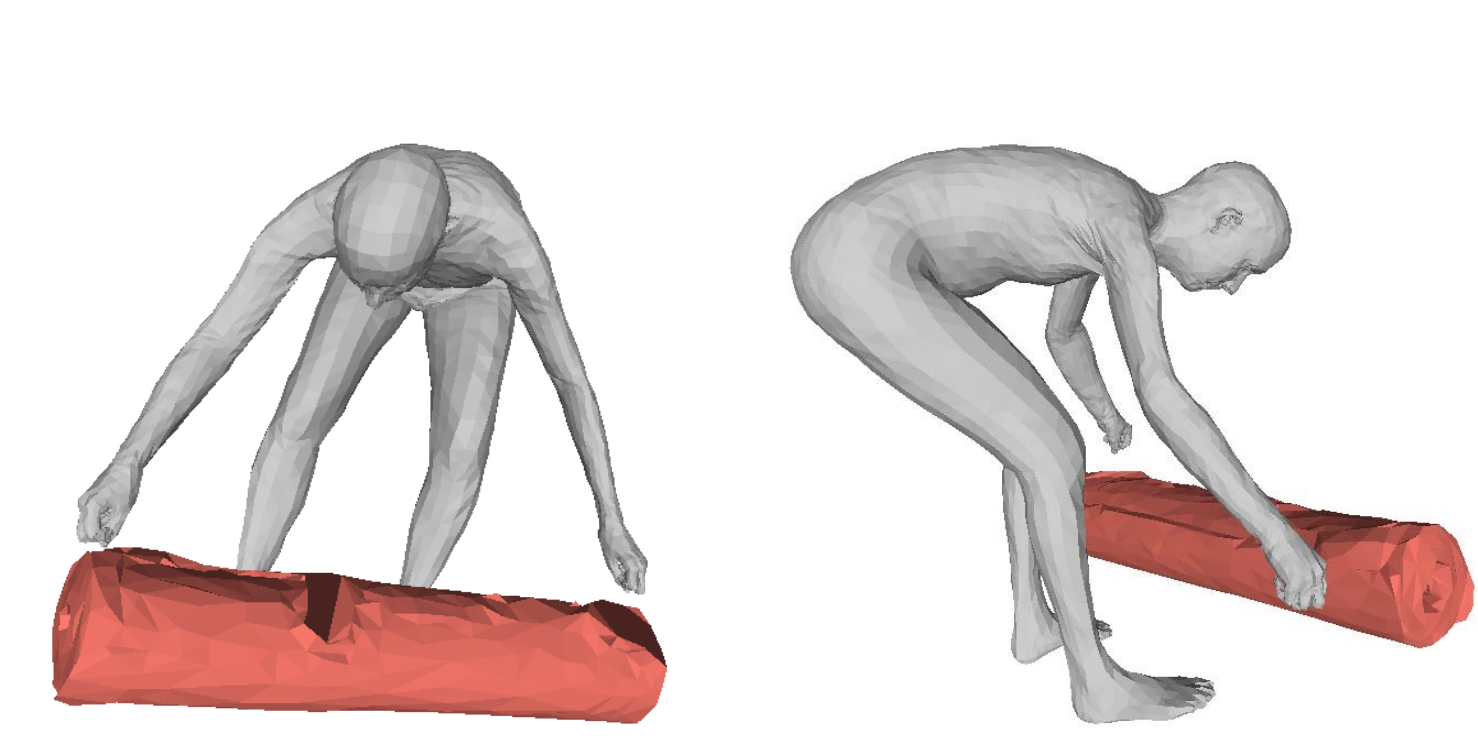} & \includegraphics[width=0.2\textwidth]{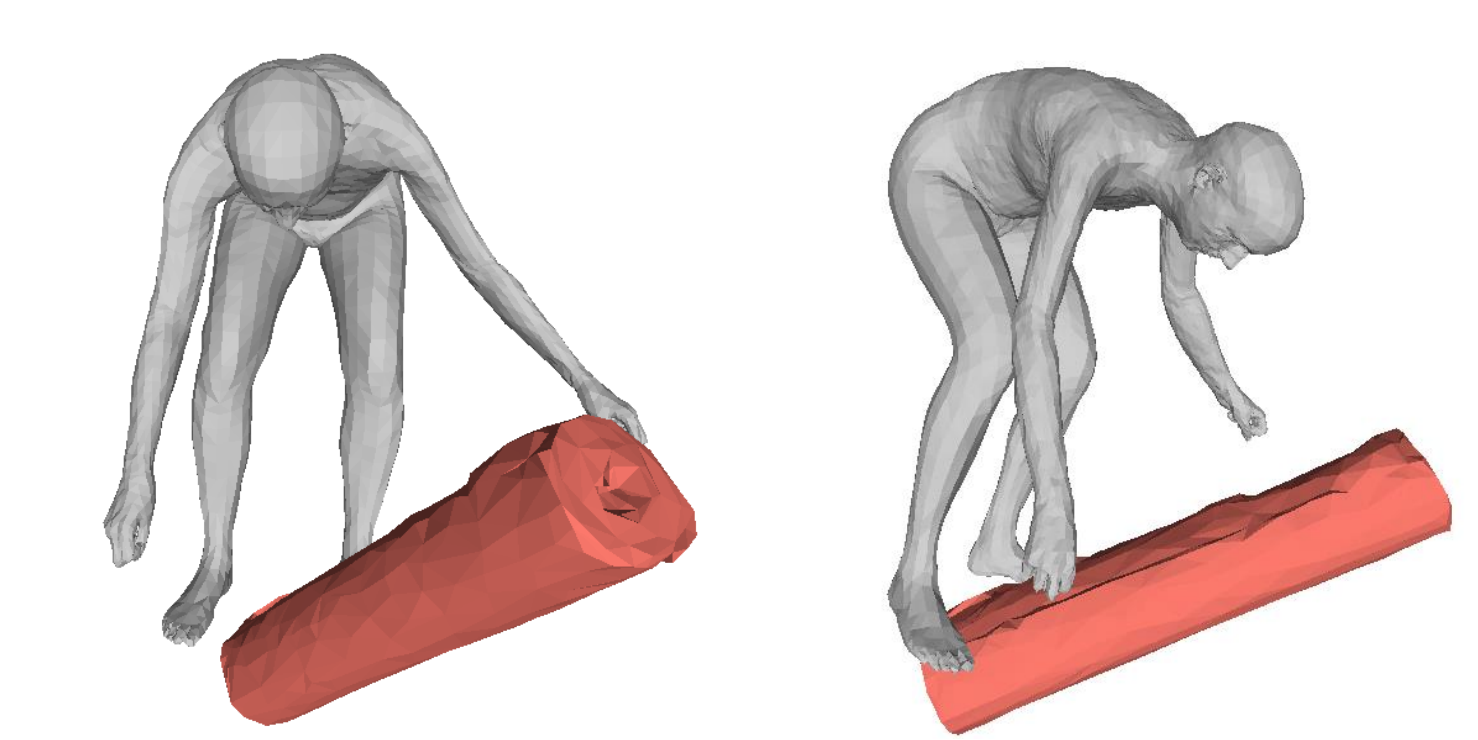} & \includegraphics[width=0.1\textwidth]{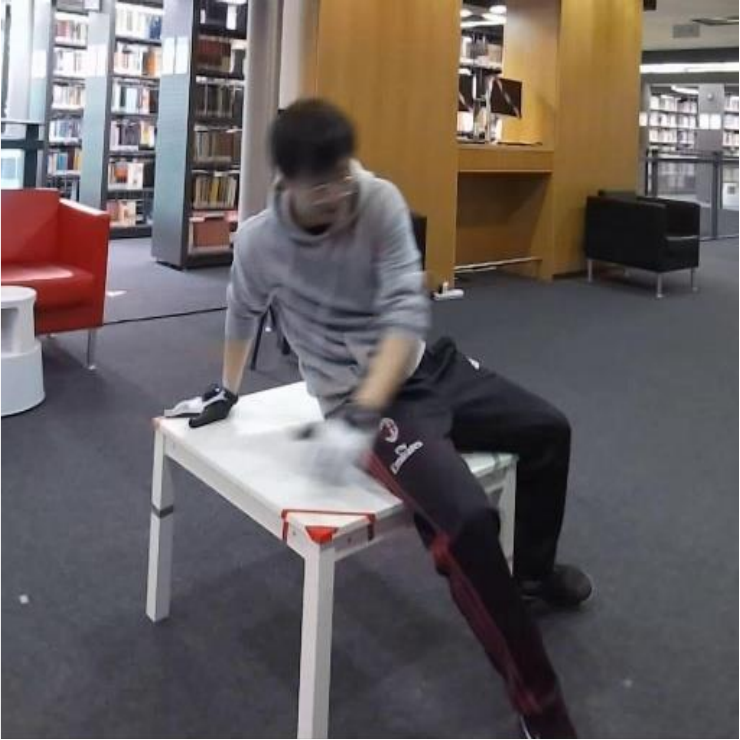} & \includegraphics[width=0.2\textwidth]{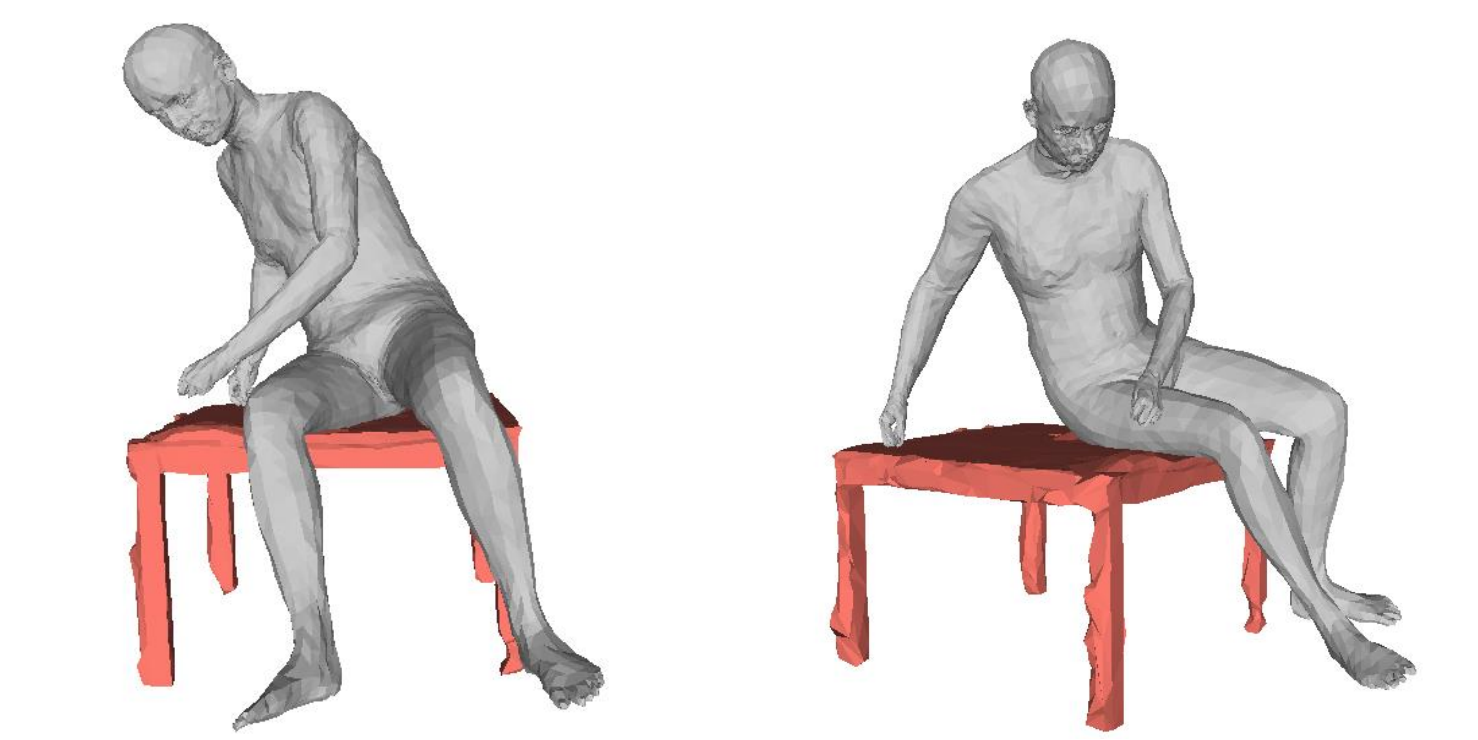} & \includegraphics[width=0.2\textwidth]{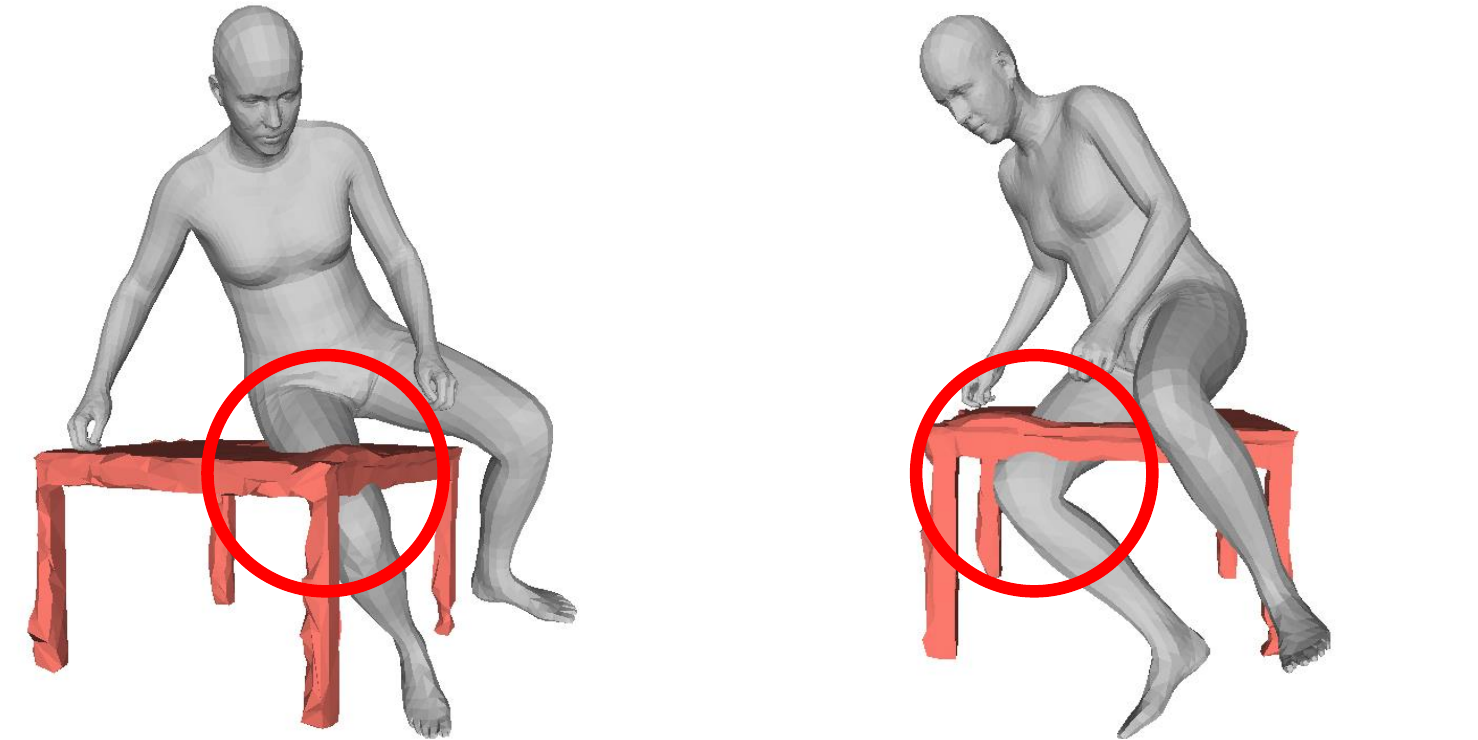} \\
    \includegraphics[width=0.1\textwidth]{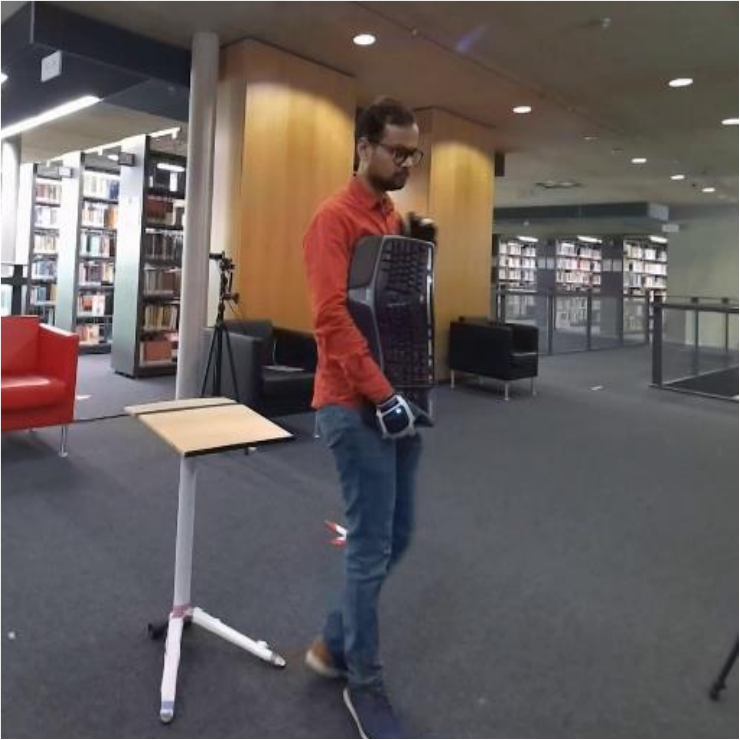}     & \includegraphics[width=0.2\textwidth]{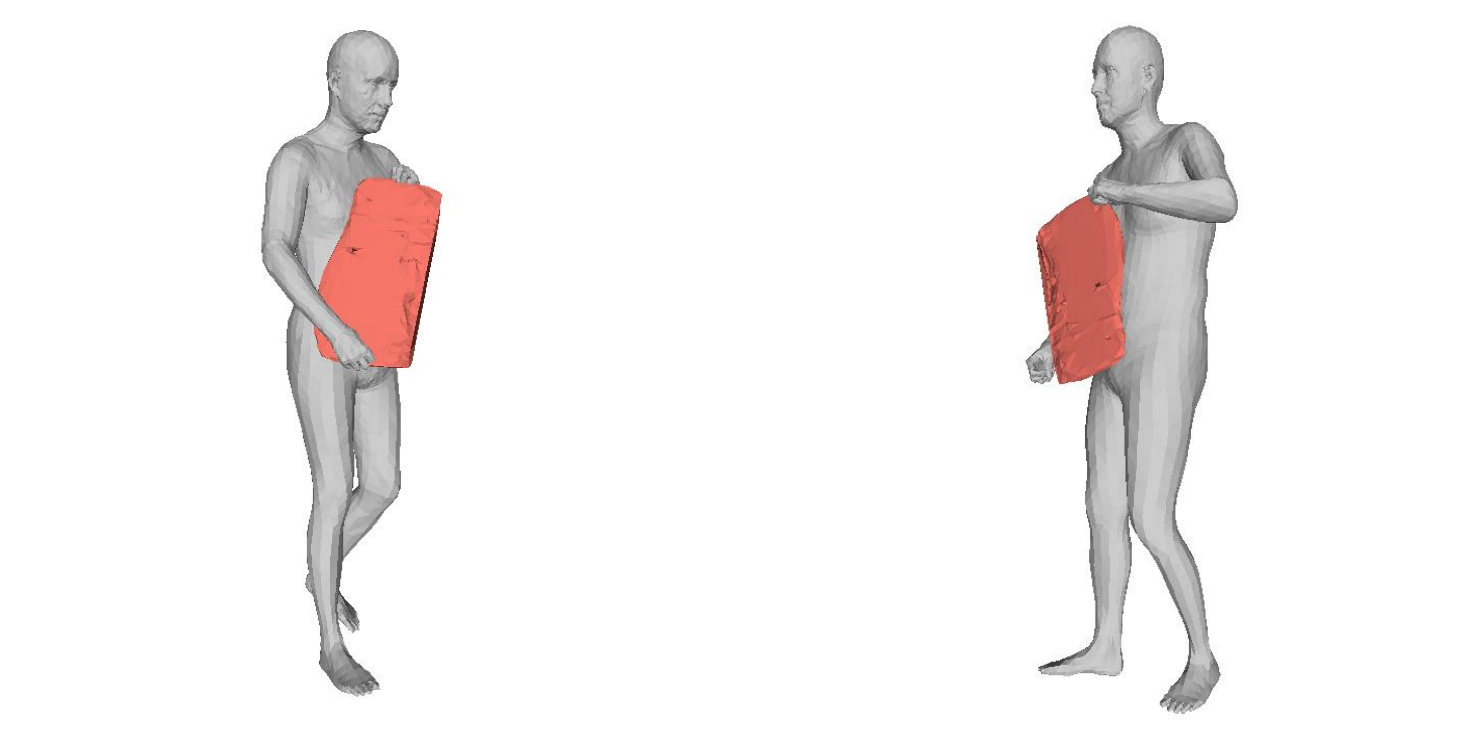} & \includegraphics[width=0.2\textwidth]{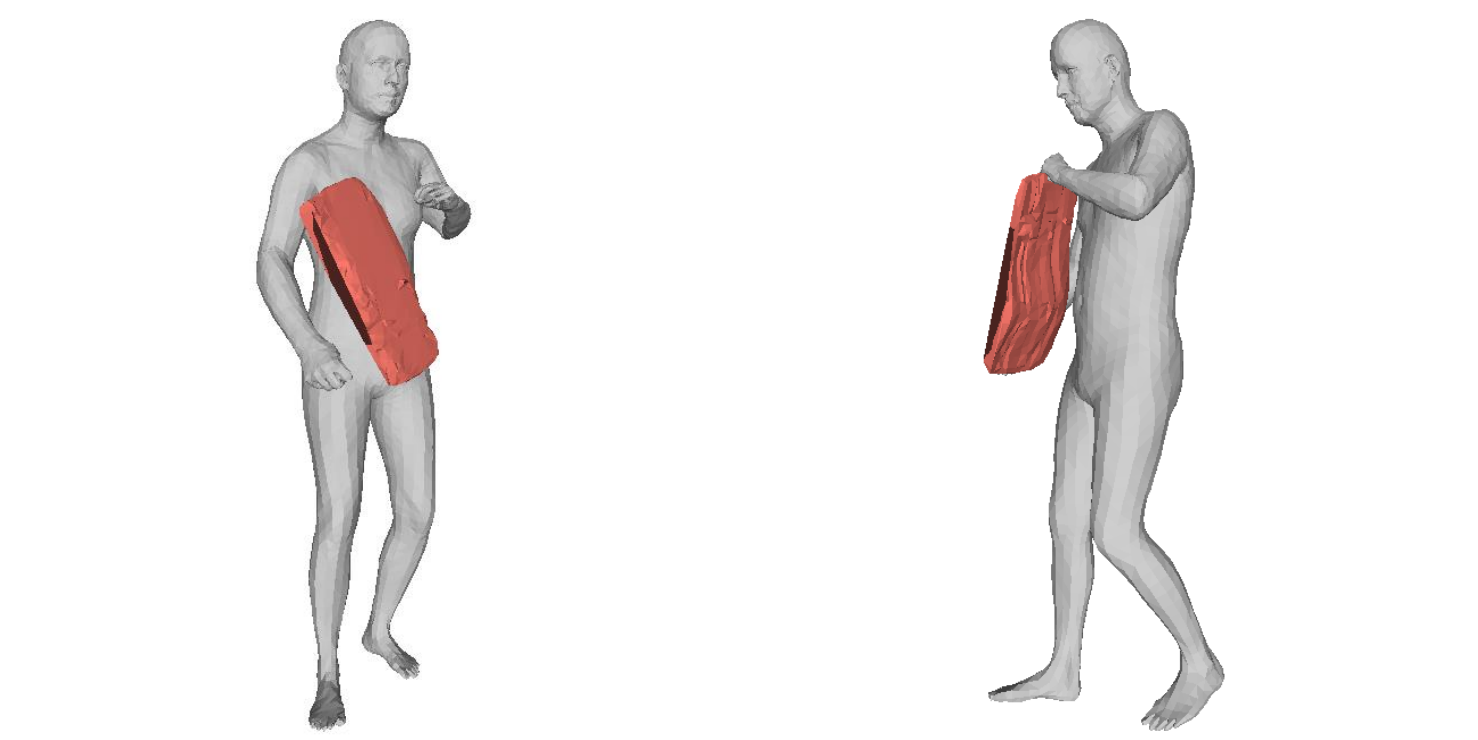} & \includegraphics[width=0.1\textwidth]{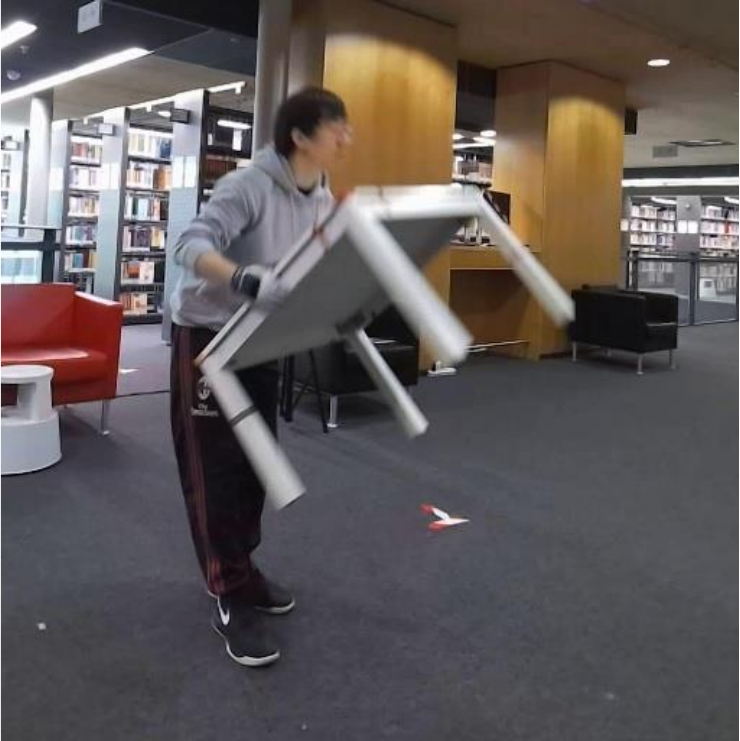} & \includegraphics[width=0.2\textwidth]{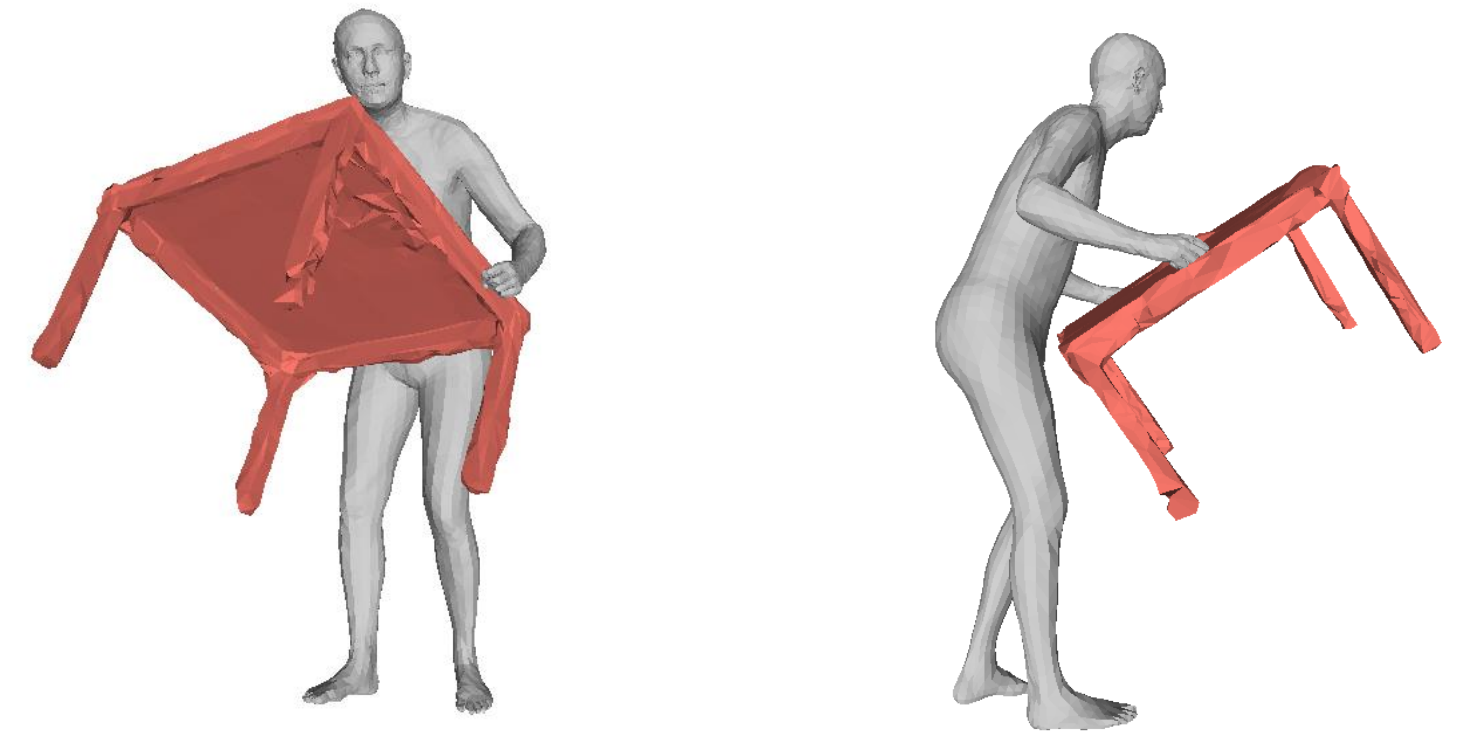} & \includegraphics[width=0.2\textwidth]{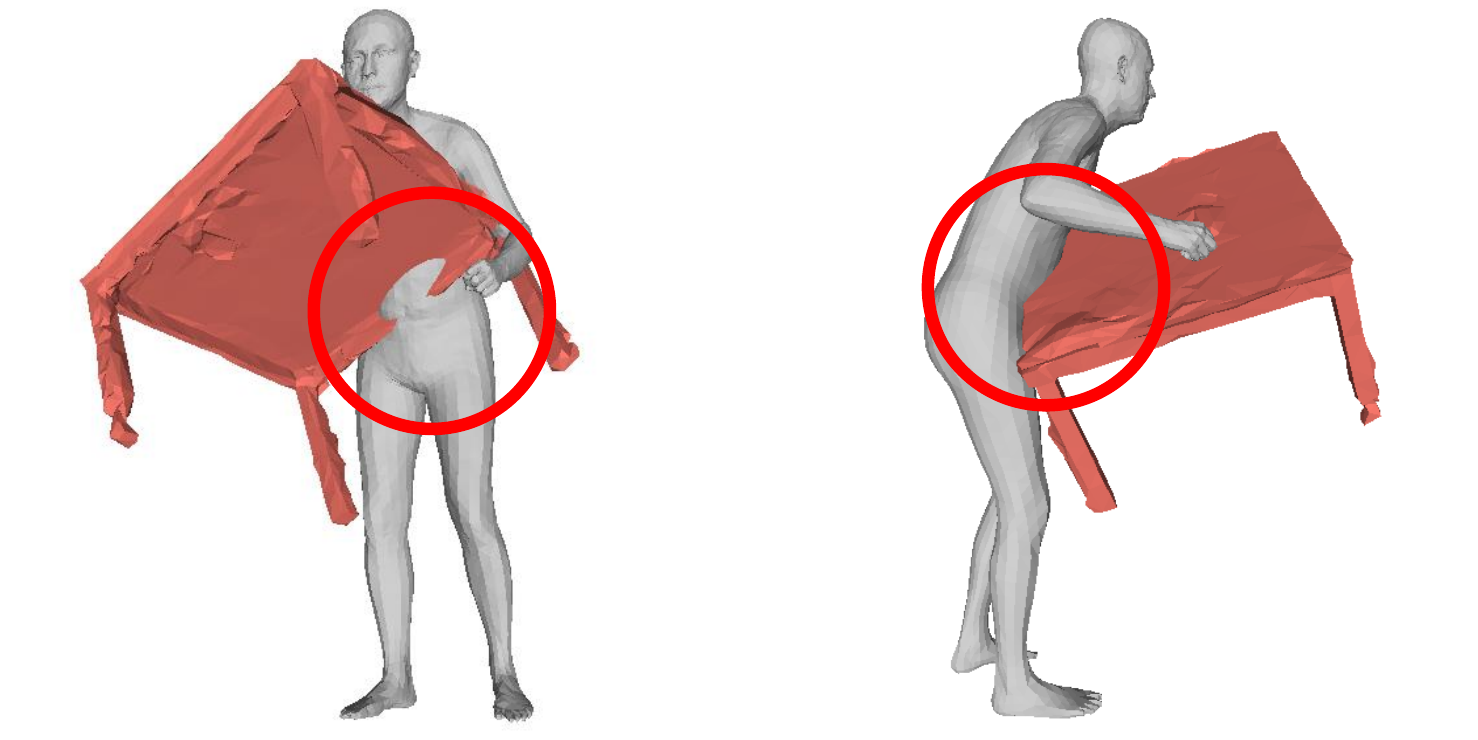} \\
    \includegraphics[width=0.1\textwidth]{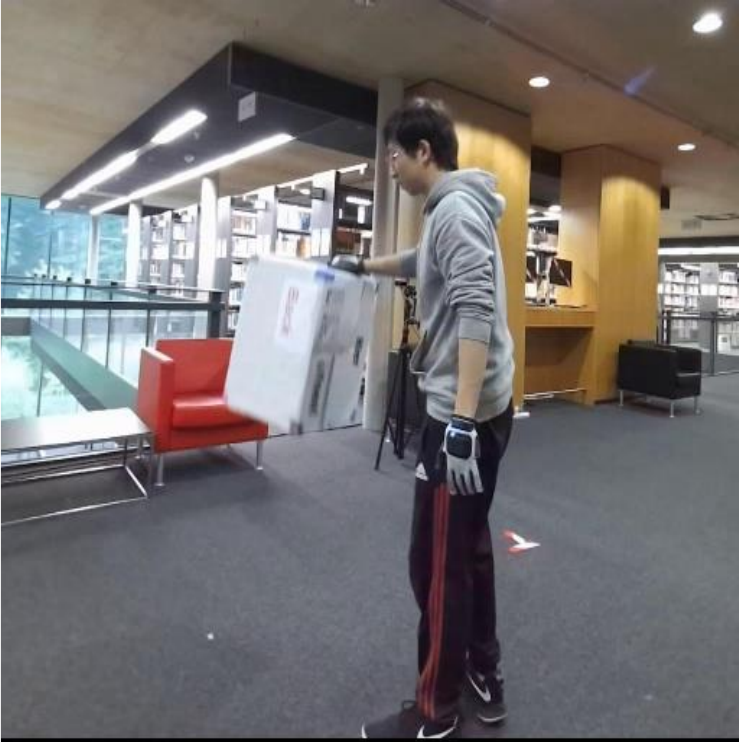}     & \includegraphics[width=0.2\textwidth]{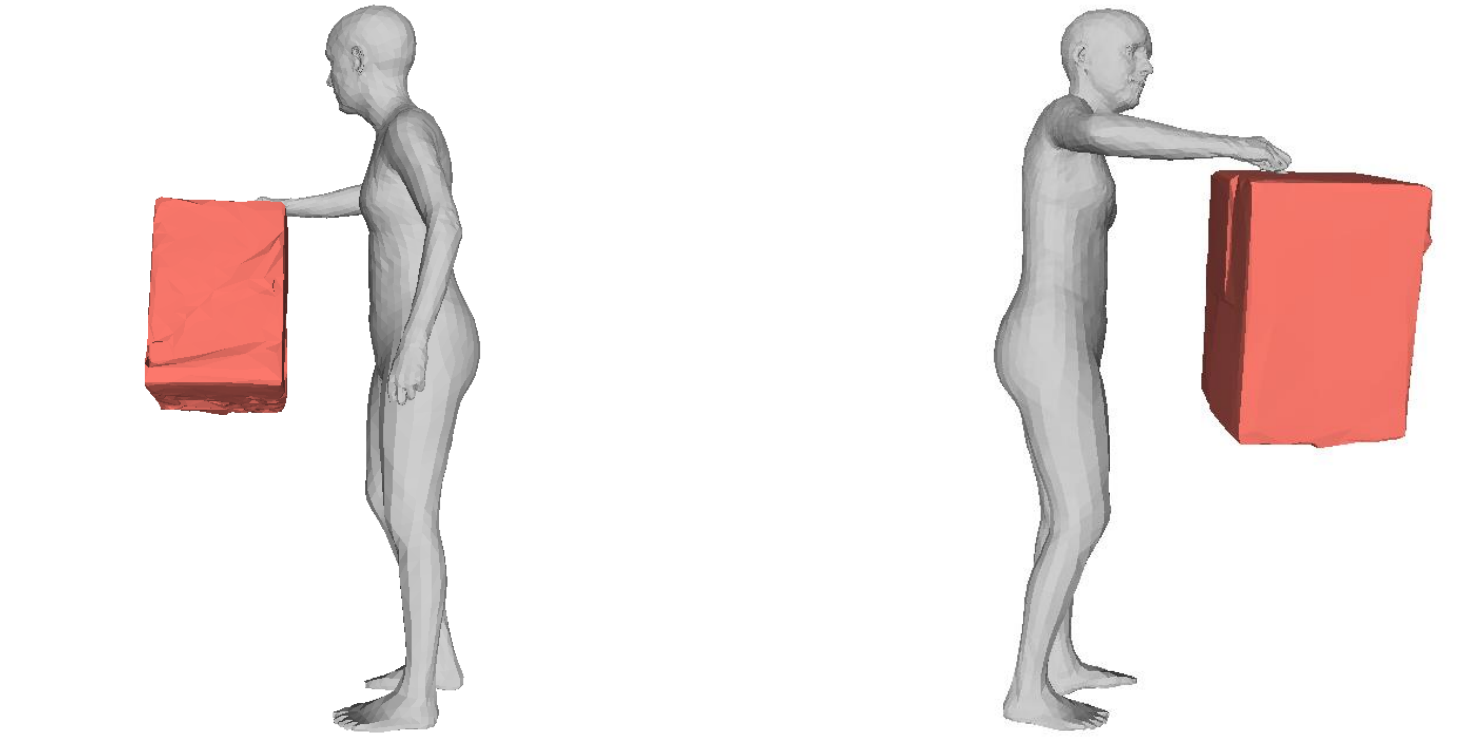} & \includegraphics[width=0.2\textwidth]{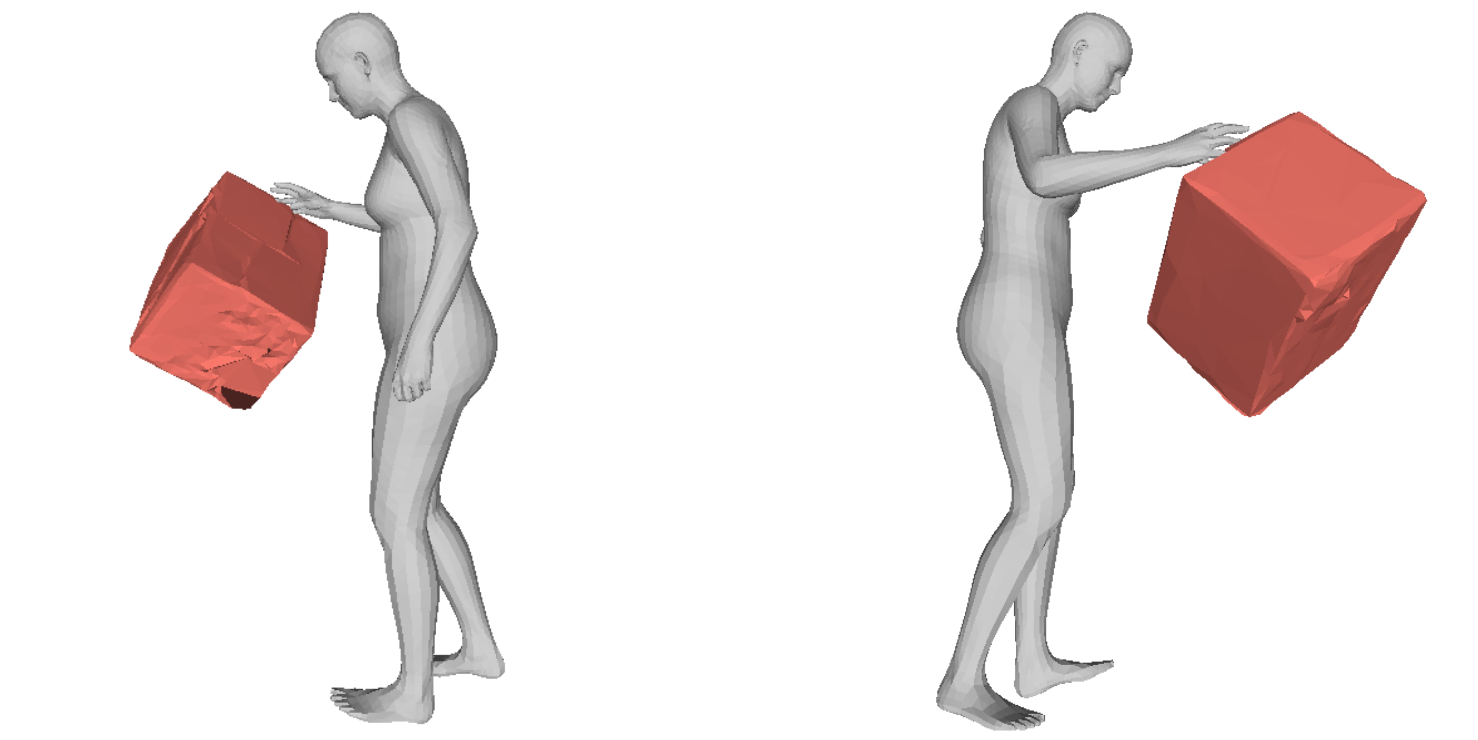} & \includegraphics[width=0.1\textwidth]{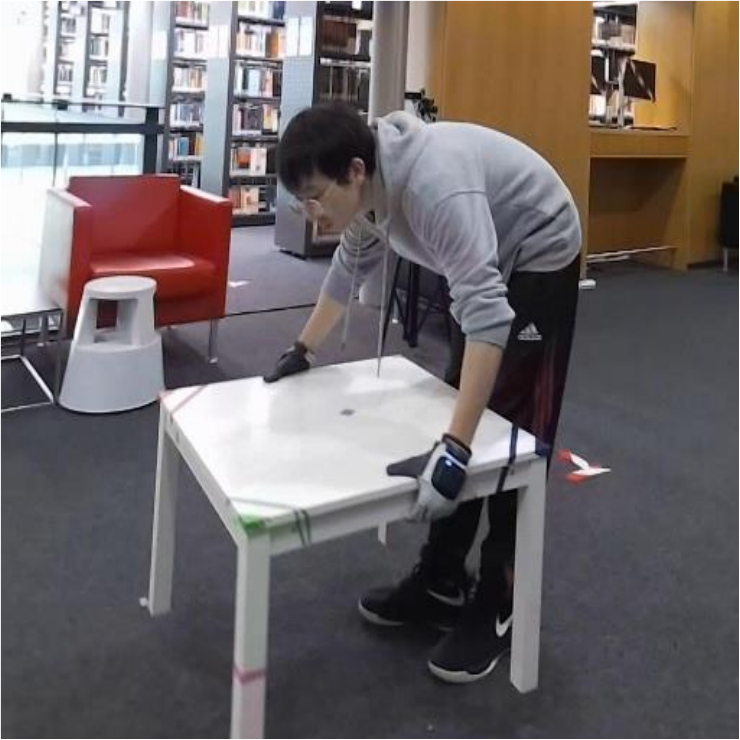} & \includegraphics[width=0.2\textwidth]{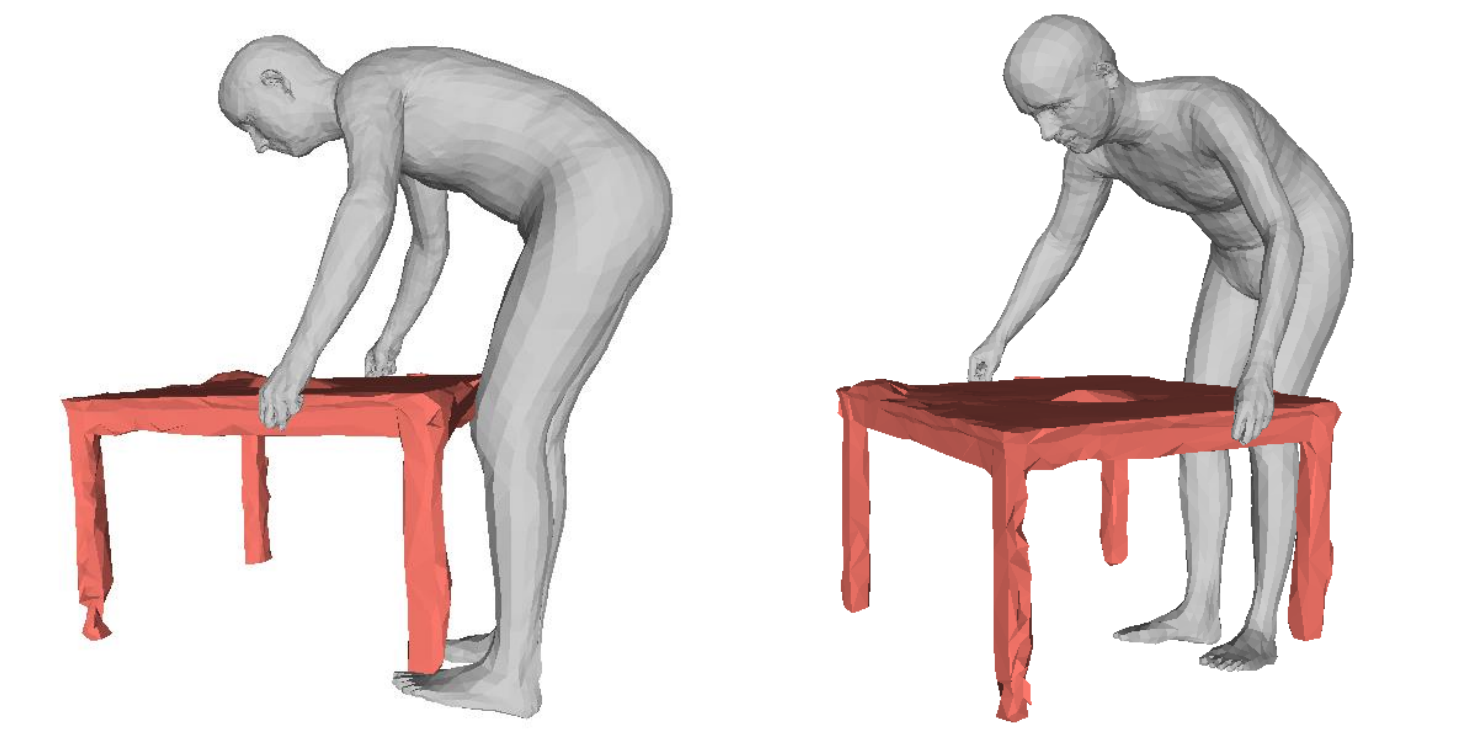} & \includegraphics[width=0.2\textwidth]{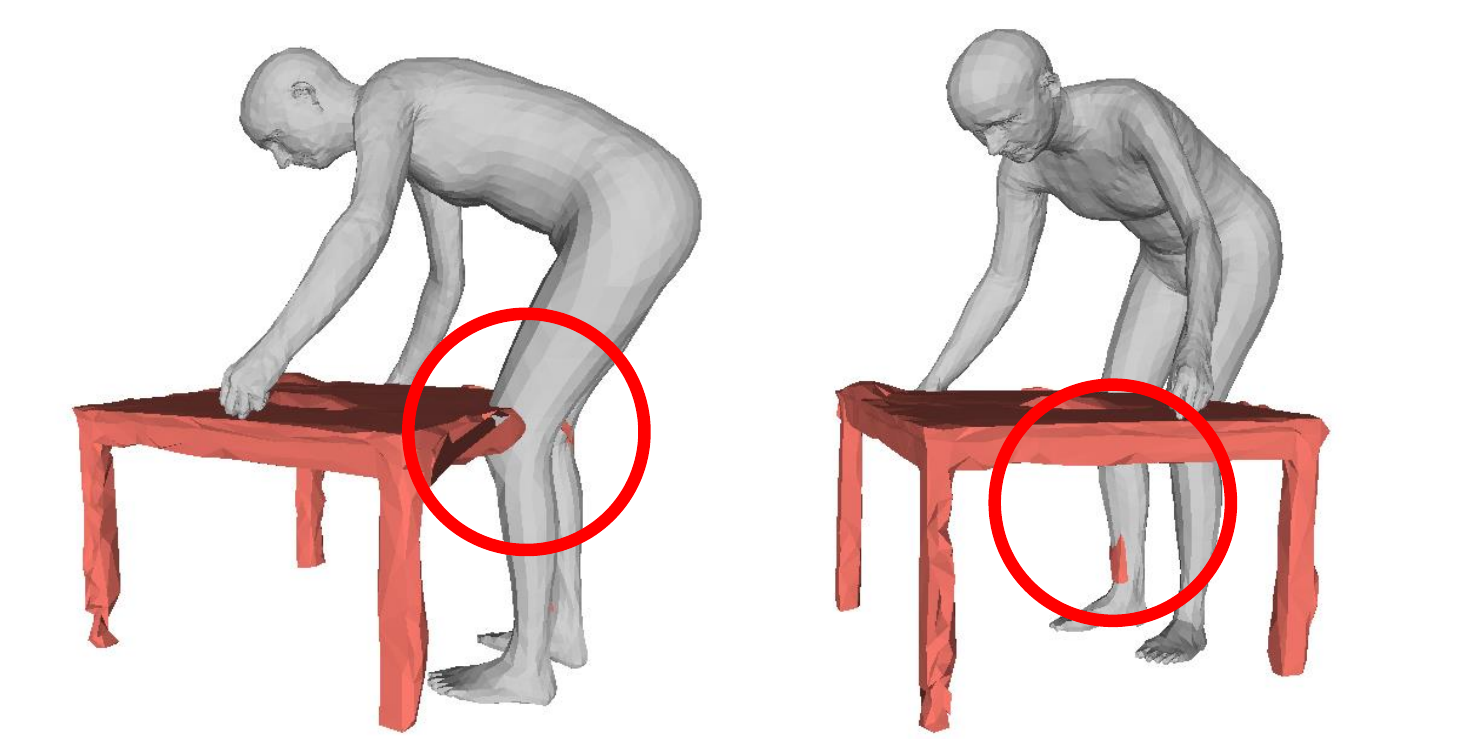} \\
     RGB &  HOI-TG (Ours) & CONTHO & RGB &  HOI-TG (Ours) &  CONTHO
    \end{tabular}
    \caption{Qualitative comparison of 3D human and object reconstruction with CONTHO \cite{joint} on BEHAVE \cite{behave}. Our HOI-TG achieves higher accuracy regarding the relative poses between the human and the object while also reducing instances of mesh penetration.}
    \label{fig:limitation1}
\end{figure*}

\subsection{Loss functions} \label{sec:loss}
Our proposed HOI-TG framework is trained in an end-to-end manner by minimizing the loss function \(\mathcal{L}\) defined as follows:
\begin{equation}
    \mathcal{L} = \mathcal{L}_{\textrm{human}} + \mathcal{L}_{\textrm{object}} + \mathcal{L}_{\textrm{hbox}},
    \label{Loss}
\end{equation}
where \(\mathcal{L}_{\textrm{human}}\) focuses on the reconstruction results of humans, \(\mathcal{L}_{\textrm{object}}\) targets the reconstruction results of objects. We designed \(\mathcal{L}_{\textrm{hbox}}\) loss function following CONTHO \cite{joint} and Hand4Whole \cite{hand4whole}. \(\mathcal{L}_{\textrm{hbox}}\) is the L1 distance between the predicted and ground truth (GT) bounding boxes of the hands. 

The term \(\mathcal{L}_{\textrm{human}}\) is defined as:
\begin{equation}
    \mathcal{L}_{\textrm{human}} = \mathcal{L}_{\textrm{human}}^{\textrm{ms-vertex}} + \mathcal{L}_{\textrm{human}}^{\textrm{param}} + \mathcal{L}_{\textrm{joint}} + \mathcal{L}_{\textrm{edge}},
    \label{Loss_human}
\end{equation}
where \(\mathcal{L}_{\textrm{joint}}\) represents the L1 distance between the predicted and GT human joint coordinates, encompassing both initial and refined 3D and 2D coordinates. The term \(\mathcal{L}_{\textrm{edge}}\) denotes the edge length consistency loss between the predicted and GT edges of the refined human meshes \(\mathbf{M}_h\). Additionally, \(\mathcal{L}_{\textrm{human}}^{\textrm{param}}\) quantifies the L1 distance between the predicted initial SMPLH parameters (\(\theta_{body}\) and \(\theta_{hand}\)) and their GT counterparts. 

\(\mathcal{L}_{\textrm{human}}^{\textrm{ms-vertex}}\) refers to the human multi-scale vertex loss. It involves a two-stage upsampling process: first, we upsample the predicted coarse mesh (431 vertices) to an intermediate resolution (1723 vertices), followed by further upsampling to the original mesh (6890 vertices for the SMPLH human mesh). The loss \(\mathcal{L}_{\textrm{human}}^{\textrm{ms-vertex}}\) applies a three-scale vertex loss between the predicted mesh and the ground truth.

The term \(\mathcal{L}_{\textrm{object}}\) is defined as:
\begin{equation}
    \mathcal{L}_{\textrm{object}} = \mathcal{L}_{\textrm{object}}^{\textrm{vertex}} + \mathcal{L}_{\textrm{object}}^{\textrm{param}},
    \label{Loss_object}
\end{equation}
where \(\mathcal{L}_{\textrm{object}}^{\textrm{vertex}}\) represents the L1 distance between the predicted and GT per-vertex 3D coordinates of the refined object meshes \(\mathbf{M}_o\), and \(\mathcal{L}_{\textrm{object}}^{\textrm{param}}\) quantifies the L1 distance between the predicted and GT 3D object rotation \(\mathbf{R}_o\) and translation \(\mathbf{t}_o\).

\section{Experiments}
\subsection{Datasets}
We conduct experiments on the BEHAVE~\cite{behave} and the InterCap~\cite{intercap} datasets. BEHAVE is an indoor dataset that captures seven subjects interacting with 20 diverse objects using a multi-view camera capture system. We follow CHORE~\cite{chore} and CONTHO \cite{joint} for the split of BEHAVE for a fair comparison. InterCap is another indoor human-object interaction dataset containing ten human subjects with ten different objects. We split the dataset following prior works \cite{vistracker,joint}.

\subsection{Evaluation metrics}
\textbf{Chamfer distance (CD$_\textrm{human}$, CD$_\textrm{object}$).} Following previous works \cite{vistracker,chore}, we evaluate human and object reconstruction using Chamfer distance between predicted and GT meshes. With the aligned 3D human and object meshes, we measure the Chamfer distance from GT separately on 3D human SMPL and 3D object in centimeters.

\noindent\textbf{Contact quality (Contact$_\textrm{p}$, Contact$_\textrm{r}$).} In addition to employing chamfer distance to evaluate the quality of human and object reconstruction, we also assess the accuracy of reconstructed contact predictions to measure human and object interaction quality. We follow \cite{joint} to generate the contact map by identifying human vertices that are within 5 cm of the object mesh. Then, we calculate precision and recall by comparing the human contact map with the ground truth.




\setlength{\tabcolsep}{2pt}
\begin{figure*}
    \centering
    \begin{tabular}{cccccc}
    \includegraphics[width=0.16\textwidth]{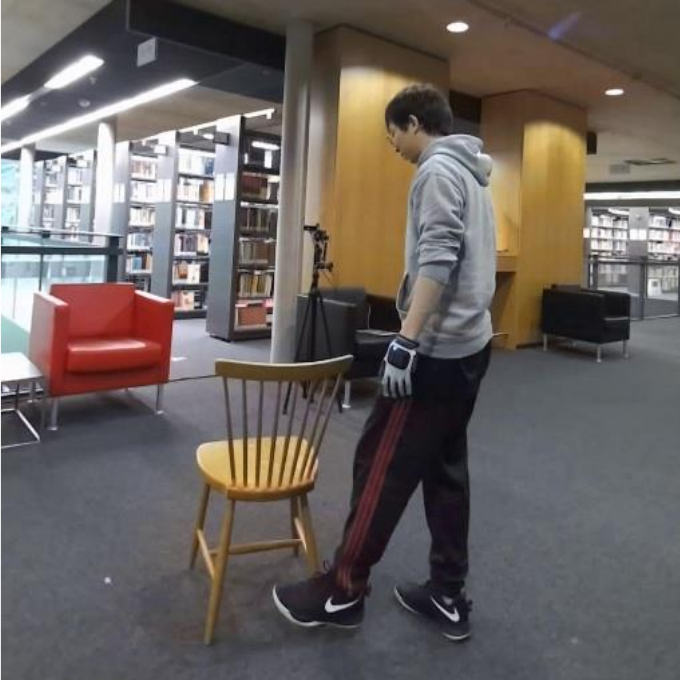}     & \includegraphics[width=0.16\textwidth]{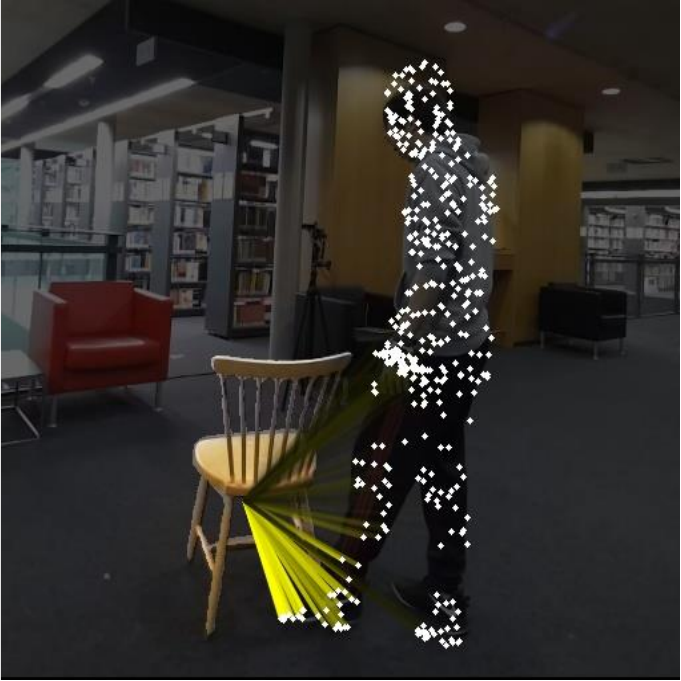} & \includegraphics[width=0.16\textwidth]{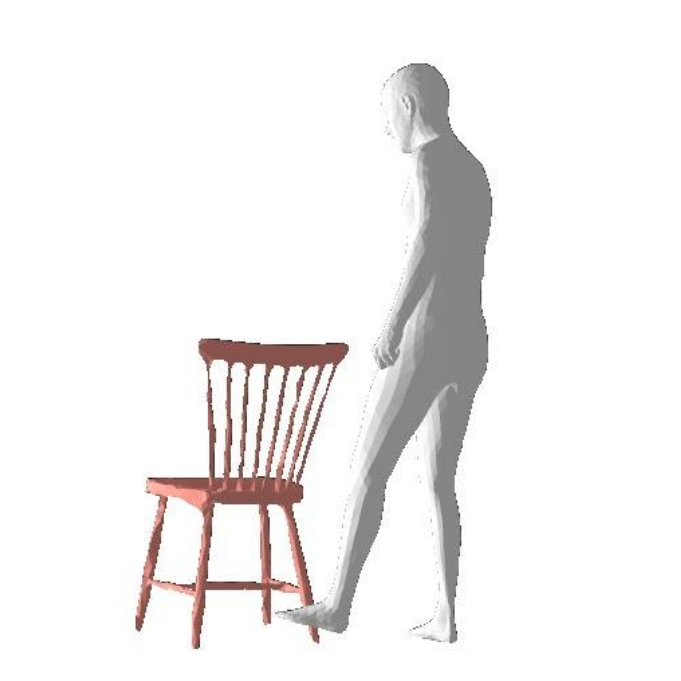} & \includegraphics[width=0.16\textwidth]{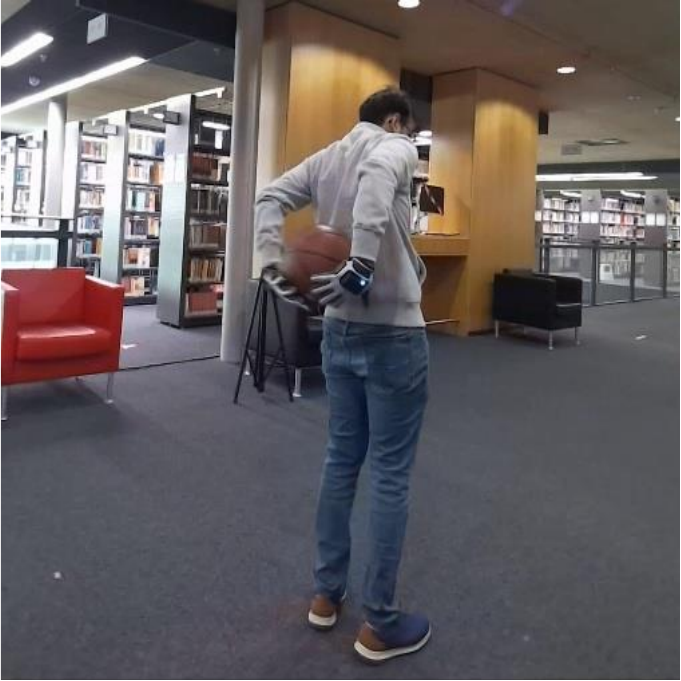} & \includegraphics[width=0.16\textwidth]{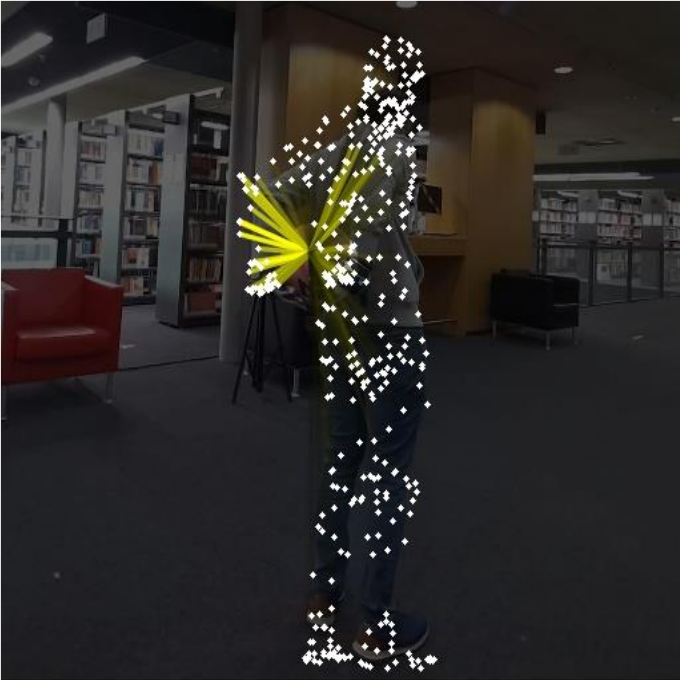} & \includegraphics[width=0.16\textwidth]{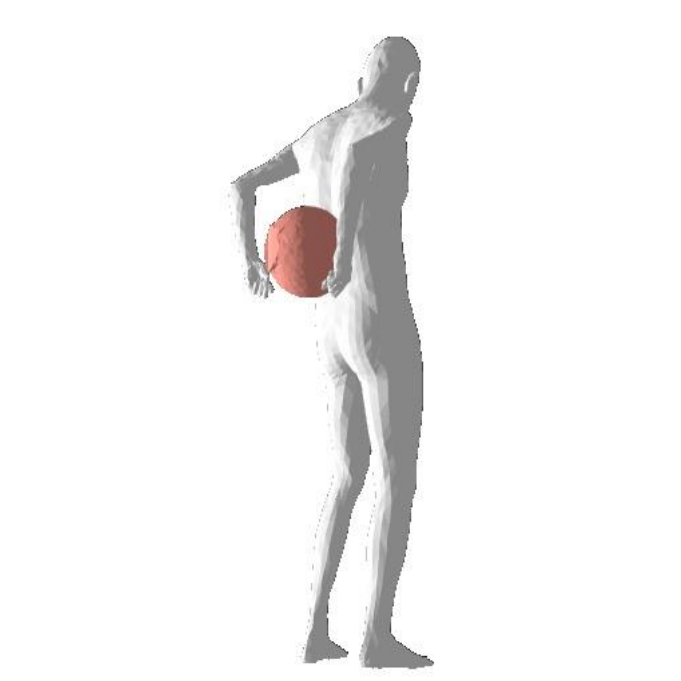} \\
    \includegraphics[width=0.16\textwidth]{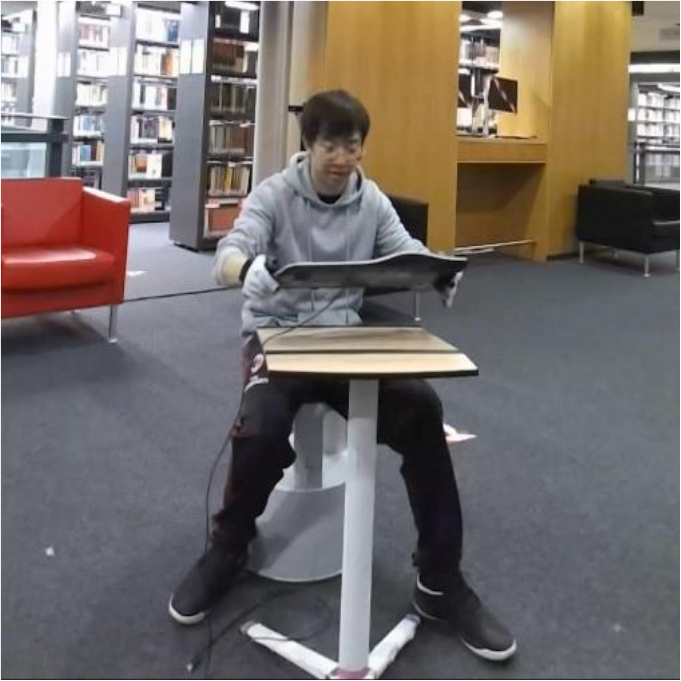}     & \includegraphics[width=0.16\textwidth]{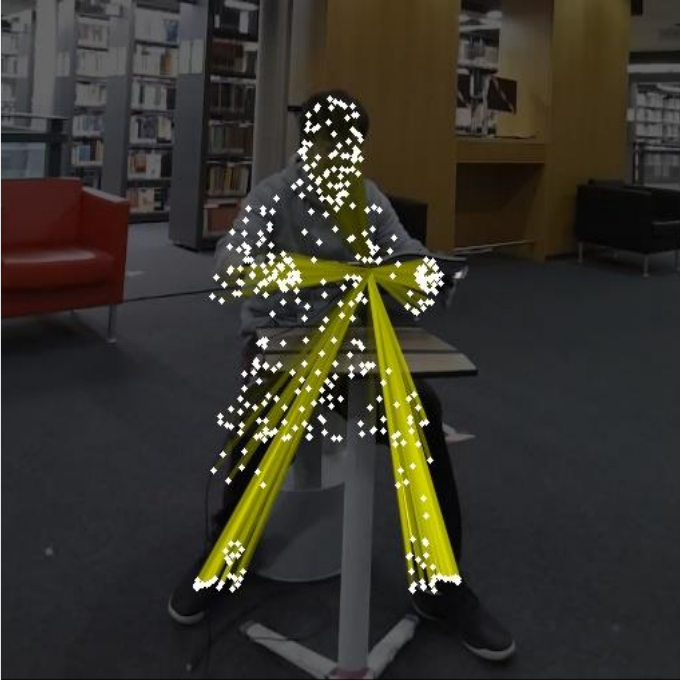} & \includegraphics[width=0.16\textwidth]{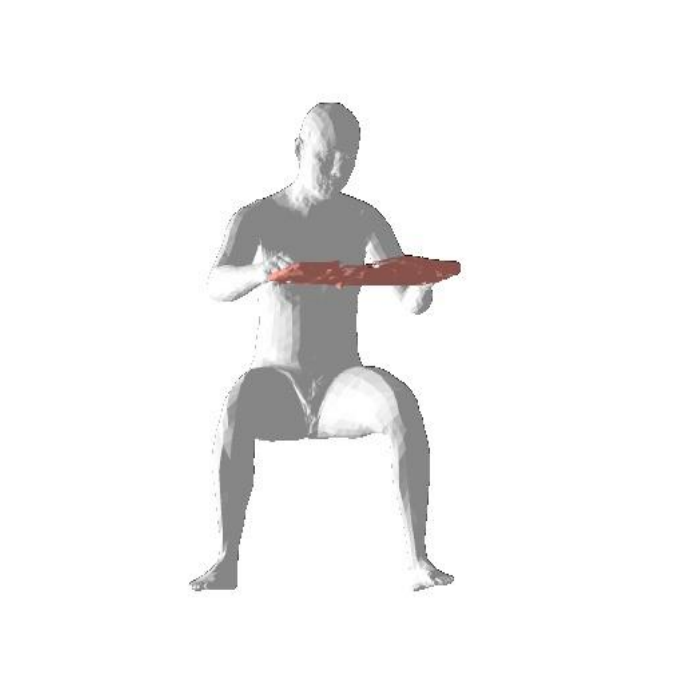} & \includegraphics[width=0.16\textwidth]{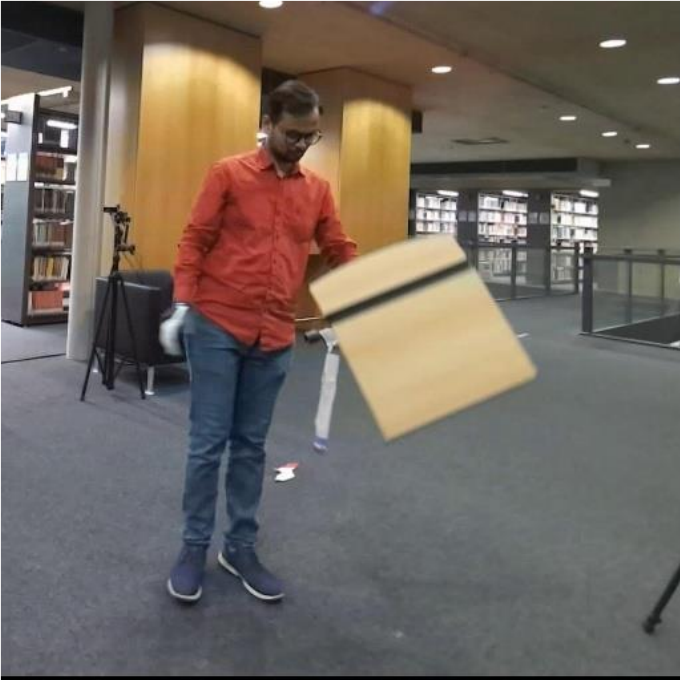} & \includegraphics[width=0.16\textwidth]{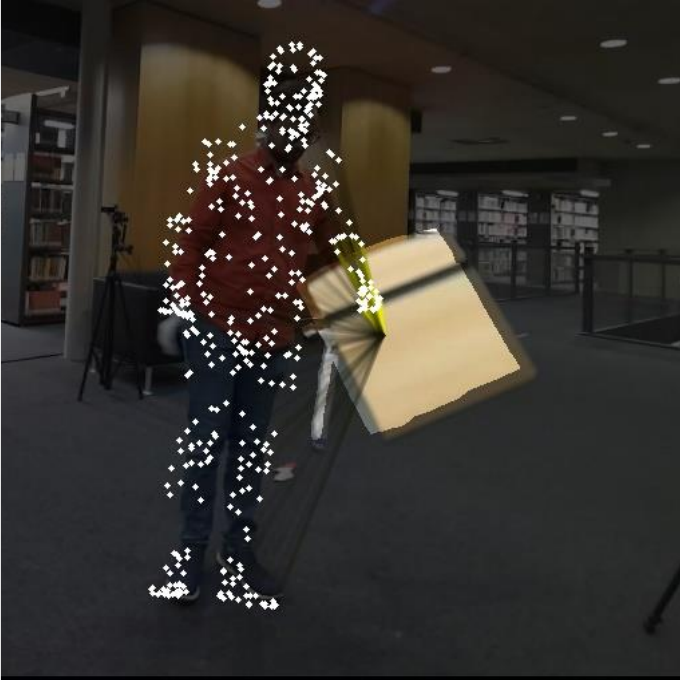} & \includegraphics[width=0.16\textwidth]{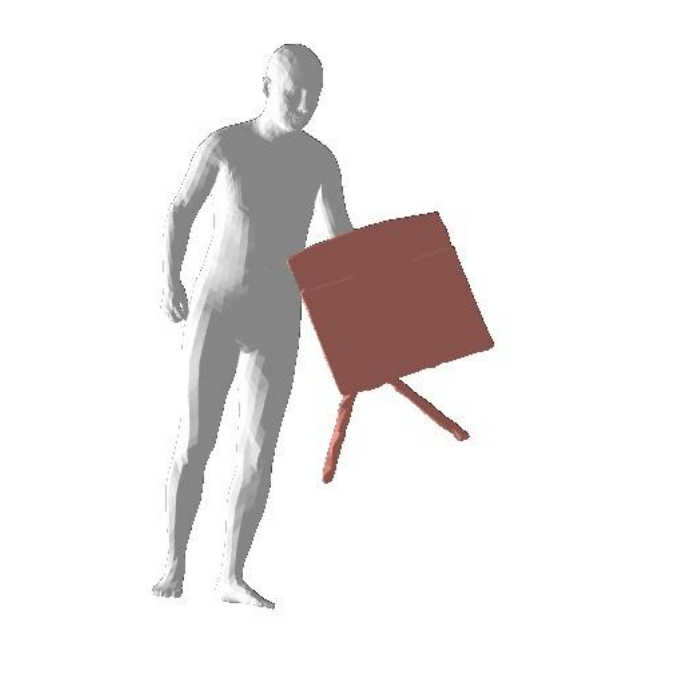} \\
    \includegraphics[width=0.16\textwidth]{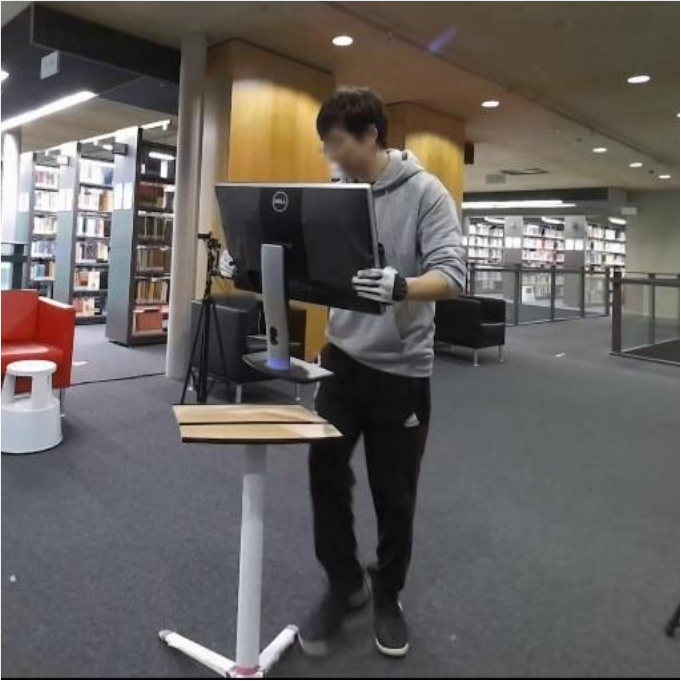}     & \includegraphics[width=0.16\textwidth]{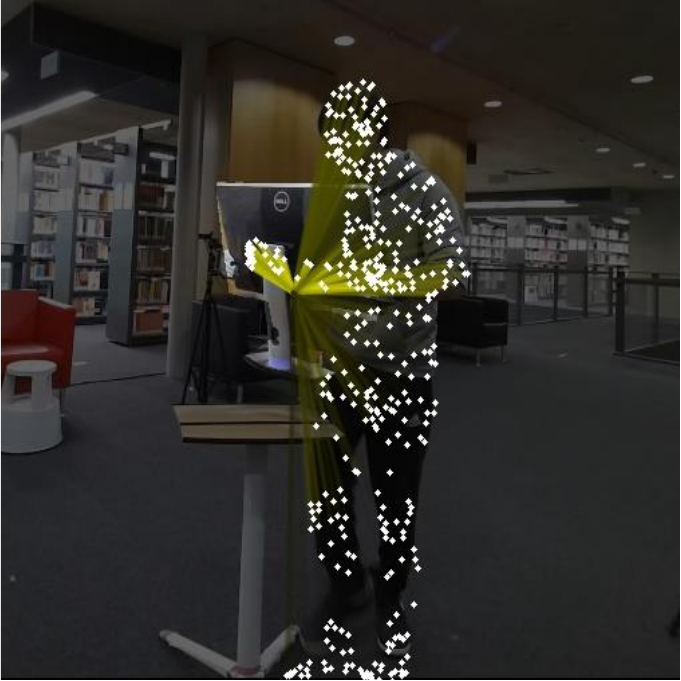} & \includegraphics[width=0.16\textwidth]{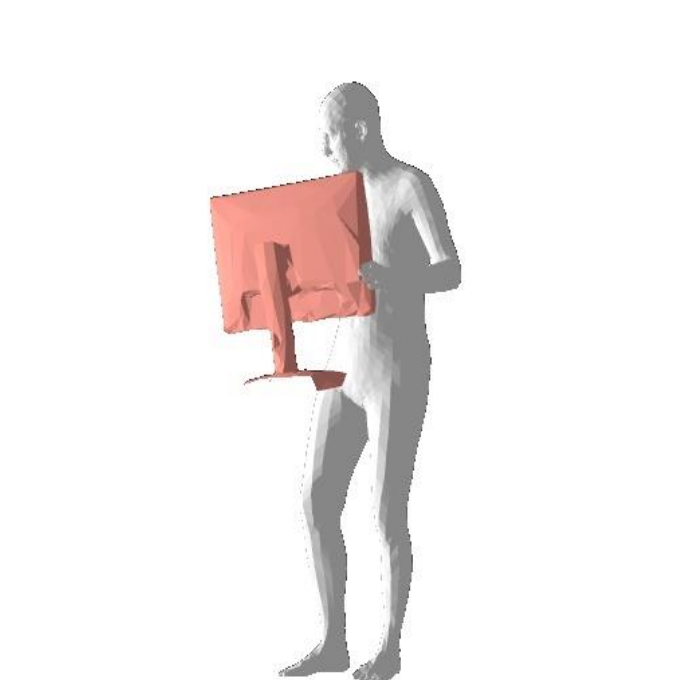} & \includegraphics[width=0.16\textwidth]{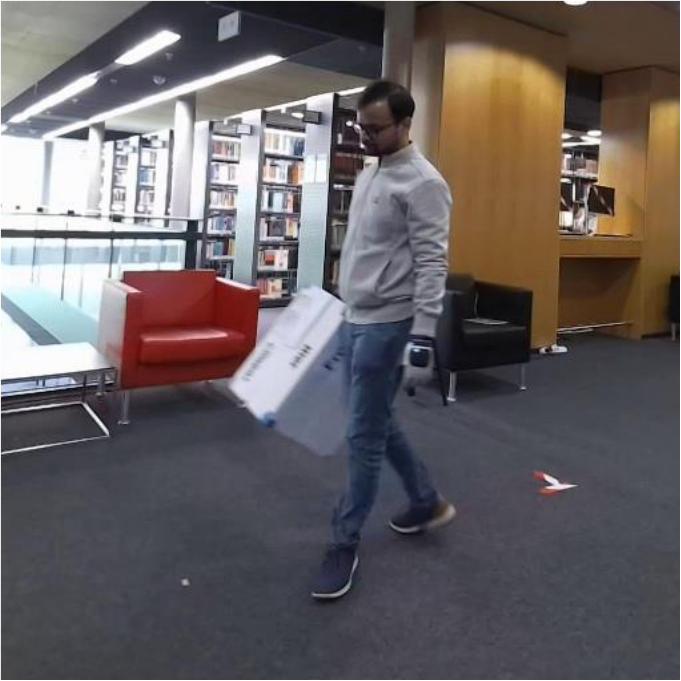} & \includegraphics[width=0.16\textwidth]{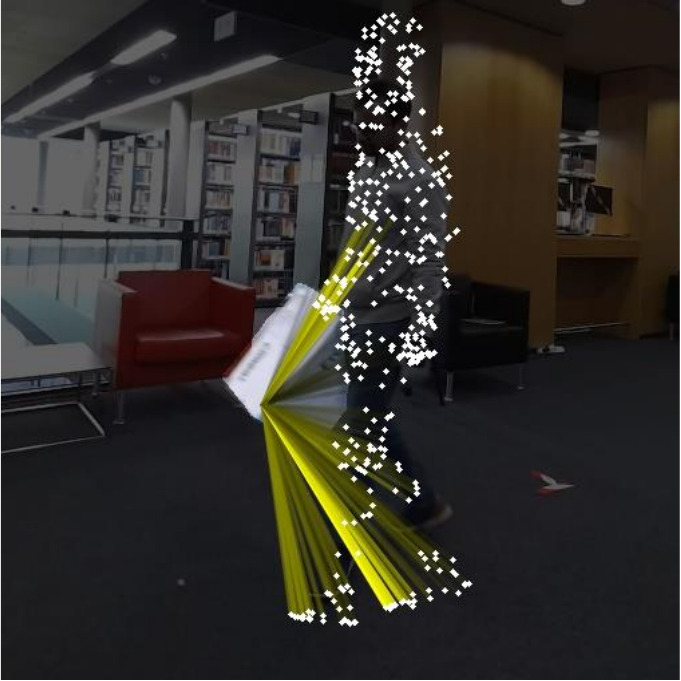} & \includegraphics[width=0.16\textwidth]{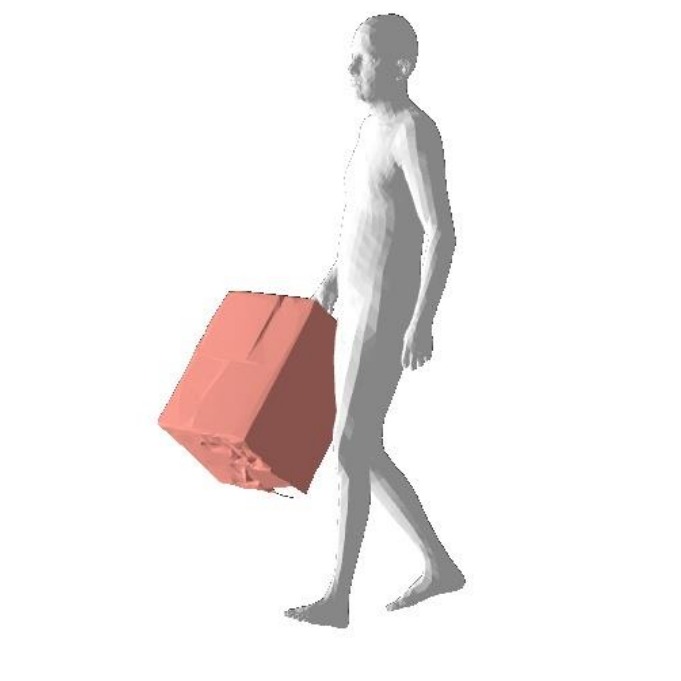} \\
     RGB &  HOI att. & HOI recon. & RGB &  HOI att. &  HOI recon. 
    \end{tabular}
    \caption{Visualization of the attention distribution (HOI att.) between human mesh vertices and the object and corresponding reconstruction results (HOI recon.) from our HOI-TG. The brighter color indicates more intensive attention.
    }
    \label{fig:attention}
\end{figure*}

\subsection{Comparison with state-of-the-art methods}
We compare our method with previous approaches on BEHAVE \cite{behave} and InterCap \cite{intercap} datasets.
The counterparts include a classical optimization-based method PHOSA \cite{PHOSA}, a representative implicit neural field-based model CHORE~\cite{chore}, 
and the latest 3D vertex refinement-based model CONTHO~\cite{joint}. 
According to \cref{tab:comparison}, our method outperforms the previous state-of-the-art on both datasets. 
The results of PHOSA \cite{PHOSA} on BEHAVE are not satisfactory, particularly regarding the prediction of reconstructed contact outcomes. It indicates the shortcoming of reliance on pre-set physical priors. CHORE \cite{chore} improves significantly over PHOSA, which shows the advantage of jointly modeling humans and objects via neural fields. CONTHO \cite{joint} outperforms the previous two approaches by devising an initialization-refinement pipeline.

Compared to CONTHO~\cite{joint}, our HOI-TG improves the reconstruction of humans and objects by 8.0\% and 5.0\%, respectively. 
For the reconstructed contact, our method promotes precision and recall by 3.4\% and 5.8\%, respectively. The results demonstrate a better reconstruction of global meshes and a higher quality of the contact areas. Since HOI-TG and CONTHO start from the same initial meshes, such improvements directly show the effectiveness of our straightforward transformer encoder and graph convolutional structures that implicitly learn the interactions between humans and objects. 

The InterCap dataset contains complex topological structures (\eg, umbrellas and skateboards) and is, therefore, more challenging. As a result, the performance of PHOSA and CHORE deteriorates heavily. It reveals the limitation of physical priors and unsigned neural fields.
In contrast, vertex regression-based methods (\ie, CHONTHO and our HOI-TG) achieve relatively stabler performance.
Our method improves 3D mesh reconstruction by 8.9\% and 8.6\% for humans and objects, respectively. Meanwhile, it increases precision and recall by 3.9\% and 4.1\% for reconstructed contact accuracy. These results highlight the advantages of HOI-TG in implicitly modeling both global posture and local interactions simultaneously.

Examples in the left half of \cref{fig:limitation1} illustrate that, for relatively symmetrical graph structures, the previous state-of-the-art method CONTHO~\cite{joint} fails to address the topological variations among different objects. Such deficiency prevents accurate pose predictions for those symmetrical entities. We introduced an object graph residual block to enhance the modeling of associations between object vertices. Along with the self-attention mechanism, the block allows our HOI-TG to model confusing symmetrical objects better and predict their poses more accurately.
As demonstrated in the right half of \cref{fig:limitation1}, our model performs well in reconstructing complex human-object interactions. CONTHO~\cite{joint} often struggles in identifying the relative positioning of humans and objects. It is likely due to the explicit constraints imposed on interactions, which can lead to mesh penetration. Benefiting from implicit interaction modeling, our HOI-TG produces a more robust human-object interaction reconstruction. 



\subsection{Visualization and ablation study}
To thoroughly analyze HOI-TG, we conduct essential visualization to unveil the attention it learns and a series of ablation studies on the BEHAVE dataset. Please refer to the supplementary material for more results. 


\textbf{Human and object interactions:} To further understand the effect of multi-layer HOI transformer encoder in learning interactions, we visualize the attention distribution (HOI att.) among human and object mesh vertices and also the reconstruction results (HOI recon.) in
\cref{fig:attention}. 
We use the average of attention values from each human vertex to all object vertices as the attention value of that human vertex towards the object. Specifically, the dimension of the attention matrix between human and object is $431{\times} 64$. The average operation produces a $431$-dim vector for visualization. The brighter color indicates more intensive attention. For simple and static interactions, the positions of objects only relate to local body parts. For instance, the positions of a chair, table, monitor, and basketball solely depend on the locations of the interacting body parts. In contrast, for complex interactions, such as sitting on a chair while using a keyboard or moving with a suitcase, our model successfully attends to non-local body parts and leverages non-local vertices to predict the positions of the objects.

\setlength{\tabcolsep}{5pt}
\begin{table}[h]
\centering
\begin{tabular}{lcccc}
\hline
        & CD$_\textrm{human}$$\downarrow$   & CD$_\textrm{object}$$\downarrow$   & Contact$_\textrm{p}$$\uparrow$    & Contact$_\textrm{r}$$\uparrow$     \\ \hline
\textit{Static} & 4.95     & 8.90     & 0.632      & 0.472      \\
\textit{Initial}  & \textbf{4.59} & \textbf{8.00} & \textbf{0.662} & \textbf{0.554} \\ \hline
\end{tabular}
\caption{Ablation study of 3D query features.}
\label{tab:input}
\end{table}

\textbf{Representations of 3D query features:} We study the behavior of our HOI-TG architecture when using two different 3D query features as inputs. 1) We utilize the global pooling of image features $\mathbf{F}$ along with static template coordinates (\textit{Static}). 2) We employ the initial vertices for 2D projection, performing grid sampling of features in $\mathbf{F}$ and concatenate them with the initially estimated coordinates (\textit{Initial}), which is our design in \cref{sec:input}. As shown in \cref{tab:input}, using the initial estimation significantly outperforms the \textit{Static} variant. We consider two reasons for the phenomenon: 1) The effectiveness of 3D models relies on the ability to recognize and differentiate individual points belonging to different entities. If all points share the same descriptive features, their uniqueness in 3D space becomes obscured. 2) Interactions cover diverse poses between humans and objects, even for the same object. Learning such interactions directly from static templates becomes challenging because these templates do not capture dynamic variations or relative motions essential for understanding interactions.

\begin{table}[h]
\centering
\begin{tabular}{lcccc}
\hline
Pipeline                      & \small{CD$_\textrm{human}$$\downarrow$}   & \small{CD$_\textrm{object}$$\downarrow$}   & \small{Contact$_\textrm{p}$$\uparrow$}    & \small{Contact$_\textrm{r}$$\uparrow$}         \\ \hline
Transformer                   & 4.73          & 8.55 & 0.606 & \textbf{0.559}      \\
\hspace{0.2cm}+H\_g          & 4.61          & 8.11 & 0.651 & 0.539      \\
\hspace{0.2cm}+H\_g+O\_g & \textbf{4.59} & \textbf{8.00} & \textbf{0.662} & 0.554 \\ \hline
\end{tabular}
\caption{Ablation study of graph residual blocks.}
\label{tab:graph}
\end{table}

\textbf{Graph residual blocks:} We investigate the impact of Human Graph Residual Block (H\_g) and Object Graph Residual Block (O\_g) in \cref{tab:graph}. The first row corresponds to the transformer architecture that uses no graph convolution in the network. The second row shows the results of adding a human graph convolution layer to every encoder block, and the last row lists the results of adding both human and object graph convolution layers. We can conclude that: 1) H\_g improves the reconstruction of humans and objects for both global meshes and local contact. It indicates that relying solely on transformers to model the interactions between them can cause ambiguity in independent topological structures. 
2) Adding O\_g on top of H\_g further promotes the global reconstruction of humans and further increases the reconstruction accuracy of the contact area. Overall, the design of HOI-TG that involves H\_g and O\_g achieves good results for HOI reconstruction.



\begin{table}[h]
\centering
\begin{tabular}{cccccc}
\hline
JR & MS & \small{CD$_\textrm{human}$$\downarrow$}   & \small{CD$_\textrm{object}$$\downarrow$}   & \small{Contact$_\textrm{p}$$\uparrow$}    & \small{Contact$_\textrm{r}$$\uparrow$}      \\ \hline
$\times$                 & $\times$                 & 5.01          & 8.22  & 0.612          & 0.559                 \\
\checkmark                 & $\times$                & 4.76          & 8.30    & 0.631          & 0.563                 \\
$\times$                 & \checkmark                 & 4.72          & 8.29  & 0.626          & 0.553                \\
\checkmark                 & \checkmark                 & \textbf{4.62} & \textbf{8.05} & \textbf{0.644} & \textbf{0.573} \\ \hline
\end{tabular}
\caption{Ablation study of \(\mathcal{L}_{\textrm{joint}}^\textrm{refine}\) (JR) and \(\mathcal{L}_{\textrm{human}}^{\textrm{ms-vertex}}\) (MS).}
\label{tab:loss}
\end{table}

\textbf{Loss functions:} Compared with CONTHO~\cite{joint}, we use 1) a different $\mathcal{L}_{\textrm{joint}}$ containing an extra refinement of joints, which we denote as \(\mathcal{L}_{\textrm{joint}}^\textrm{refine}\) (JR), and 2) a new \(\mathcal{L}_{\textrm{human}}^{\textrm{ms-vertex}}\) (MS) to speed up convergence by constraining the quality of human reconstruction at different scales. Therefore, we investigate the impact of \(\mathcal{L}_{\textrm{joint}}^\textrm{refine}\) and \(\mathcal{L}_{\textrm{human}}^{\textrm{ms-vertex}}\) in \cref{tab:loss}. In \cref{tab:loss}, the proposed graph residual block is added in all encoders, while in \cref{tab:comparison}, we only add the Object Graph Residual Block in the second transformer encoder module. The results indicate that adding one of them alone slightly hurts the reconstruction of objects, yet combining them improves global and local reconstruction for both humans and objects.


\begin{figure}[h]
    \centering
    \begin{subfigure}{0.23\textwidth}
        \includegraphics[width=\linewidth]{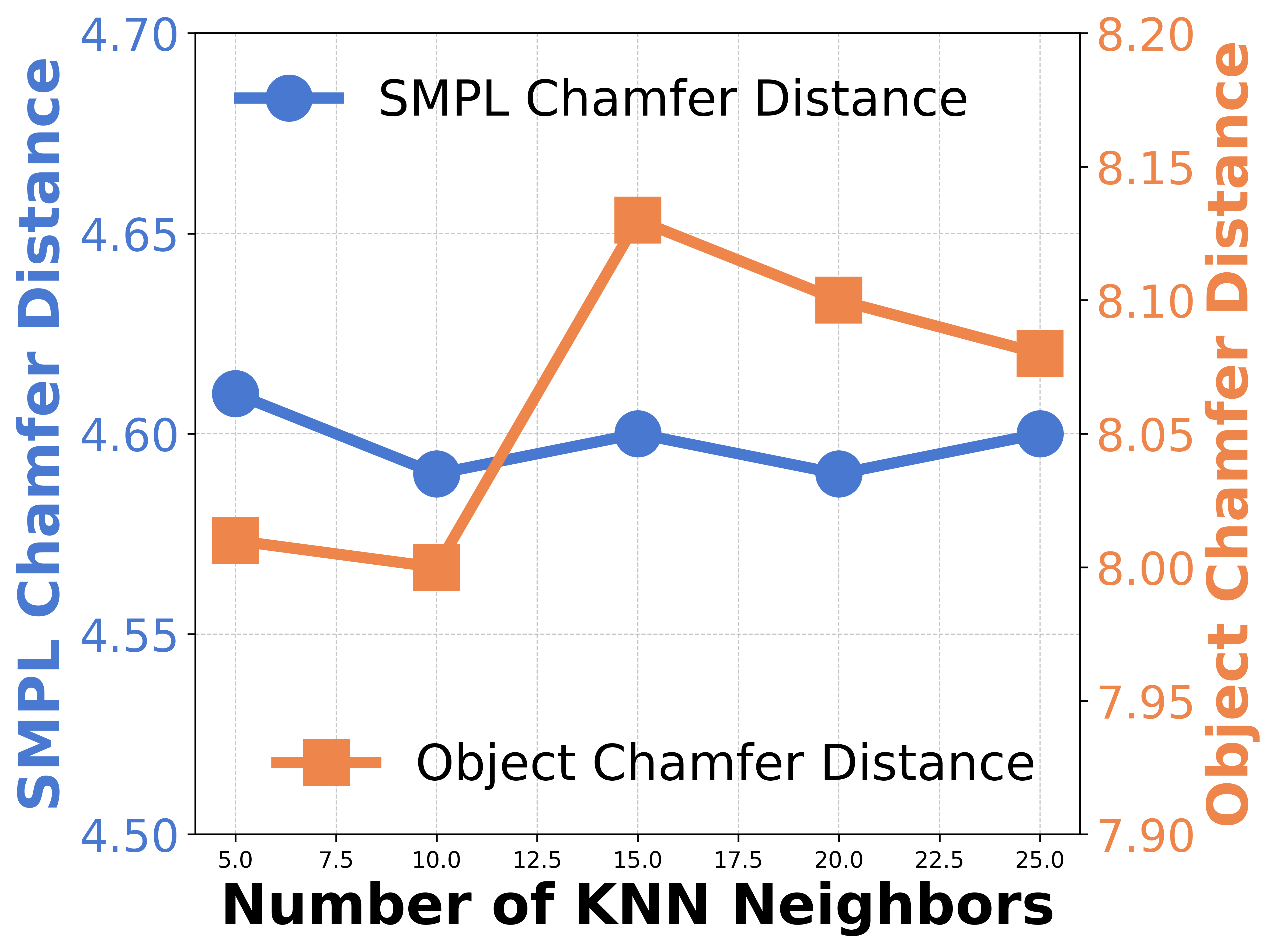}
        \caption{Reconstruction distance.}
        \label{fig:KNNCD}
    \end{subfigure}
    \hfill
    \begin{subfigure}{0.23\textwidth}
        \includegraphics[width=\linewidth]{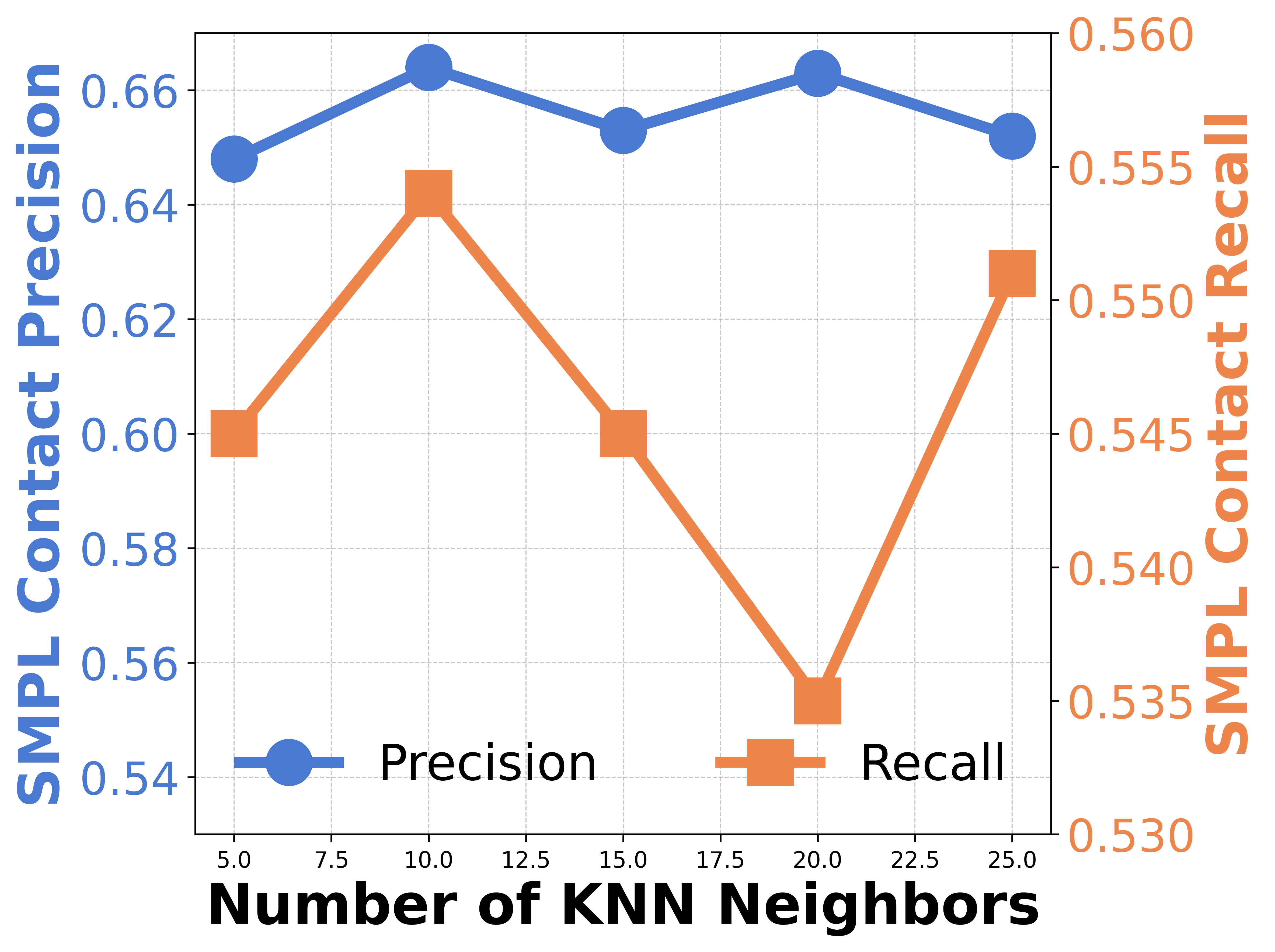}
        \caption{Contact precision \& recall}
        \label{fig:KNNcontact}
    \end{subfigure}
    \caption{Ablation study of numbers of KNN neighbors.}
    \label{fig:KNN}
\end{figure}

\textbf{KNN neighbors:} In \cref{fig:KNN}, we investigated the optimal number of KNN neighbors in the object graph residual block while constructing the adjacency matrix. Our results indicate that setting the number of neighbors to 10 leads to good performance in both the reconstructed Chamfer distance and the accuracy of contact predictions. We analyze that when a low number of neighbors is used, the model struggles to represent the adjacency relationships between nodes. Conversely, employing a high number of neighbors may result in the incorporation of redundant feature information from non-local adjacent nodes, thus obscuring the model's ability to capture these relationships accurately.

\section{Conclusion}
We have developed an end-to-end HOI Reconstruction Transformer with Graph-based Encoding, namely HOI-TG, for 3D human and object mesh reconstruction from a single image. By leveraging transformer architecture to implicitly model interactions and integrating graph convolutional networks for vertex interactions, our approach enhances the capture of both global and local interactions. This innovative methodology has demonstrated state-of-the-art performance on the BEHAVE and the InterCap datasets. The \textbf{limitations} of HOI-TG lie in dealing with lying poses and purely symmetric objects. A detailed analysis is provided in the supplementary material.

{
    \small
    \bibliographystyle{ieeenat_fullname}
    \bibliography{main}
}

\clearpage
\setcounter{page}{1}
\maketitlesupplementary
\newcommand{\rgbwidth}{0.09}
\newcommand{\meshwidth}{0.14}

%
This supplementary material elaborates on the proposed HOI-TG regarding methodology details and experimental results. Such information includes grid sampling and positional encoding, the construction of graph adjacency weight matrix for objects, and expanding loss functions. The results cover the comparison with StackFLOW~\cite{stackflow}, the variants of removing the annotated segmentation masks and a replacement by DetectronV2~\cite{wu2019detectron2} masks, and the location of graph residual blocks. Then, we comprehensively compare the computational efficiency and discuss the limitations of our HOI-TG. To further validate the effectiveness of our method, we present more qualitative comparisons in the end. Along with this material, we also provide the code for reproducibility.

\section{More details of our HOI-TG}
\subsection {Grid sampling and positional encoding}
We introduce grid sampling and positional encoding to provide the encoder with more 3D informative input. First, we project the human joints $\mathbf{M}_j^\textrm{init}$, human mesh $\mathbf{M}_h^\textrm{init}$ and object mesh $\mathbf{M}_o^\textrm{init}$ obtained in the init stage onto the 2D image plane using the camera parameters predicted in the init stage. Then, we apply bilinear interpolation on $\mathbf{F}$ to obtain the feature corresponding to each vertex. Finally, we concatenate the interpolated feature and the 3D coordinates of each vertex to produce our 3D queries.

\subsection{Graph adjacency matrix for objects}
For the adjacency matrix of objects with different templates, assuming that the 3D coordinate point of an object template is $P \in \mathbb{R}^{n\times3}$, we first calculate the distance $d(p_i, p_j)$ between each two points,
\begin{equation}
d(p_i, p_j) = \|p_i-p_j\|_2
\end{equation}
Then for each point, we select the $K$ points with the closest distance as its neighbors and calculate the un-normalized adjacency matrix $\bar{\mathbf{A}}^{'}$,
\begin{equation}
\bar{\mathbf{A}}^{'}(i,j) = 
\begin{cases} 
 {d(p_i, p_j)}& \text{if } p_j \text{ is one of the } K{\text{ neighbors}} \text{ of } p_i, \\
0 & \text{otherwise.}
\end{cases}
\end{equation}
$K$ represents the pre-set number of neighbors. We use the distance as the weight for neighbor nodes and 0 for non-neighbor nodes. Finally, $\bar{\mathbf{A}}^{'}$ is normalized to obtain the adjacency matrix $\bar{\mathbf{A}}$.

\subsection{Loss functions and implementation details.}
Our loss function consists of three parts, which are $\mathcal{L}_{\textrm{human}}$, $\mathcal{L}_{\textrm{hbox}}$ focusing on human reconstruction, and $\mathcal{L}_{\textrm{object}}$ focusing on object reconstruction:
\begin{equation}
    \mathcal{L} = \mathcal{L}_{\textrm{human}} + \mathcal{L}_{\textrm{object}} + \mathcal{L}_{\textrm{hbox}}.
    \label{suppLoss}
\end{equation}
The $\mathcal{L}_{\textrm{hbox}}$ represents the L1 loss of the predicted hand bounding box and GT. We follow previous works \cite{joint,hand4whole} for the design.

$\mathcal{L}_{\textrm{human}}$ is defined as:
\begin{equation}
    \mathcal{L}_{\textrm{human}} = \mathcal{L}_{\textrm{human}}^{\textrm{ms-vertex}} + \mathcal{L}_{\textrm{human}}^{\textrm{param}} + \mathcal{L}_{\textrm{joint}} + \mathcal{L}_{\textrm{edge}}.
    \label{suppLoss_human}
\end{equation}
\textbf{Human vertex multi-scale loss $\mathcal{L}_{\textrm{human}}^{\textrm{ms-vertex}}$:} We upsample $\mathbf{M}_h \in \mathbb{R}^{431 \times 3}$ twice to get $\mathbf{M}_h^{*} \in \mathbb{R}^{1723 \times 3}$ and $\mathbf{M}_h^{**} \in \mathbb{R}^{6890 \times 3}$. $\mathcal{L}_{\textrm{human}}^{\textrm{ms-vertex}}$ represents the L1 loss between the multi-scale human vertices($\mathbf{M}_h$, $\mathbf{M}_h^{*}$ and $\mathbf{M}_h^{**}$) and GT vertices.

\noindent\textbf{Human parameters loss $\mathcal{L}_{\textrm{human}}^{\textrm{param}}$:} $\mathcal{L}_{\textrm{human}}^{\textrm{param}}$ represents the L1 loss between the predicted parameters (human body mesh $\theta_{\textrm{body}}$ and human hand mesh $\theta_{\textrm{hand}}$) in the init stage and GT parameters. 

\noindent\textbf{Human joint loss $\mathcal{L}_{\textrm{joint}}$:} Our model has three joint outputs in total: i) the 3D joint predicted in the init stage, ii) the init 3D joint and init 2D joint obtained by the SMPLH model through the human parameters predicted in the init stage, iii) the 3D joint and 2D joint coordinates we reconstruct with the transformer. We all utilize L1 loss to minimize the loss between them and the corresponding GT.

\noindent\textbf{Human edge length consistency loss $\mathcal{L}_{\textrm{edge}}$:} $\mathcal{L}_{\textrm{edge}}$ is the L1 loss between up-sampled predicted human mesh $\mathbf{M}_h^{**}$ edges and GT edges.

$\mathcal{L}_{\textrm{object}}$ is defined as:
\begin{equation}
    \mathcal{L}_{\textrm{object}} = \mathcal{L}_{\textrm{object}}^{\textrm{vertex}} + \mathcal{L}_{\textrm{object}}^{\textrm{param}}.
    \label{suppLoss_object}
\end{equation}

\noindent\textbf{Object vertices loss $\mathcal{L}_{\textrm{object}}^{\textrm{vertex}}$:} $\mathcal{L}_{\textrm{object}}^{\textrm{vertex}}$ is the L1 loss between the reconstruction object vertices $\mathbf{M}_o$ and GT. 

\noindent\textbf{Object parameters loss $\mathcal{L}_{\textrm{object}}^{\textrm{param}}$:} $\mathcal{L}_{\textrm{object}}^{\textrm{param}}$ is the L1 loss between the init object parameters ($\mathbf{R}_{\textrm{init}}$ and $\mathbf{T}_{\textrm{init}}$) and GT.

\noindent\textbf{Implementation details.}
We train our HOI-TG framework using the Adam optimizer with an initial learning rate of $1{\times}10^{-4}$ for both the transformer and the ResNet50 backbone. The pipeline is trained for 50 epochs, with the learning rate decaying by 0.1 after 30 epochs. All transformer weights are randomly initialized, except that the ResNet backbone is initialized with weights from Hand4Whole~\cite{hand4whole}. We set the mini-batch size to 16 on an NVIDIA A100 80GB GPU.

\begin{table*}[ht]
\centering
\begin{tabular}{lcccc}
\hline
Method                 & CD$_\textrm{human}$$\downarrow$   & CD$_\textrm{object}$$\downarrow$   & Contact$_\textrm{p}$$\uparrow$    & Contact$_\textrm{r}$$\uparrow$              \\ \hline
StackFLOW (w/o post-optim.) & 5.98          & 12.6          & 0.429          & 0.521          \\
StackFLOW (w post-optim.)   & 6.27          & 11.5          & 0.465          & 0.538          \\
\textbf{Ours}          & \textbf{4.87} & \textbf{7.49} & \textbf{0.647} & \textbf{0.539} \\ \hline
\end{tabular}
\caption{Comparison with StackFLOW~\cite{stackflow} on the BEHAVE~\cite{behave} dataset. We use the officially released checkpoints by StackFLOW~\cite{stackflow} for the comparison.}
\label{tab:comparestack}
\end{table*}

\section{More experimental results}
\subsection{Comparison with StackFLOW}
Table \ref{tab:comparestack} presents a comparison between our HOI-TG and StackFLOW~\cite{stackflow}. StackFLOW infers the posterior distribution of spatial relationships between people and objects from the input image and utilizes GT offsets to optimize their positions and postures during the inference stage, namely post-optimization. Although StackFLOW provides experimental results on the BEHAVE dataset, it splits the dataset differently from other approaches~\cite{joint,chore,xie2024rhobin}. 
Specifically, StackFLOW samples more instances for testing within the BEHAVE dataset's test set. 
For a fair comparison, we report results on the intersection of its test split and the generally used test set. Our method outperforms StackFLOW, regardless of whether it includes the post-optimization stage. The results show the effectiveness of our implicit contact modeling over the explicit human-object offset constraint.

\subsection{Segmentation}
Following CONTHO \cite{joint}, we use human and object segmentations provided by the datasets as the inputs. For completeness, we investigate the necessity of the segmentation masks. As shown in \cref{tab:seg}, without segmentation, the reconstruction results for both humans and objects are subpar, particularly for objects. Our analysis reveals that this is primarily due to information interference from the background image. When the object's color closely matches the background, the model struggles to accurately discern the object's depth position, which significantly hinders human-object interaction (HOI) reconstruction. By incorporating segmentation, the model can more effectively identify the relative depth positions of humans and objects, resulting in improved reconstruction accuracy.

\setlength{\tabcolsep}{1pt}
\begin{table}[h!]
\centering
\begin{tabular}{lcccc}
\hline
        & \small{CD$_\textrm{human}$$\downarrow$}   & \small{CD$_\textrm{object}$$\downarrow$}   & \small{Contact$_\textrm{p}$$\uparrow$}    & \small{Contact$_\textrm{r}$$\uparrow$}     \\ \hline
CONTHO & 4.99     & 8.42     & 0.628      & 0.496      \\
CONTHO (w/o seg.) & 6.16     & 19.23     & 0.440     & 0.348      \\ \hline
Ours  & \textbf{4.59} & \textbf{8.00} & \textbf{0.662} & \textbf{0.554} \\
Ours (w/o seg.) & 5.67     & 19.39     & 0.473      & 0.446      \\ \hline
\end{tabular}
\caption{Ablation study of segmentation in the inputs.}
\label{tab:seg}
\end{table}

Except using the segmentation masks provided by the datasets, we can also extract the masks with off-the-shelf segmentation models such as DetectronV2~\cite{wu2019detectron2}. This way, the only input will be the RGB image. We evaluate the extra time cost of obtaining the masks by DetectronV2 in \cref{tab:seginfer}. 
After incorporating human and object segmentation into the pipeline, the reasoning time slightly increases from 0.208 to 0.264 seconds.
Benefiting from the extracted masks, our model achieves much better HOI reconstruction results than the `w/o seg.' variant.
Since the extracted segmentations have not undergone manual correction, some inaccuracies account for a slight decrease in global mesh reconstruction. 

\setlength{\tabcolsep}{1pt}
\begin{table}[h!]
\centering
\begin{tabular}{lccccc}
\hline
     & \small{CD$_\textrm{human}$$\downarrow$}   & \small{CD$_\textrm{object}$$\downarrow$}   & \small{Contact$_\textrm{p}$$\uparrow$}    & \small{Contact$_\textrm{r}$$\uparrow$} & \small{Time(s)}     \\ \hline
Ours    & \textbf{4.59} & \textbf{8.00} & 0.662 & \textbf{0.554} & \textbf{0.208} \\
Ours (w Det.) & 4.66          & 8.10          & \textbf{0.664}          & 0.550          & 0.264          \\
 \hline
\end{tabular}
\caption{Comparison of segmentations provided by the dataset and extracted by DetectronV2.}
\label{tab:seginfer}
\end{table}


\setlength{\tabcolsep}{6pt}
\begin{table*}[ht]
\centering
\begin{tabular}{ccc c ccccccc}
\toprule
\multicolumn{3}{c}{Human Graph Residual Block} && \multicolumn{3}{c}{Object Graph Residual Block} & \\
\cline{1-3} \cline{5-7}
Block1 & Block2 & Block3 && Block1 & Block2 & Block3 & CD$_\textrm{human}$$\downarrow$   & CD$_\textrm{object}$$\downarrow$   & Contact$_\textrm{p}$$\uparrow$    & Contact$_\textrm{r}$$\uparrow$        \\ \midrule
\xmark & \xmark & \xmark & & \xmark & \xmark & \xmark & 4.73          & 8.55 & 0.606 & 0.559                    \\ \hline
\cmark & \cmark & \cmark & & \xmark & \xmark & \xmark & 4.61          & 8.11          & 0.651          & 0.539 \\
\cmark & \cmark & \cmark && \cmark      & \xmark      & \xmark      & 4.62          & 8.01          & 0.638          & \textbf{0.591}  \\
\cmark & \cmark & \cmark && \xmark      & \cmark      & \xmark      & \textbf{4.59} & \textbf{8.00} & \textbf{0.662} & 0.554                   \\
\cmark & \cmark & \cmark && \xmark      & \xmark      & \cmark      & 4.68          & 8.43          & 0.643          & 0.534                   \\
\cmark & \cmark & \cmark && \cmark      & \cmark      & \cmark      & 4.62          & 8.05          & 0.644          & 0.573                \\ \bottomrule
\end{tabular}
\caption{Ablation study of the location of Object Graph Residual Block.}
\label{tab:graphlocation}
\end{table*}

\subsection{Location of graph residual blocks}
We integrate the Human Graph Residual Block in all three transformer encoder blocks. As for the proposed Object Graph Residual Block, we investigate the optimal location in \cref{tab:graphlocation}.
The results indicate that: i) Incorporating the object graph residual block at any layer positively contributes to human and object reconstruction. ii) Adding a graph convolutional network (GCN) to the first and second blocks yields more significant improvements in reconstruction. This suggests that the self-attention mechanism at higher layers struggles to distinguish clear boundaries between human and object features. 
Therefore, our HOI-TG only equips the second transformer encoder block with the Object Graph Residual Block.

\section{Generalization to in-the-wild images.}

\setlength{\tabcolsep}{1pt}
\begin{figure}[ht]
    \centering
    \begin{tabular}{cccc} \includegraphics[width=\rgbwidth\textwidth]{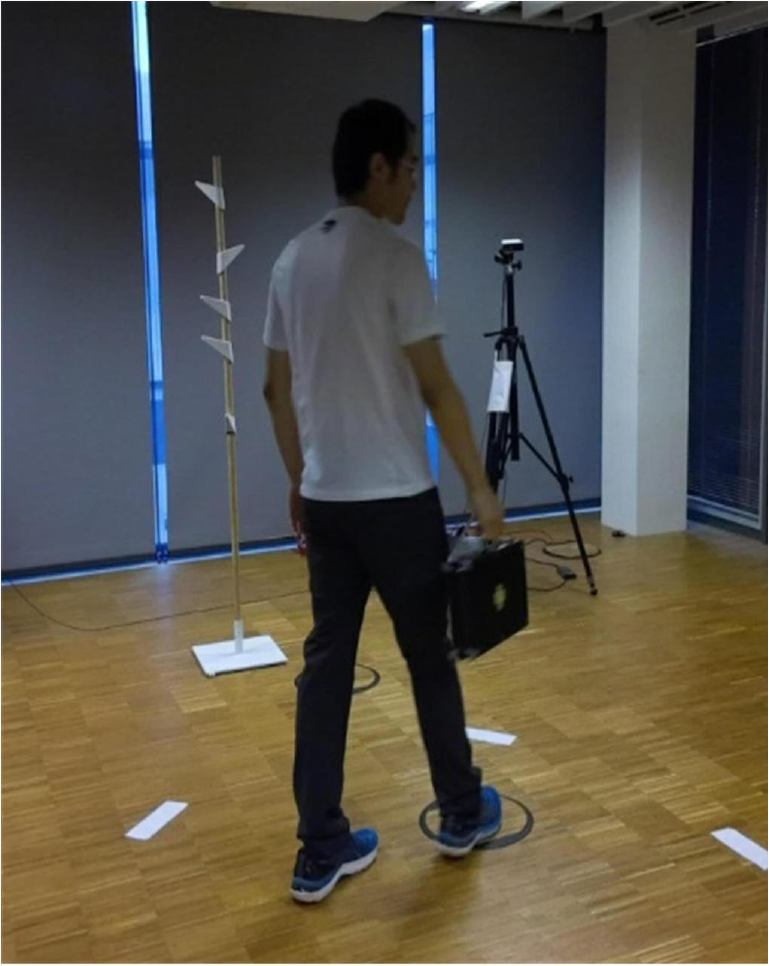} & \includegraphics[width=\meshwidth\textwidth]{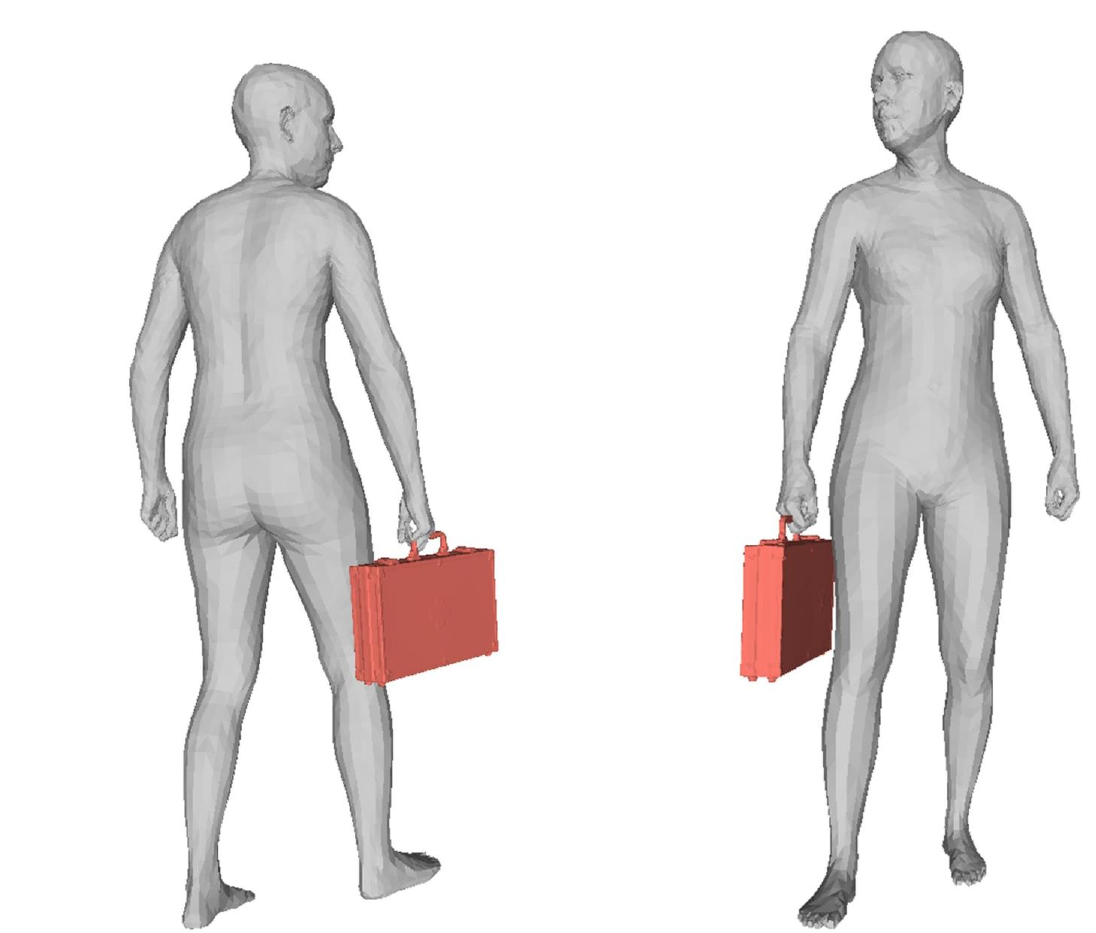} & \includegraphics[width=\rgbwidth\textwidth]{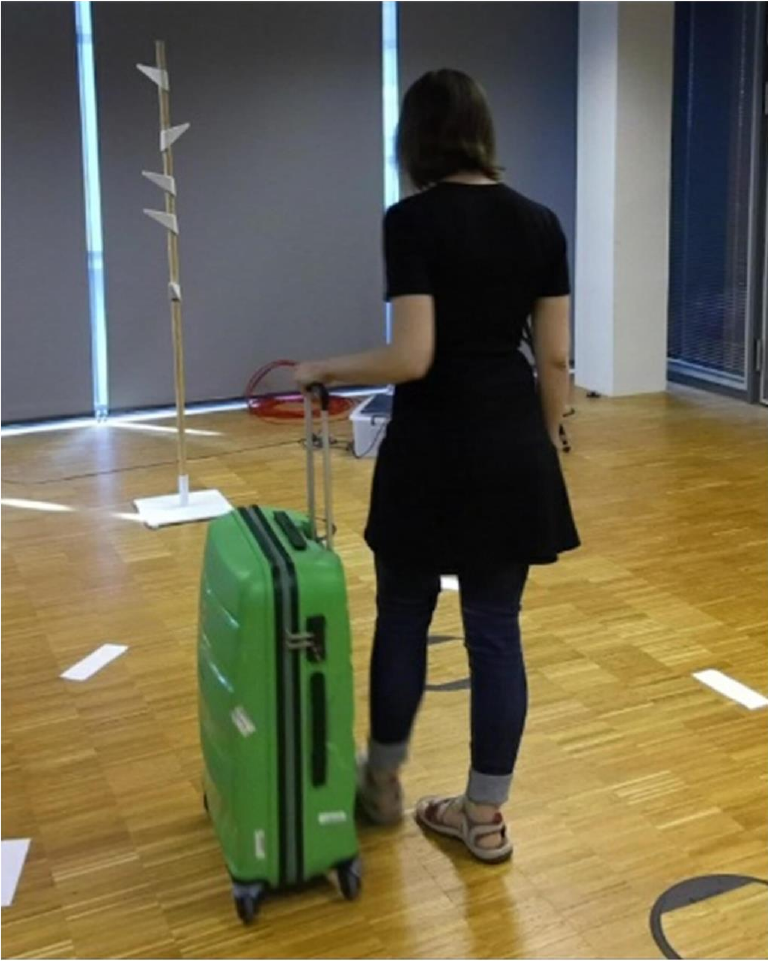} & \includegraphics[width=\meshwidth\textwidth]{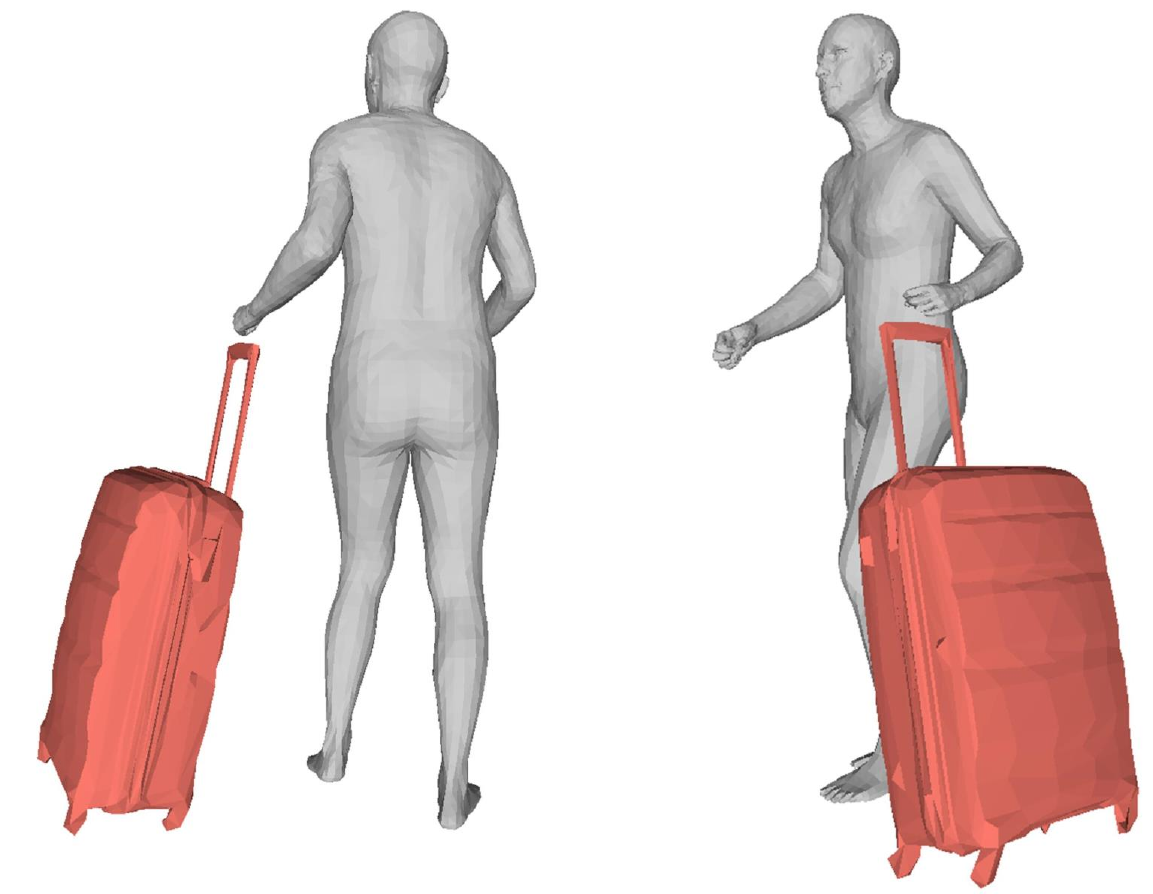} \\
    \includegraphics[width=\rgbwidth\textwidth]{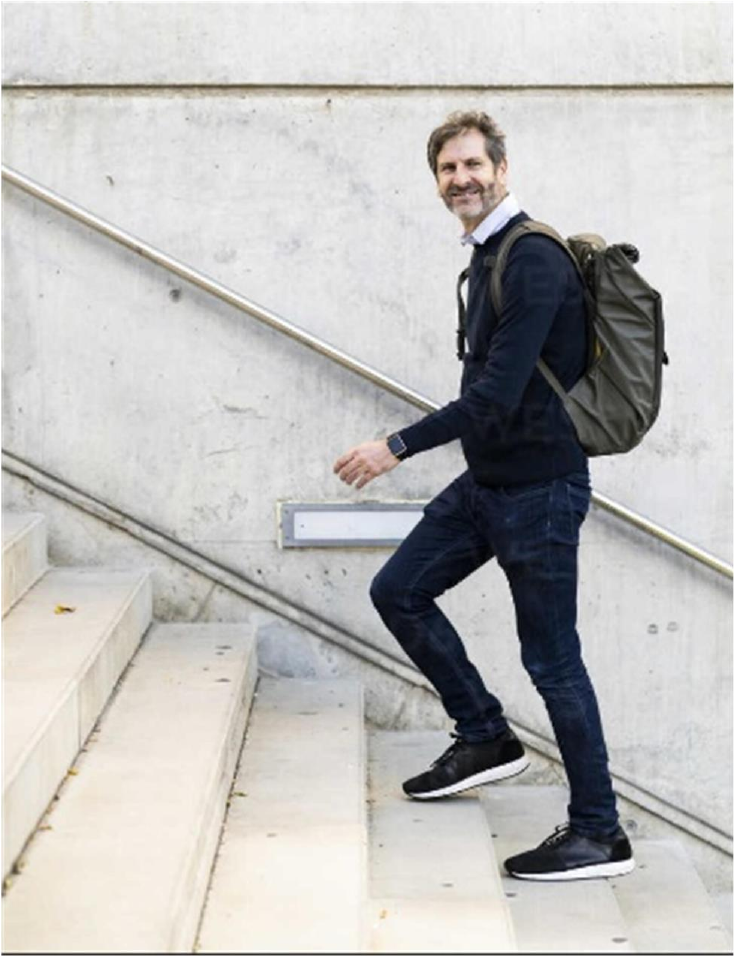}     & \includegraphics[width=\meshwidth\textwidth]{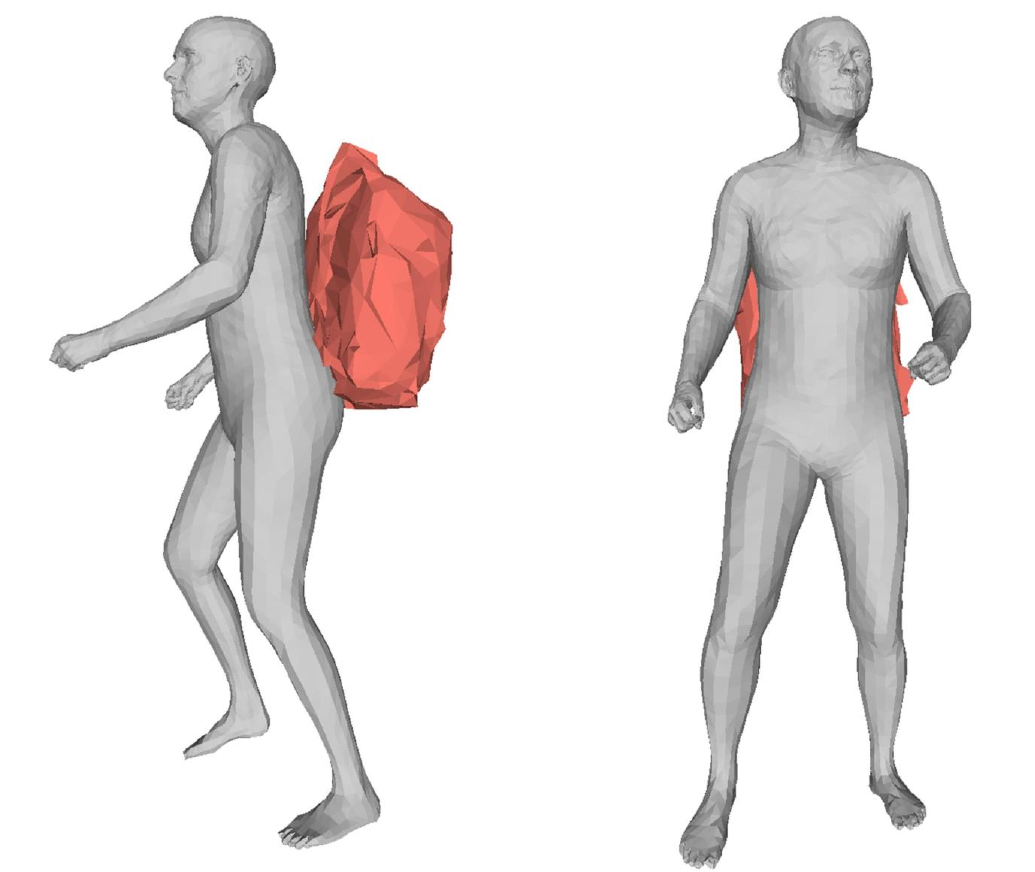} & \includegraphics[width=\rgbwidth\textwidth]{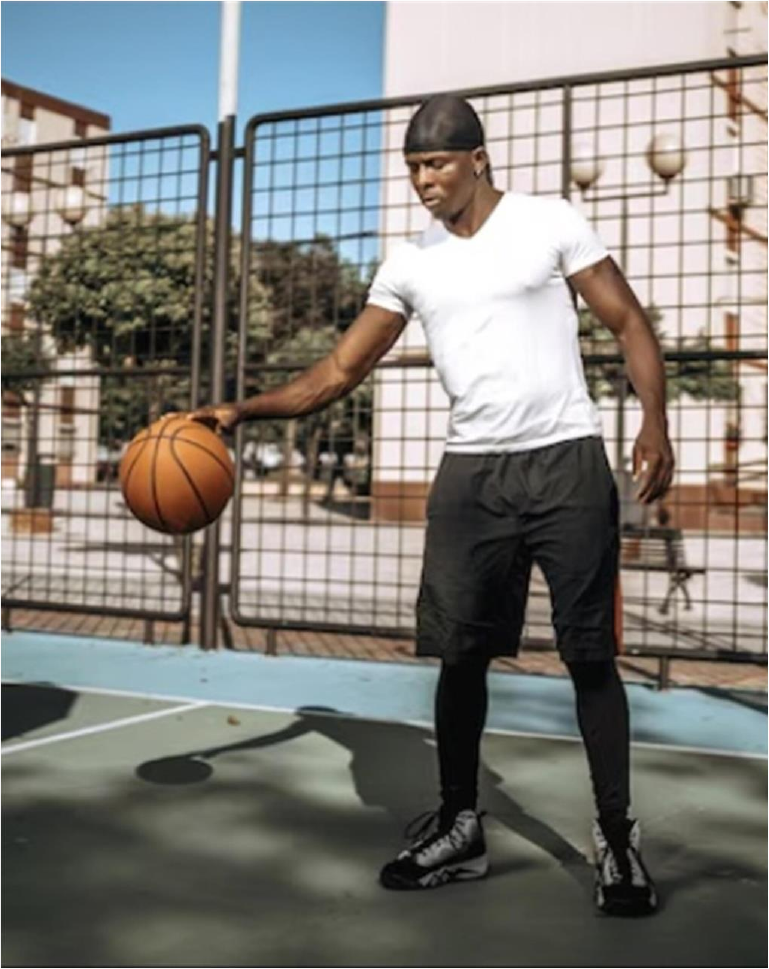} & \includegraphics[width=\meshwidth\textwidth]{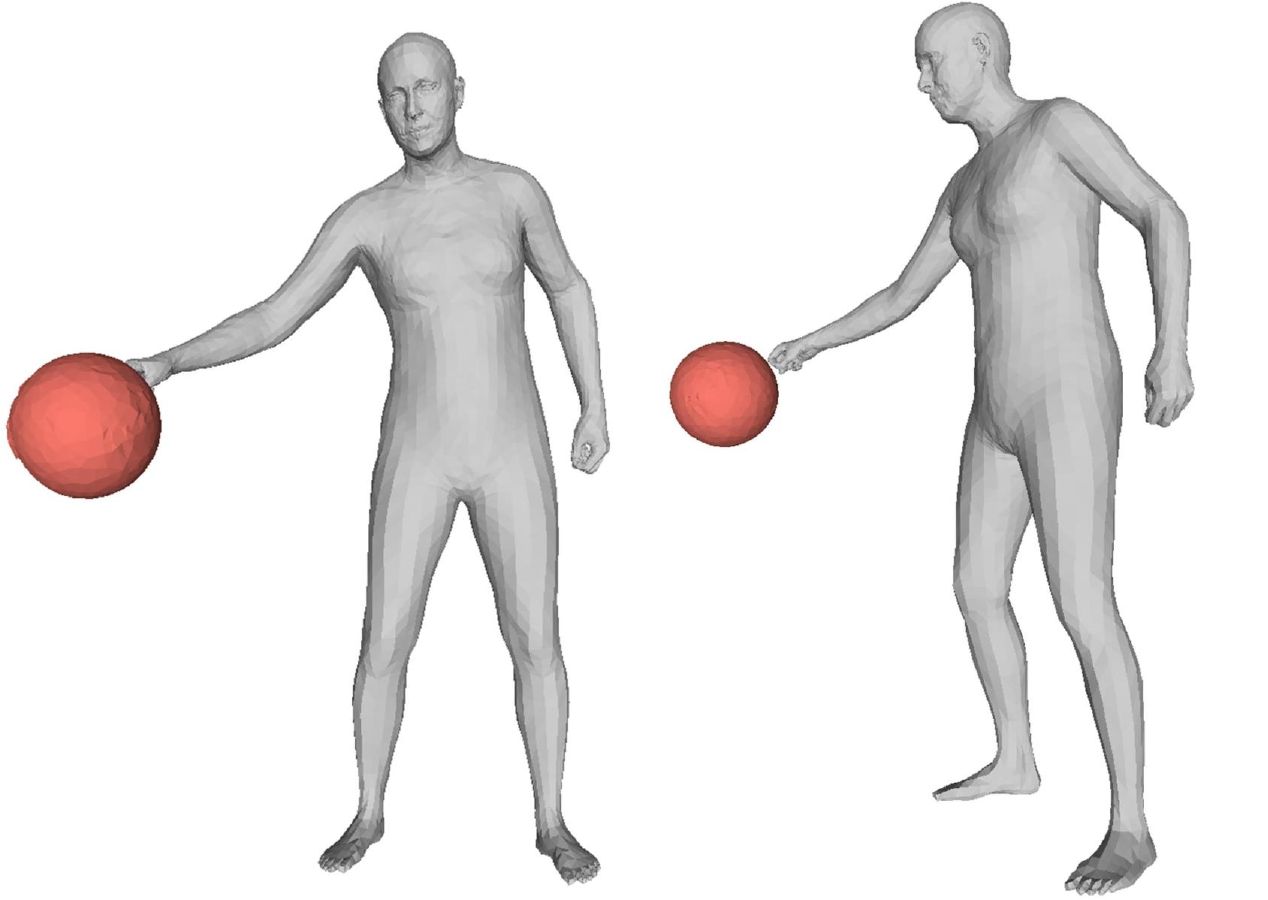}
    \end{tabular}
    \caption{Estimated HOI of InterCap samples using HOI-TG trained on BEHAVE (row 1) and of images in the wild (row 2).}
    \label{fig:Generalization}
\end{figure}

For generalization, \cref{fig:Generalization} shows the estimated HOI of InterCap samples using HOI-TG trained on the BEHAVE dataset and applying HOI-TG to images in the wild. For the first row, we use the model trained on the BEHAVE~\cite{behave} dataset directly for testing on the InterCap~\cite{intercap} dataset. For the second row, we directly use the model trained on the BEHAVE~\cite{behave} dataset to test the reconstruction results of in-the-wild images.The results indicate that HOI-TG possesses a certain degree of generalization ability. However, given that our method is data-driven, the generalization capability of HOI reconstruction in complex scenes still requires further investigation.

\section{Efficiency and limitations}

\subsection{Computational efficiency}
We summarize the parameters, running time, and performance of different methods in \cref{tab:infer}. The data related to PHOSA~\cite{PHOSA} and CHORE~\cite{chore} are provided by CONTHO~\cite{joint} and StackFLOW~\cite{stackflow}. For StackFLOW~\cite{stackflow}, CONTHO~\cite{joint}, and our HOI-TG, we evaluate the speed using the same environment on a single RTX 4090 GPU, employing a batch size of 1 for multiple inferences and calculate the average running time. We only consider the inference time, excluding data preprocessing.
According to \cref{tab:infer}, our HOI-TG outperforms previous methods in human-object interaction reconstruction while achieving the fastest inference speed. 
Both the optimization-based method PHOSA~\cite{PHOSA} and the neural field-based model CHORE~\cite{chore} require significantly longer inference time. The post-optimization process of StackFLOW~\cite{stackflow} also consumes quite a lot of time. In contrast, our HOI-TG can infer much faster and improves both global mesh reconstruction and local contact modeling. The results indicate that we have developed a more elegant and powerful architecture than previous approaches.

\setlength{\tabcolsep}{2pt}
\begin{table}[h!]
\centering
\begin{tabular}{lcccc}
\hline
Method        & \small{Params(M)} & \small{Time(s)}        & \small{Chamfer Dist$\downarrow$}   & \small{F1 Score$\uparrow$}       \\ \hline
PHOSA         & -         & 165.3          & 19.395         & 0.317          \\
CHORE         & 18.19     & 312.2          & 8.120          & 0.523          \\ \hline
StackFLOW     & 83.43     & 15.67          & 8.885          & 0.499          \\
CONTHO        & 82.80     & 0.218          & 6.705          & 0.554          \\
\textbf{Ours} & 122.81    & \textbf{0.208} & \textbf{6.295} & \textbf{0.603} \\ \hline
\end{tabular}
\caption{Comparison of model efficiency and performance on BEHAVE~\cite{behave} dataset.}
\label{tab:infer}
\end{table}

\subsection {Limitations}
This section discusses cases where our HOI-TG fails to produce satisfactory reconstruction results and analyze the reason.

\begin{figure*}
    \centering
    \begin{tabular}{cccccc}
    \includegraphics[height=0.15\textwidth]{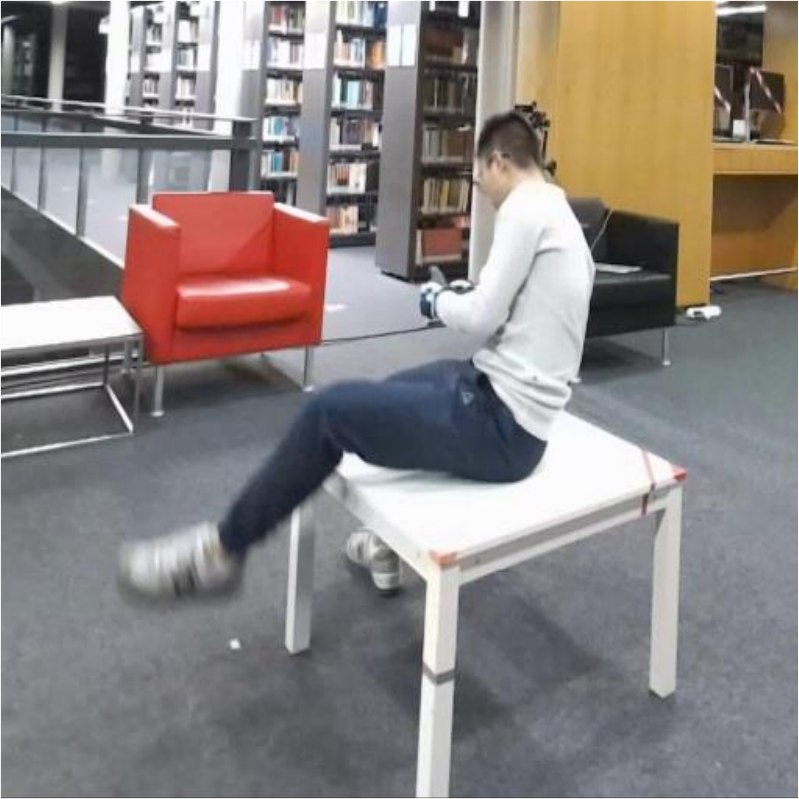}     & \includegraphics[height=0.15\textwidth]{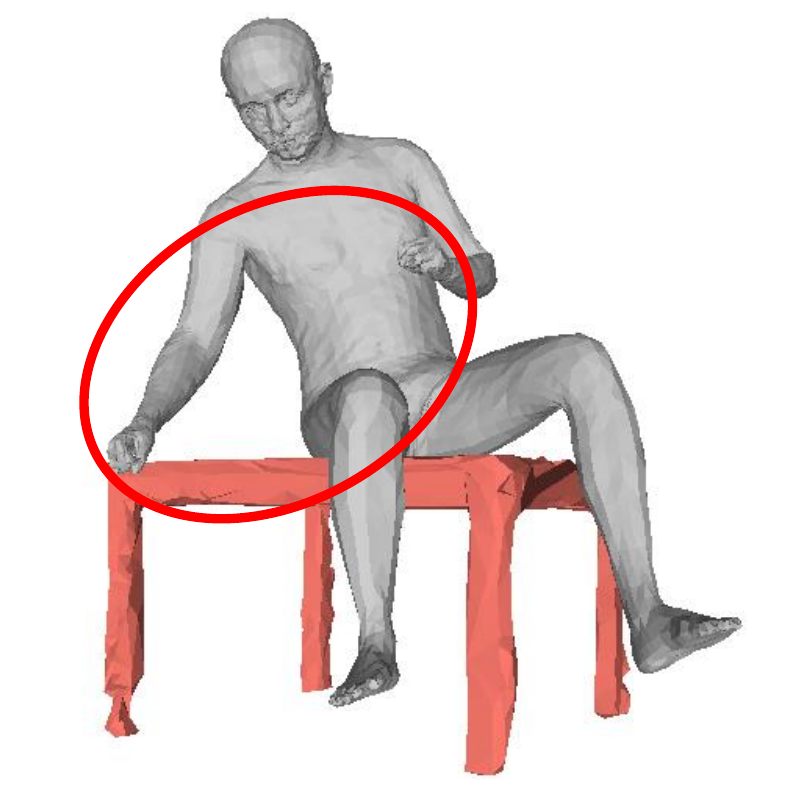} & \includegraphics[height=0.15\textwidth]{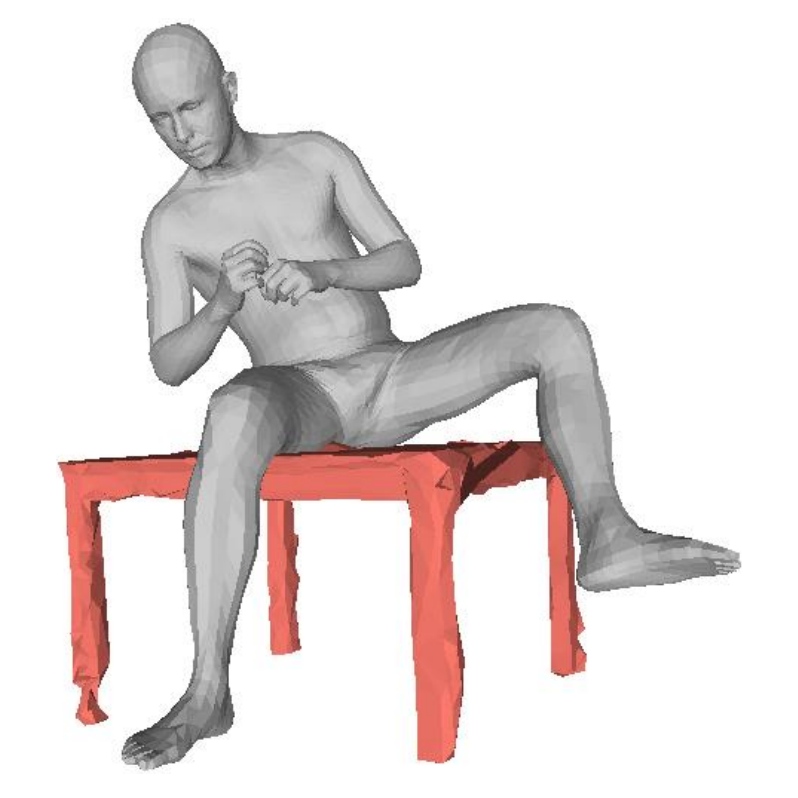}  &
    \includegraphics[height=0.15\textwidth]{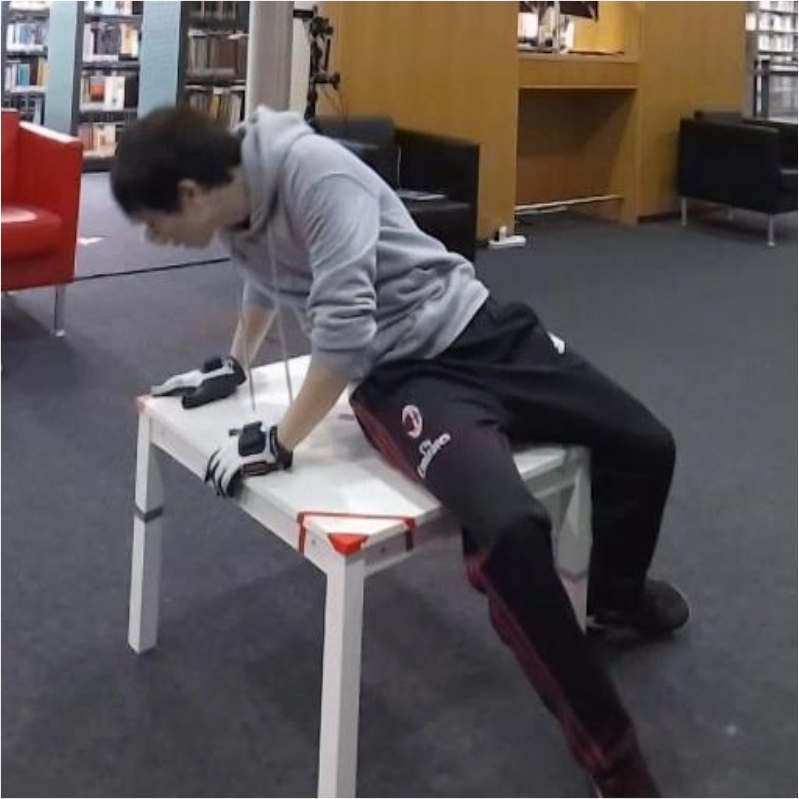}     & \includegraphics[height=0.15\textwidth]{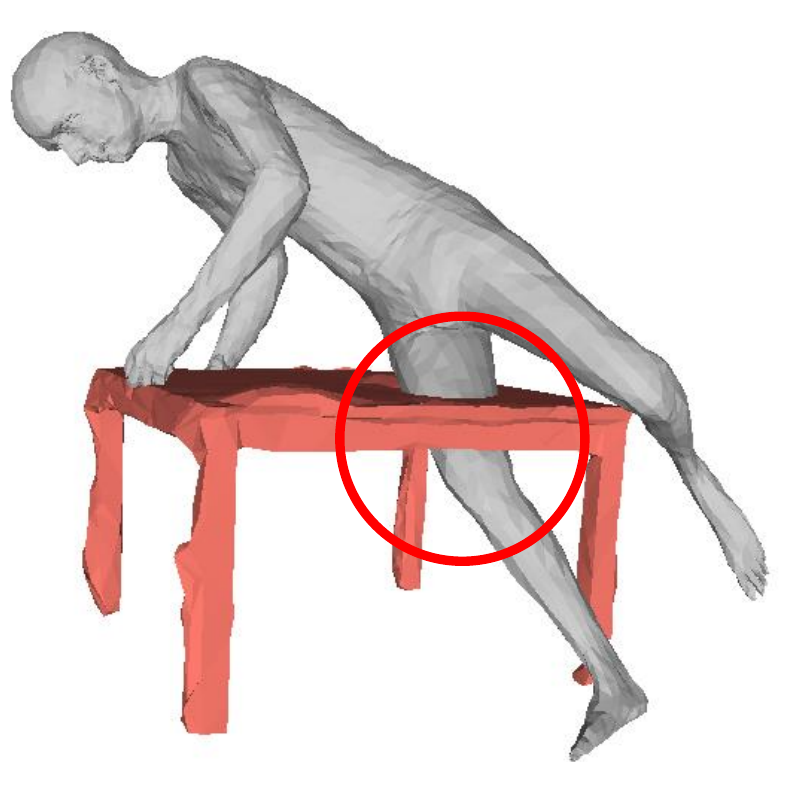} & \includegraphics[height=0.15\textwidth]{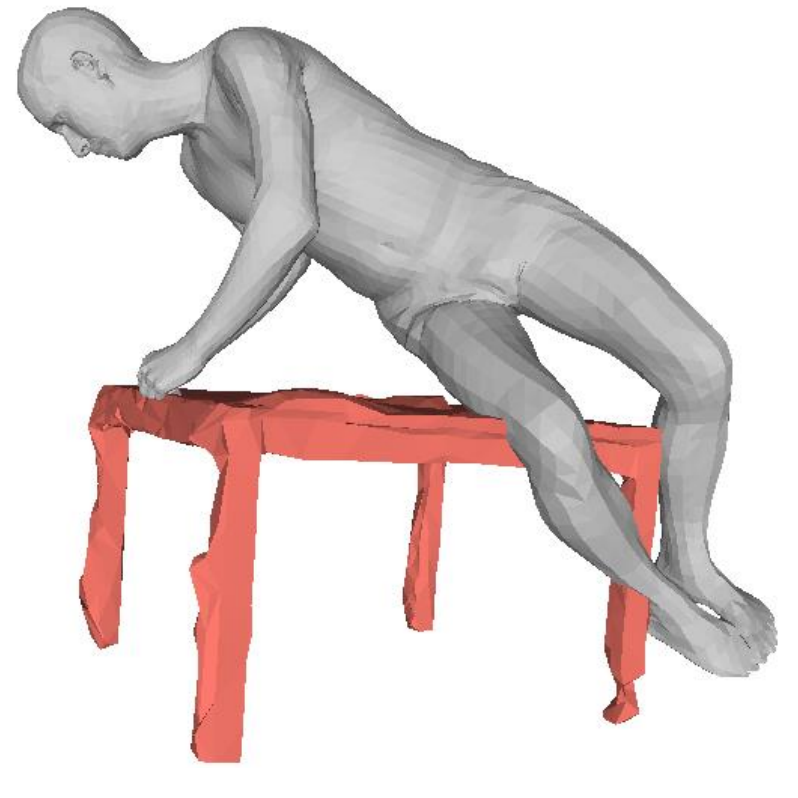}  \\
    \includegraphics[height=0.15\textwidth]{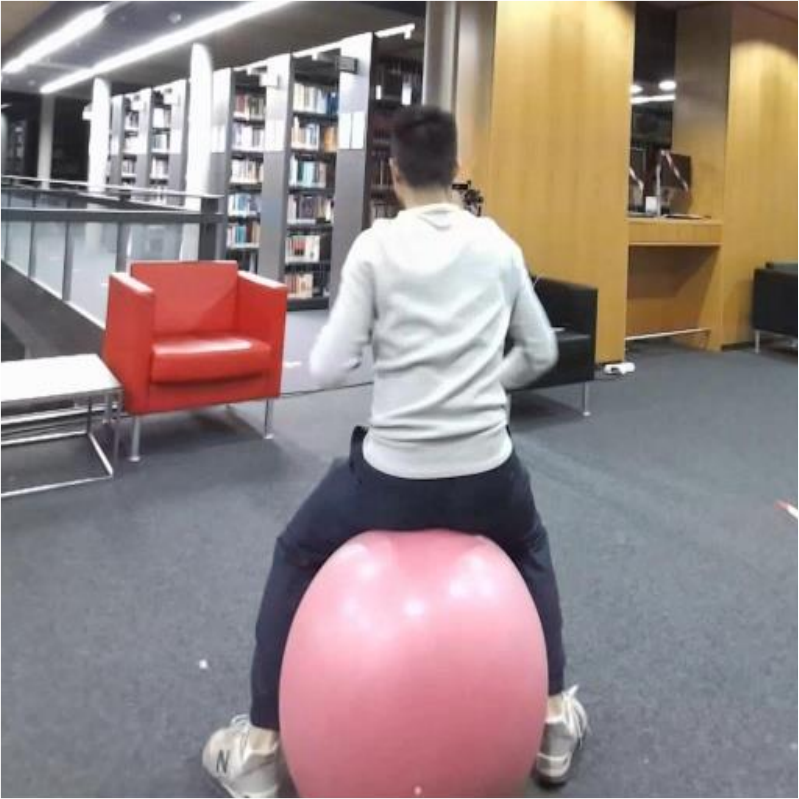}     & \includegraphics[height=0.15\textwidth]{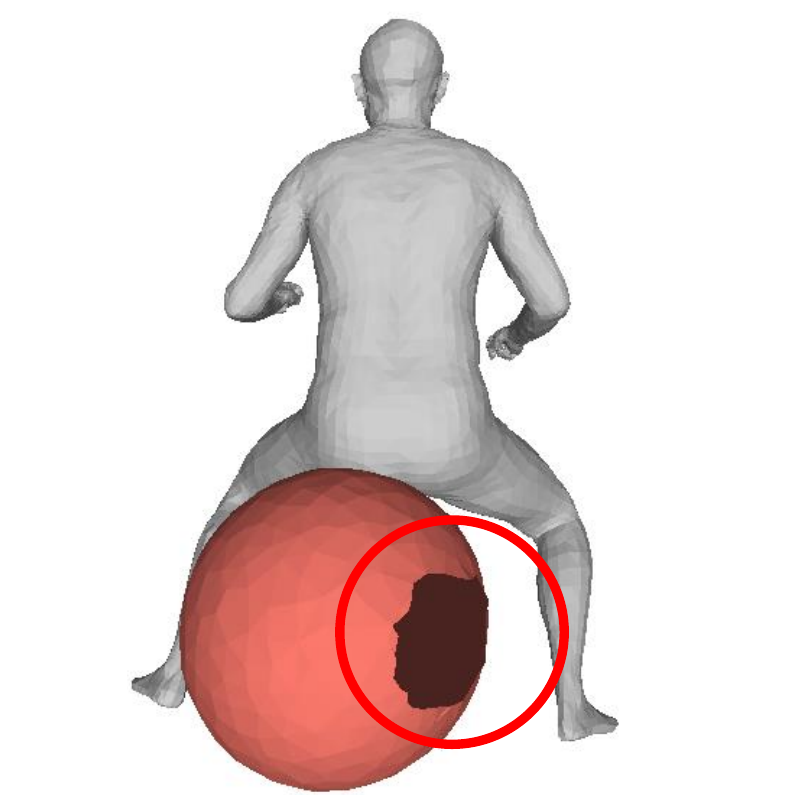} & \includegraphics[height=0.15\textwidth]{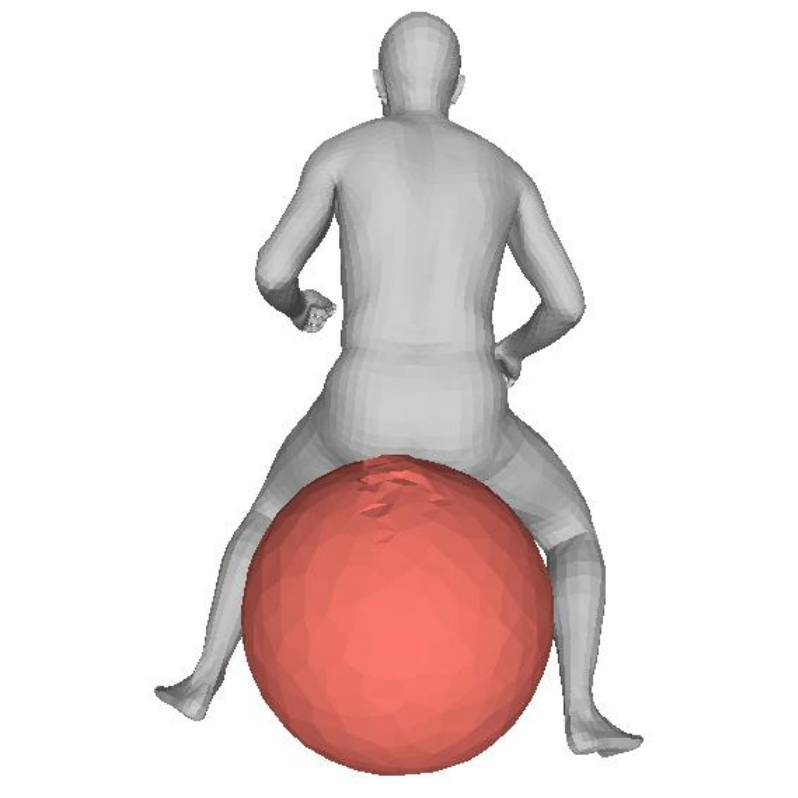} &
    \includegraphics[height=0.15\textwidth]{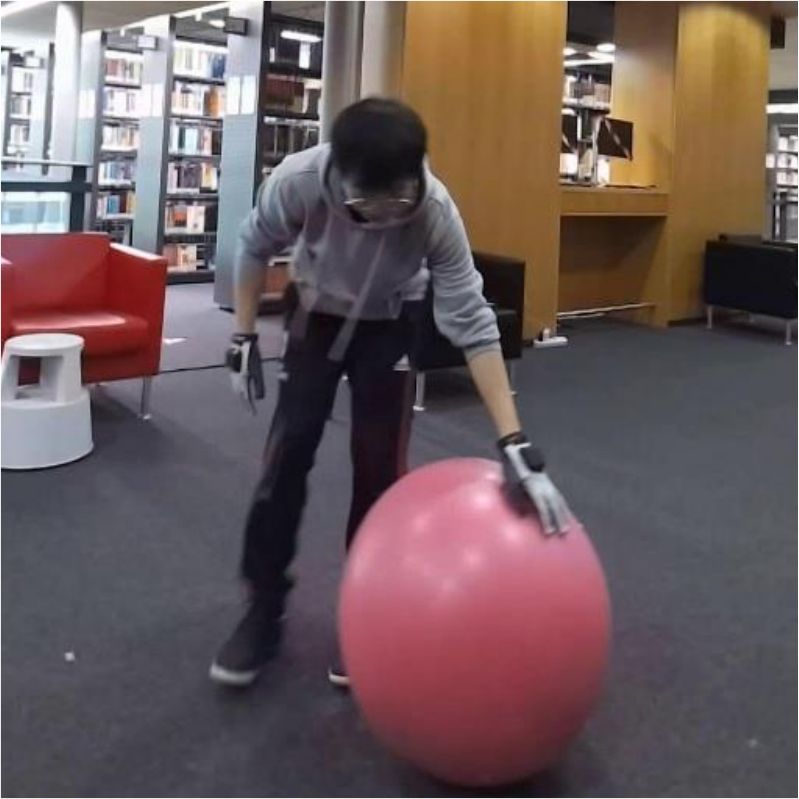}     & \includegraphics[height=0.15\textwidth]{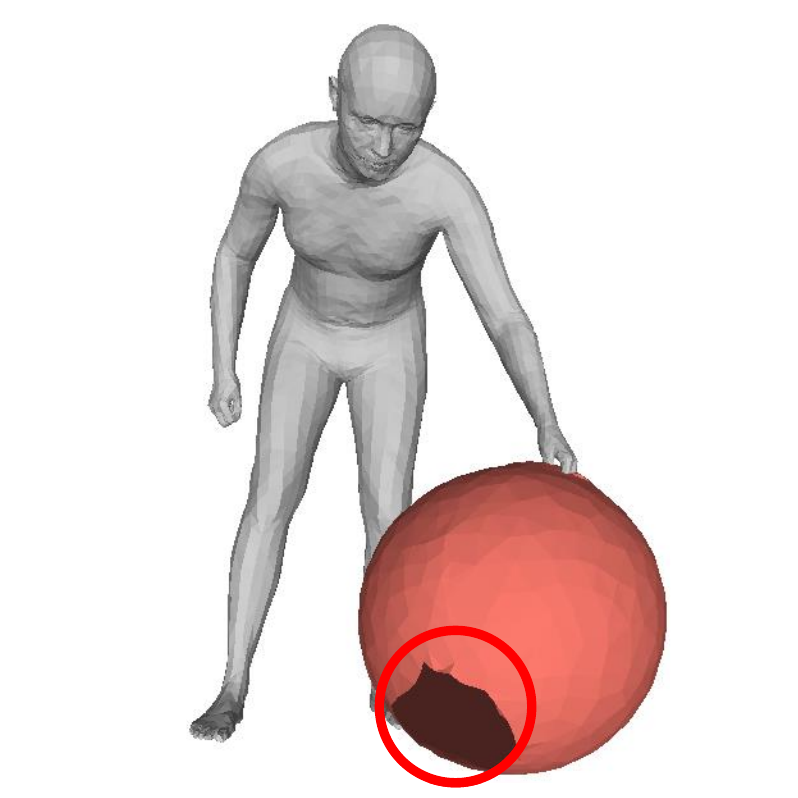} & \includegraphics[height=0.15\textwidth]{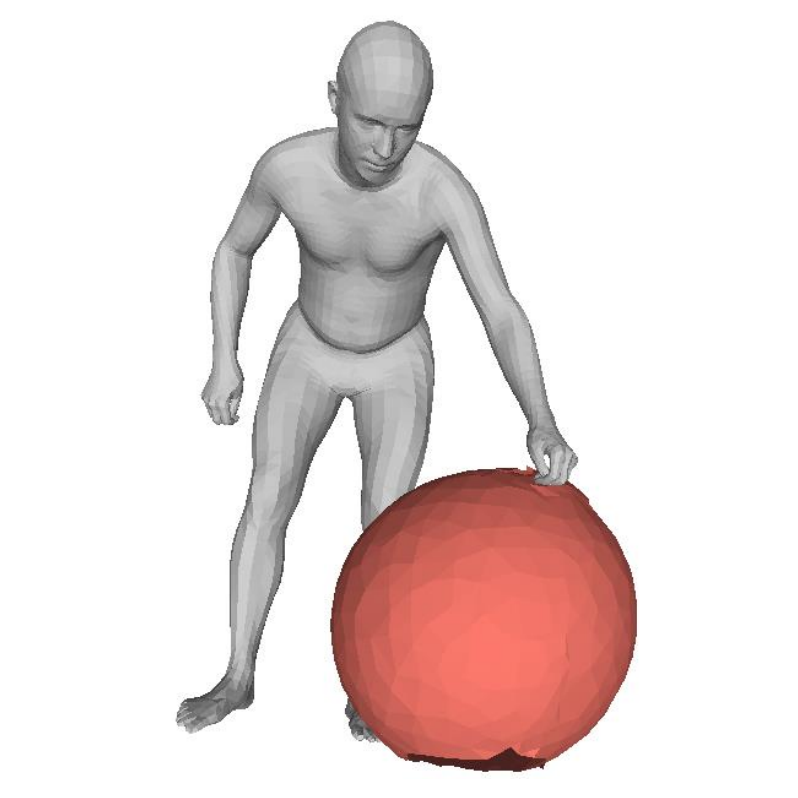}  \\
     RGB & HOI-TG (Ours) & GT  & RGB & HOI-TG (Ours) & GT
    \end{tabular}
    \caption{Failure cases of our HOI-TG. We highlight the region with red circles.}
    \label{fig:failure}
\end{figure*}

\noindent\textbf{Lying poses:} Our model may not perform well on complex or rare postures like lying down. As shown in the first row of \cref{fig:failure}, our model cannot accurately predict the posture of arms and legs and even cause mesh penetration. We consider that is mainly because actions such as lying down cause most body parts to be self-occluded. Such self-occlusion poses a big challenge for the ResNet50 backbone in producing meaningful initial human mesh. As a result, our model may fail to distinguish among different human vertices and predict inaccurate human mesh and object pose.


\noindent\textbf{Purely symmetric objects:} Accurately reconstructing highly symmetrical objects such as spherical and square has always been challenging in HOI reconstruction. The difficulty of capturing the object details may result in inaccurate rotation prediction. As shown in the second row of \cref{fig:failure}, our model cannot accurately estimate the correct rotation posture of the yoga ball.

\setlength{\tabcolsep}{1pt}

\section{More visual comparison results}
\begin{figure*}
    \centering
    \begin{tabular}{cccc}
    \includegraphics[height=0.2\textwidth]{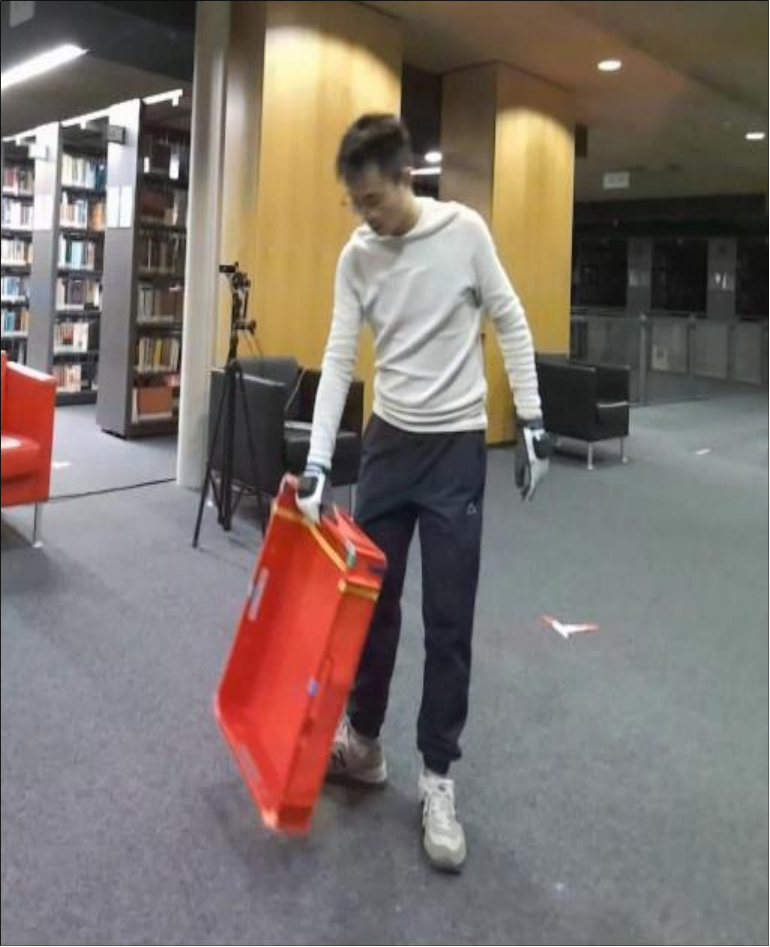}     & \includegraphics[height=0.2\textwidth]{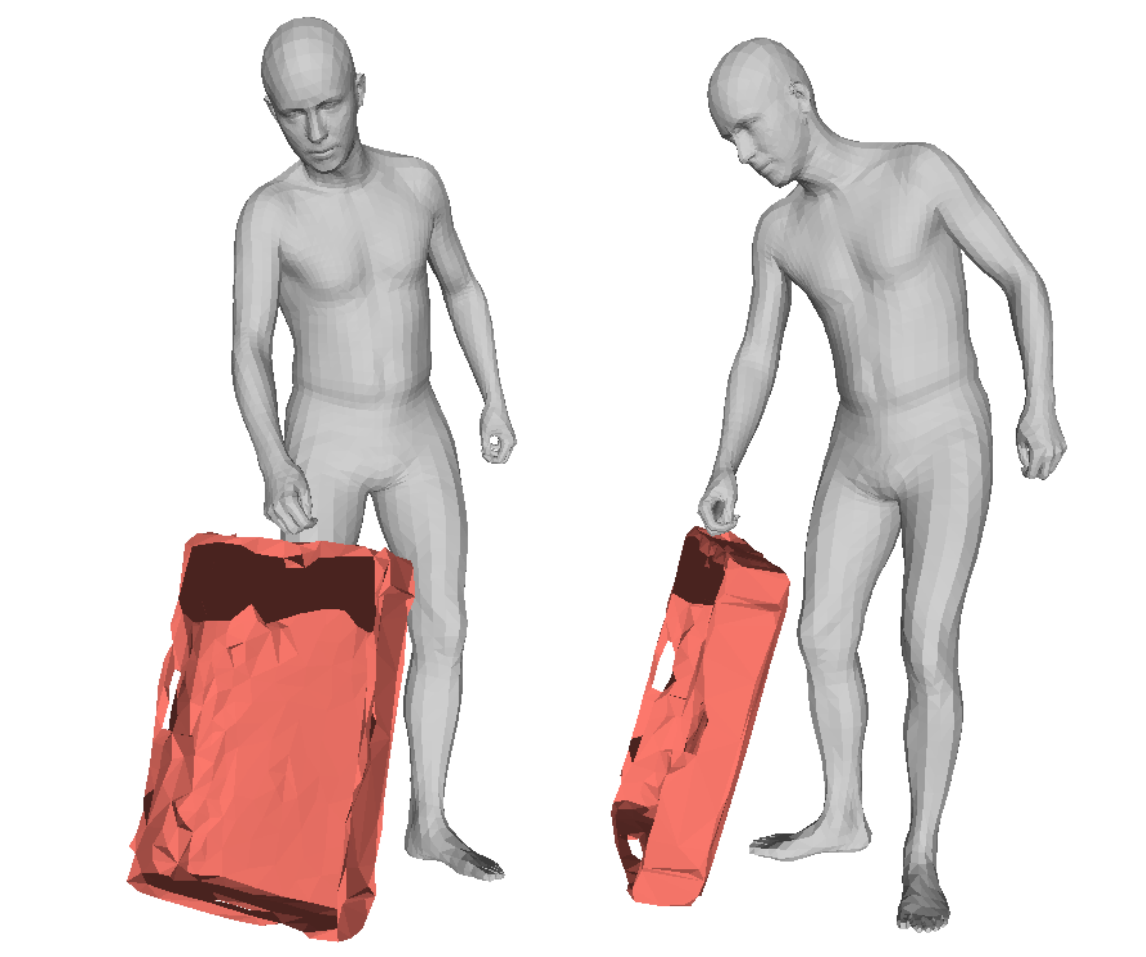} & \includegraphics[height=0.2\textwidth]{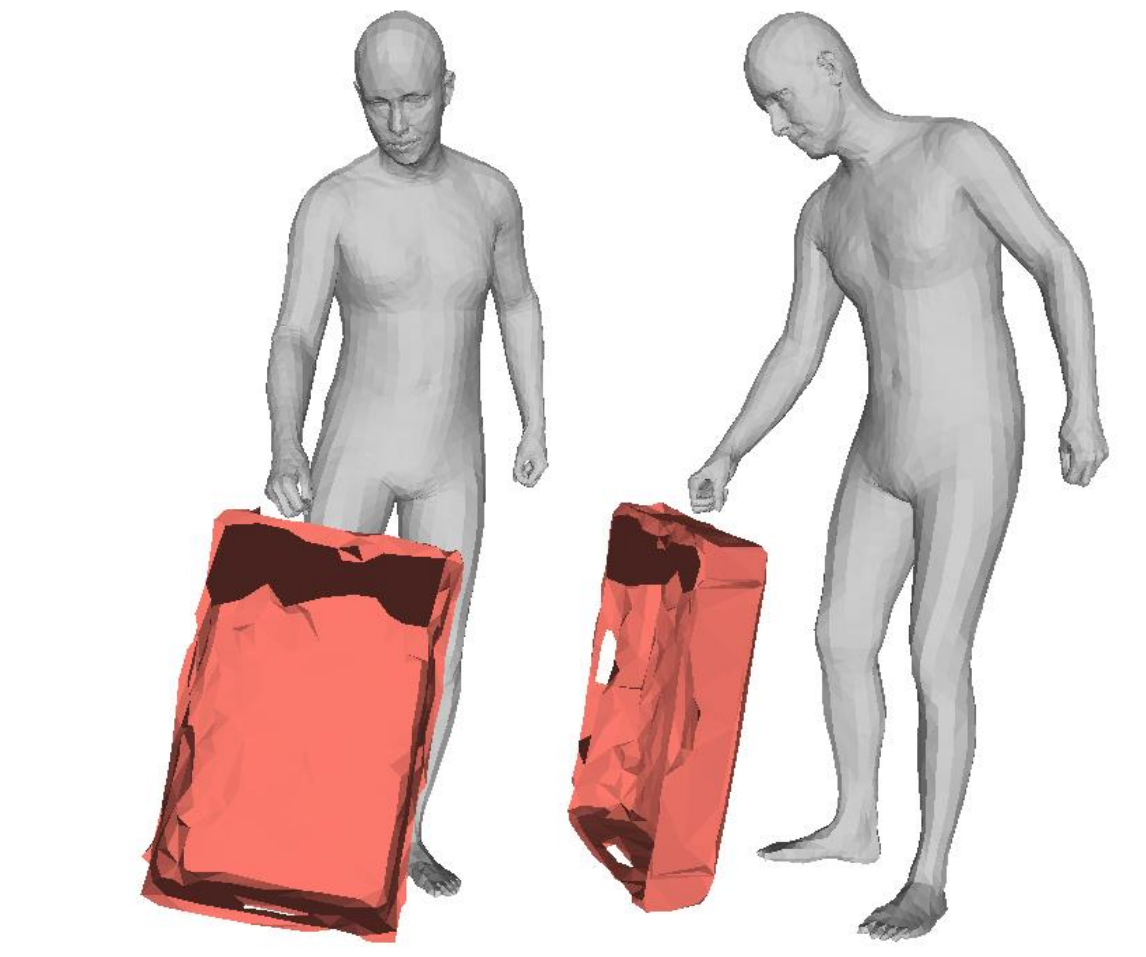} & \includegraphics[height=0.2\textwidth]{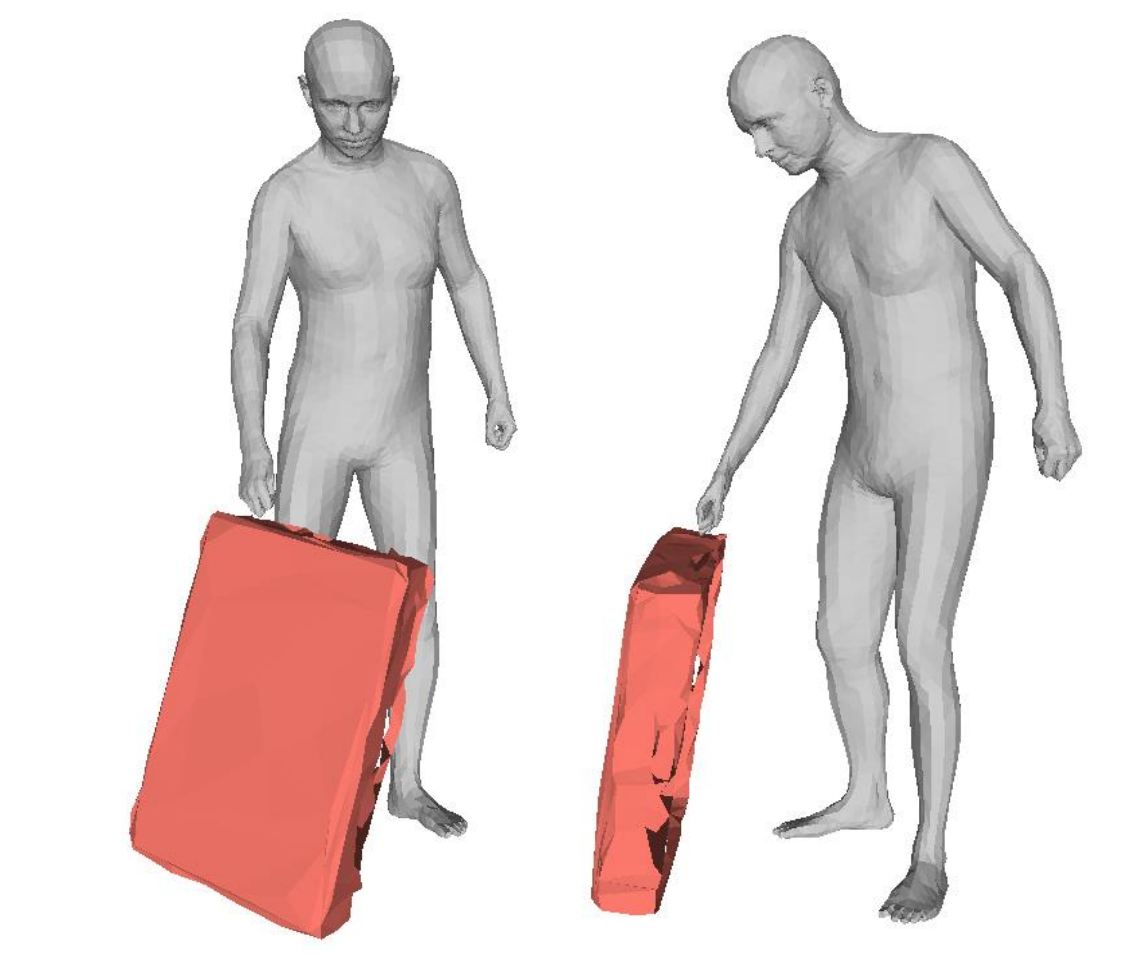} \\
    \includegraphics[height=0.2\textwidth]{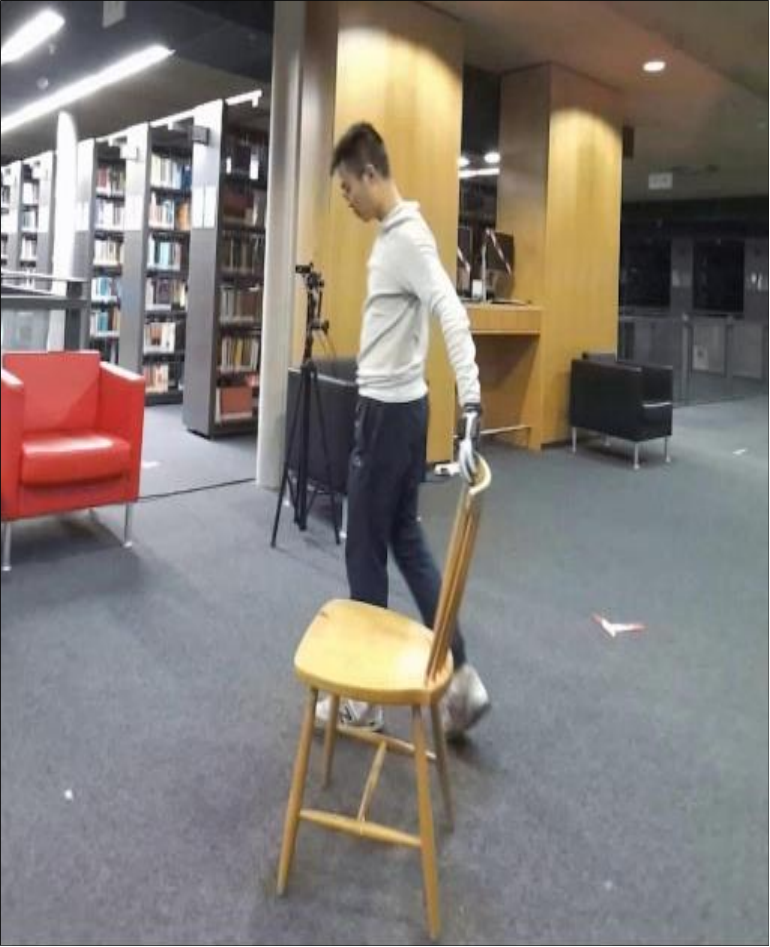}     & \includegraphics[height=0.2\textwidth]{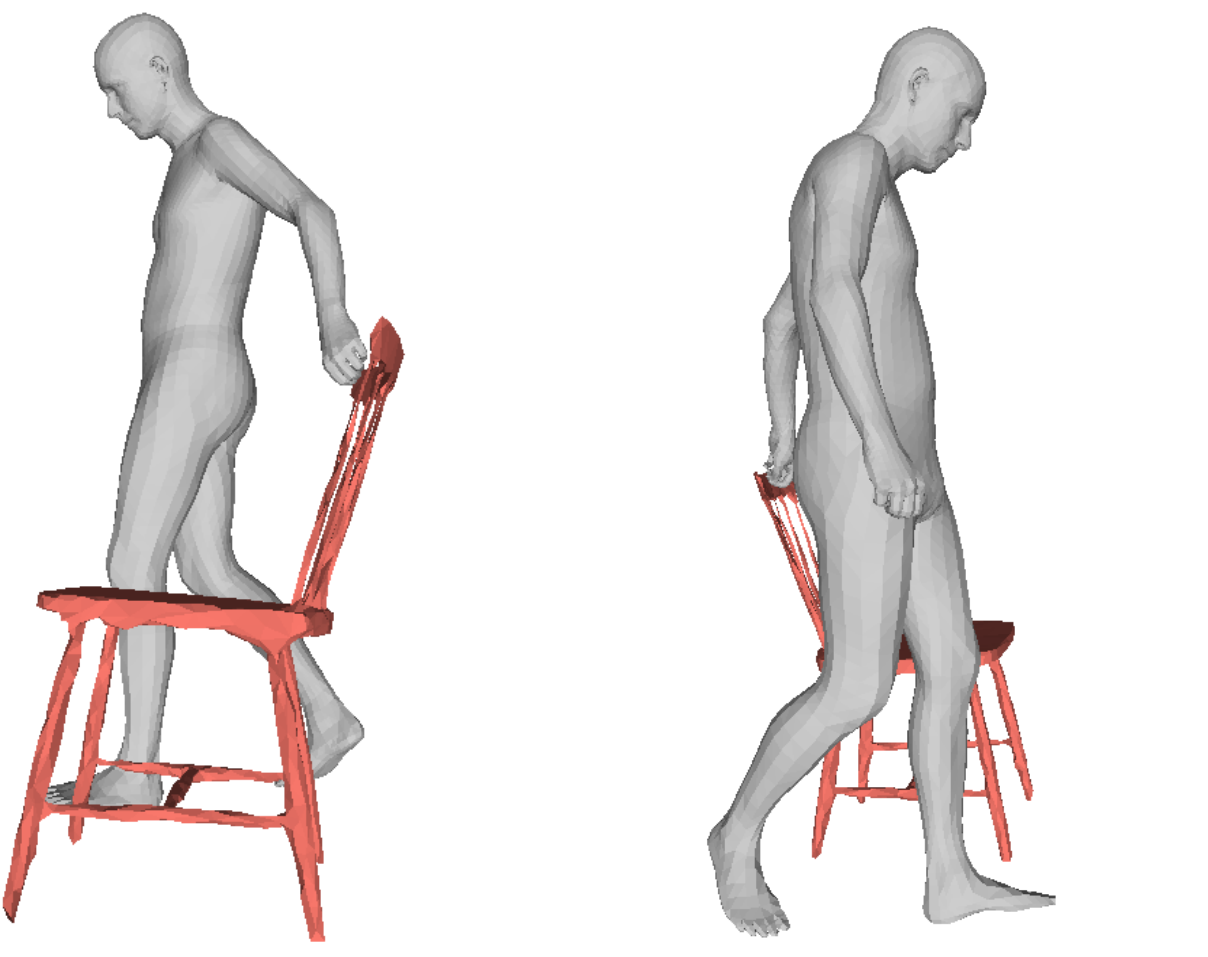} & \includegraphics[height=0.2\textwidth]{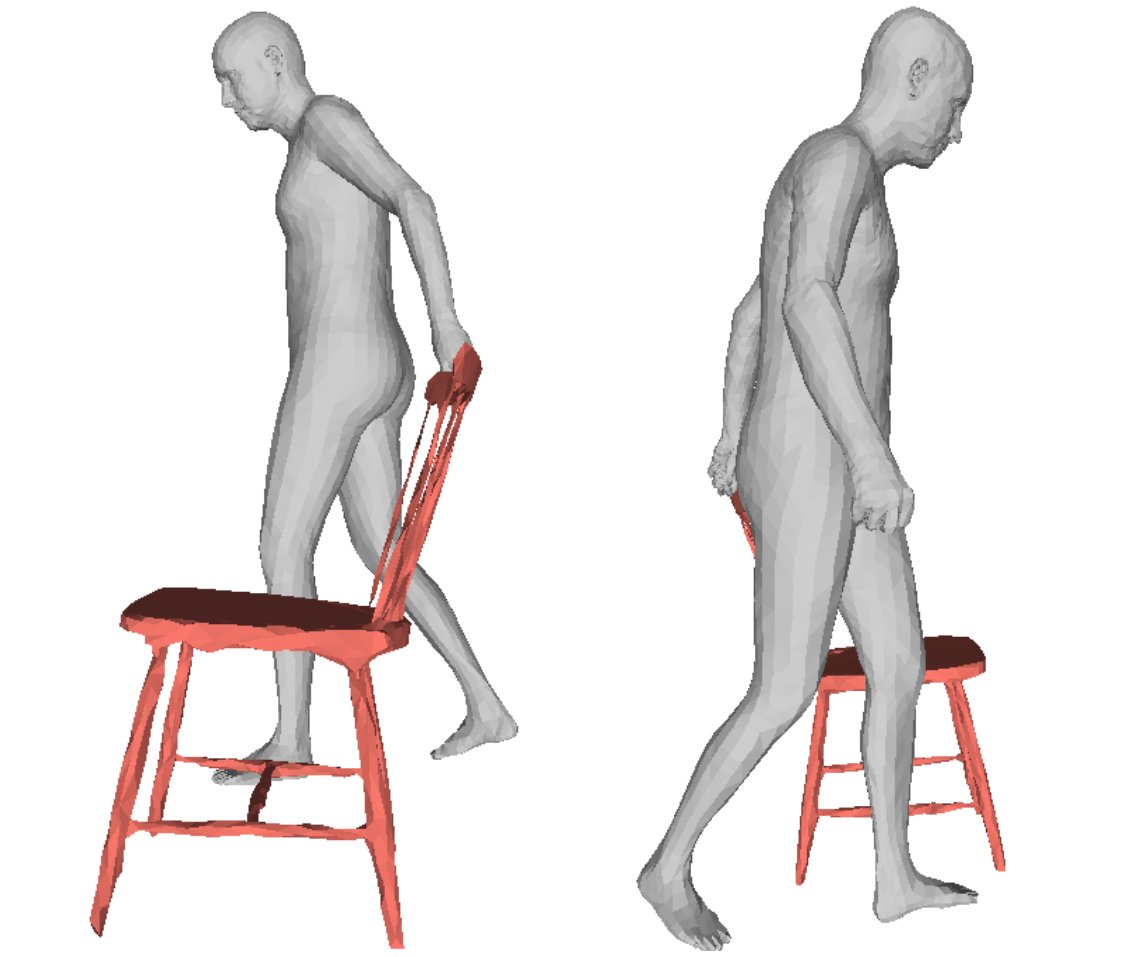} & \includegraphics[height=0.2\textwidth]{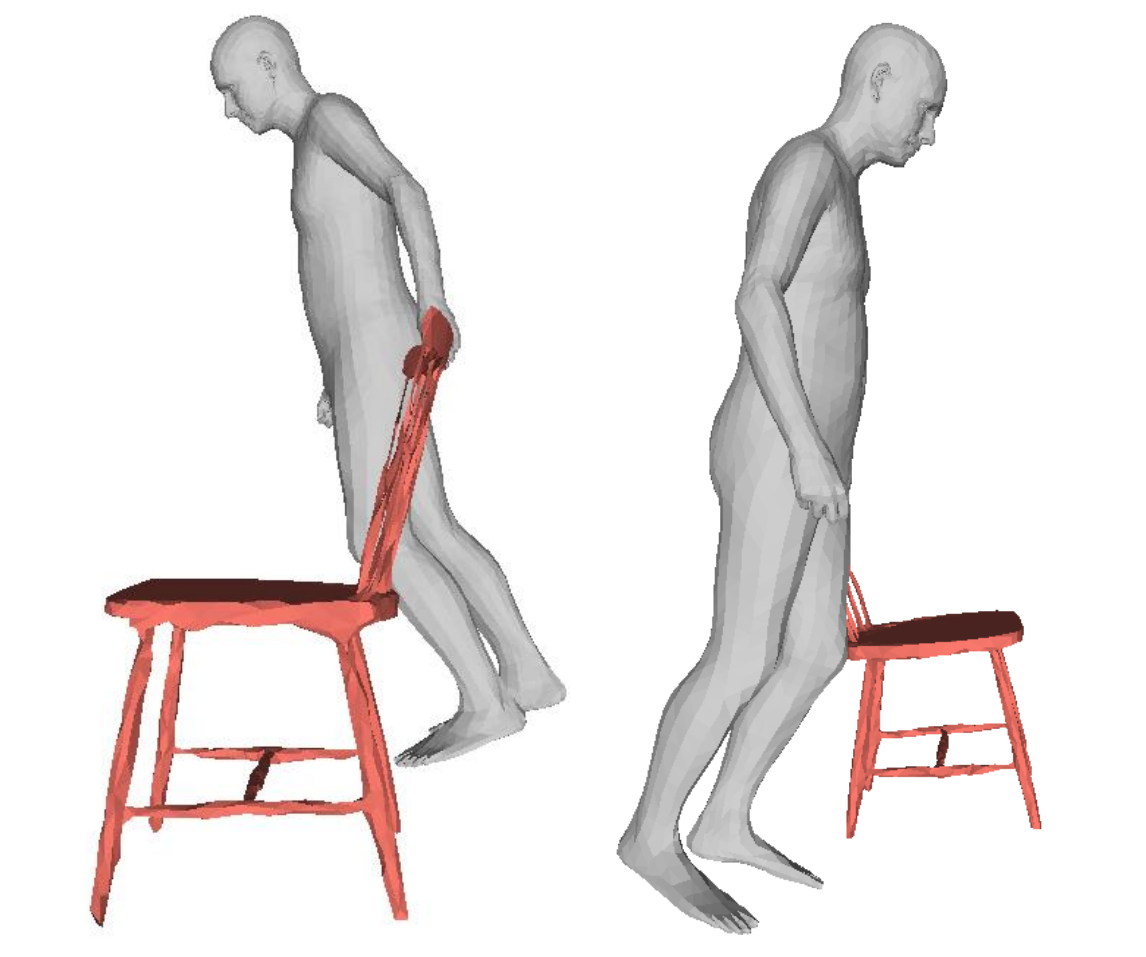} \\
    \includegraphics[height=0.2\textwidth]{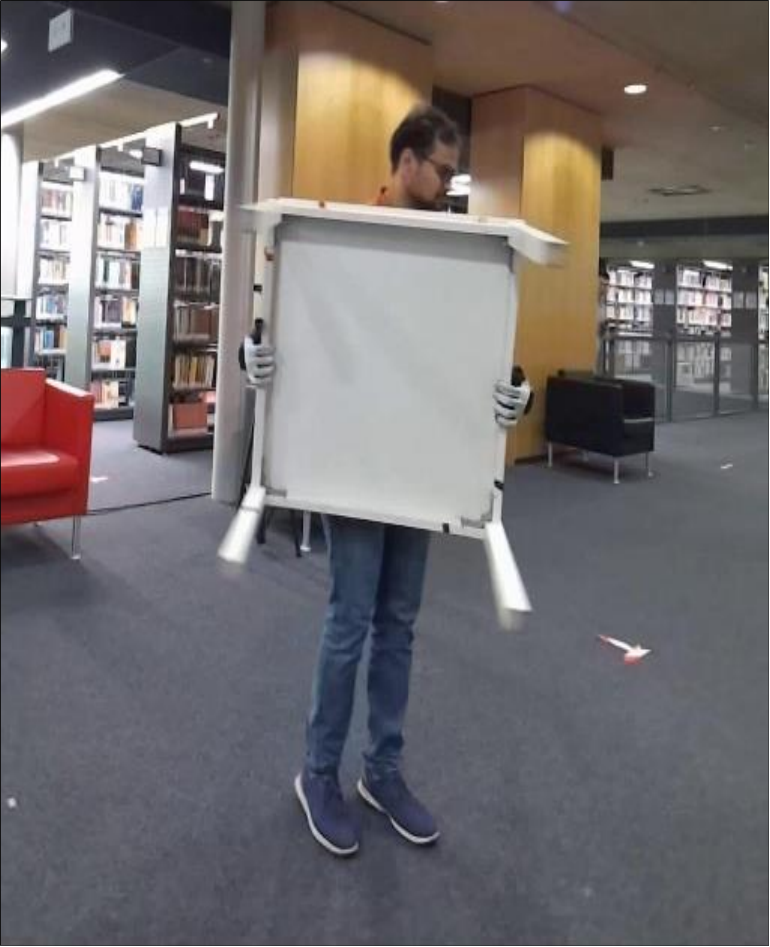}     & \includegraphics[height=0.2\textwidth]{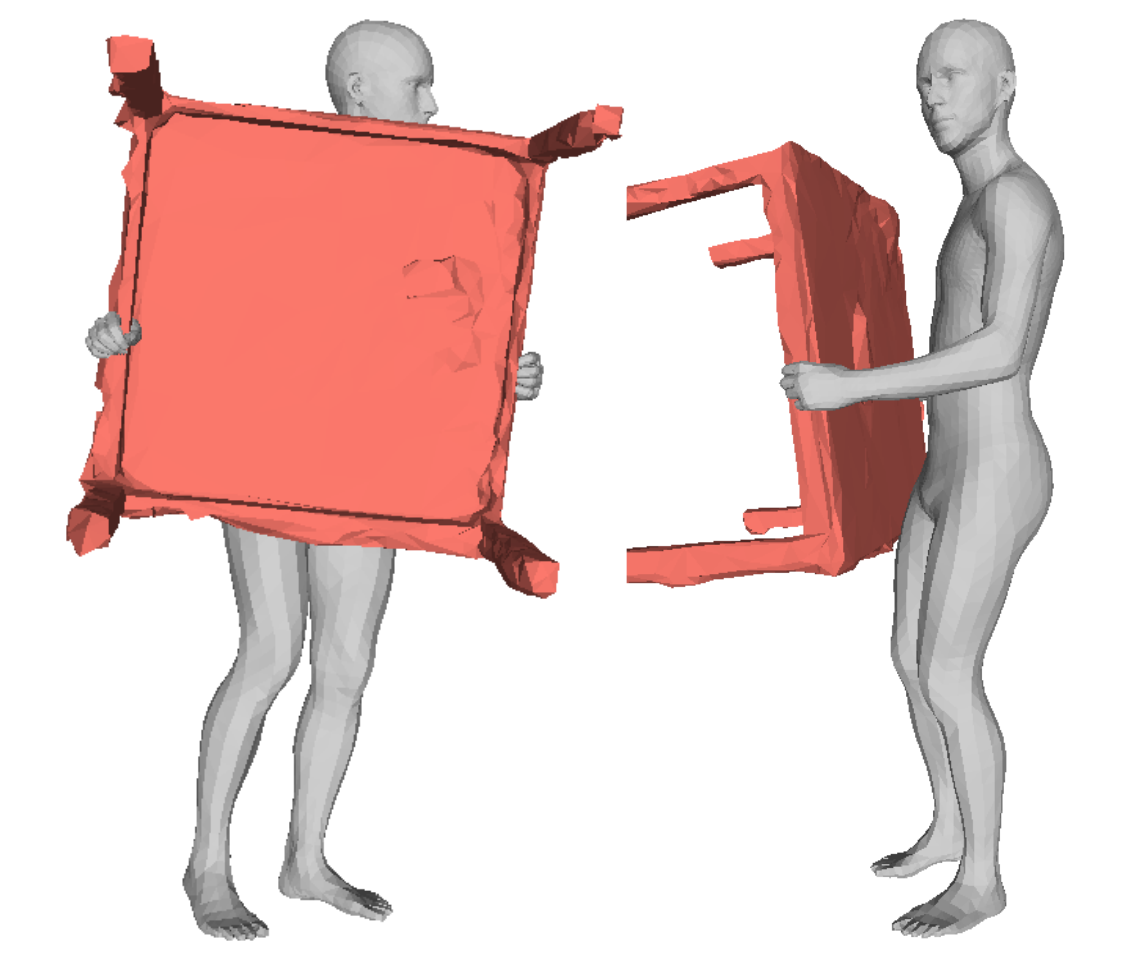} & \includegraphics[height=0.2\textwidth]{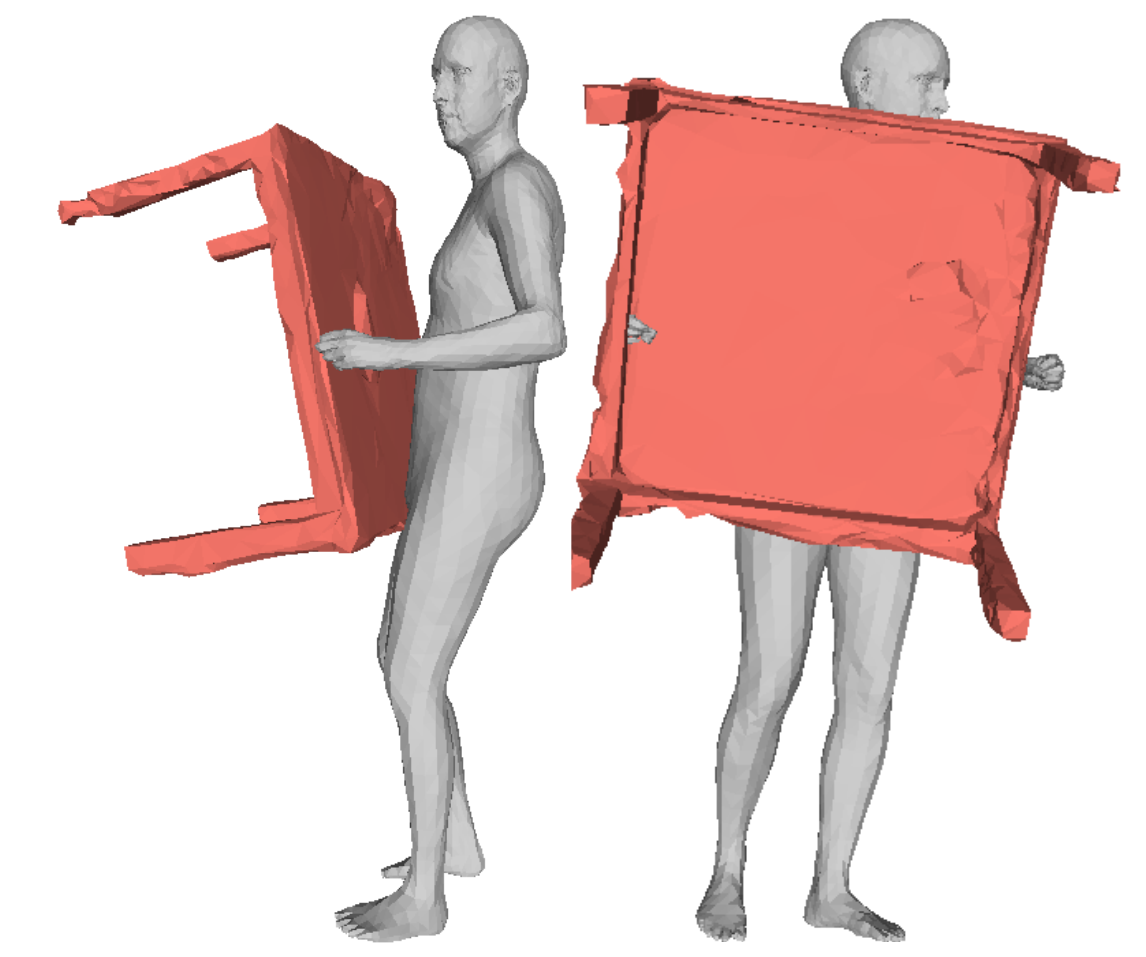} & \includegraphics[height=0.2\textwidth]{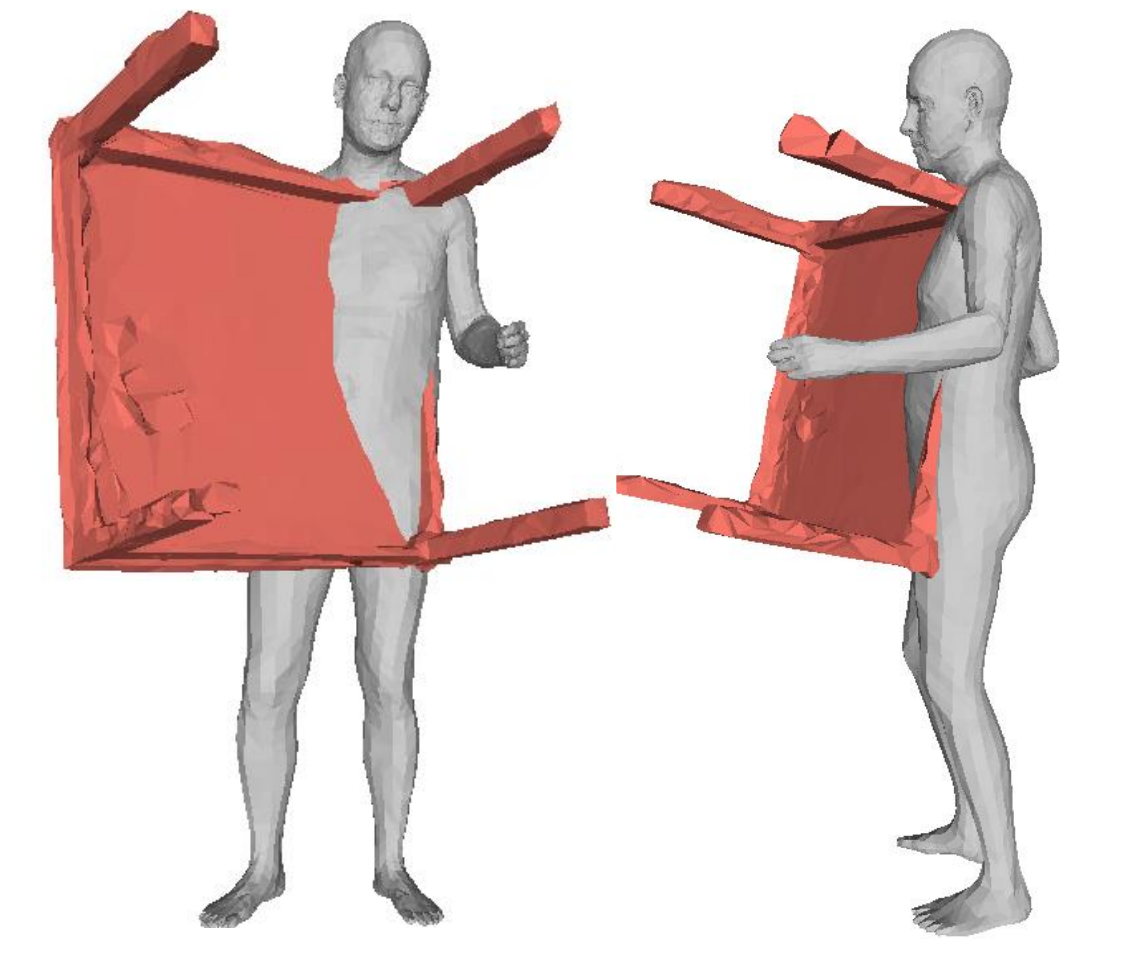} \\
    \includegraphics[height=0.2\textwidth]{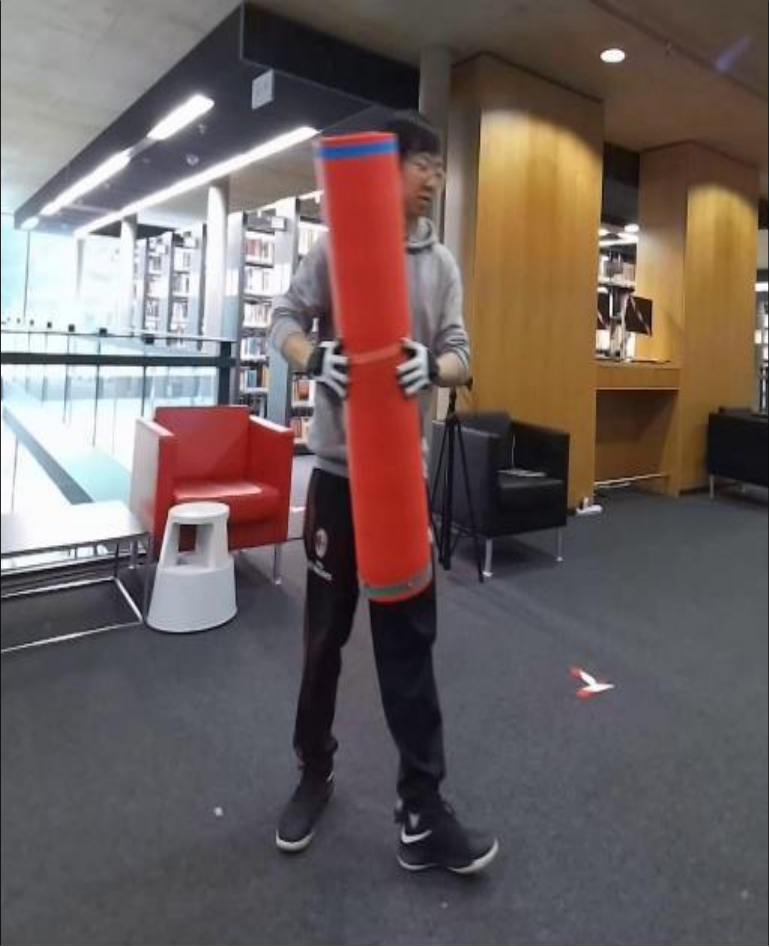}     & \includegraphics[height=0.2\textwidth]{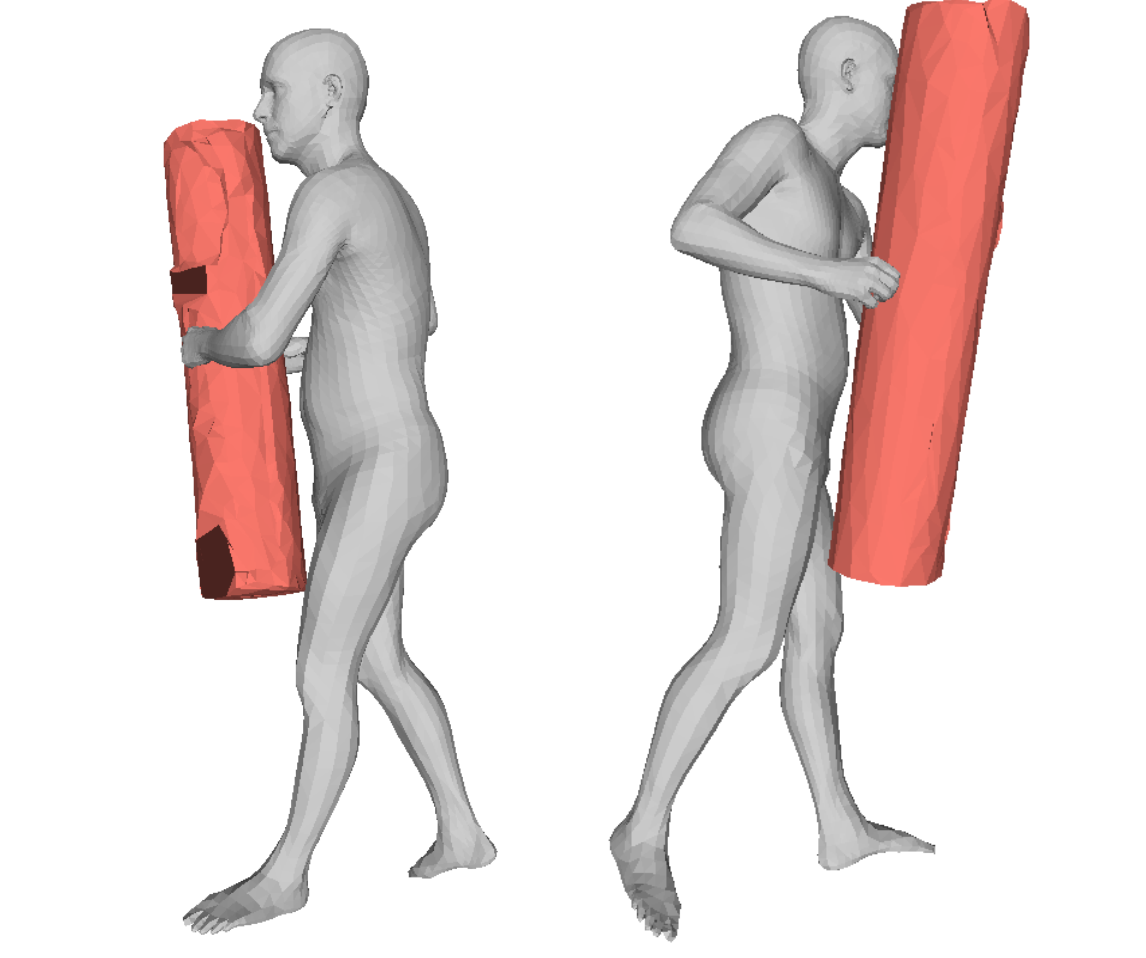} & \includegraphics[height=0.2\textwidth]{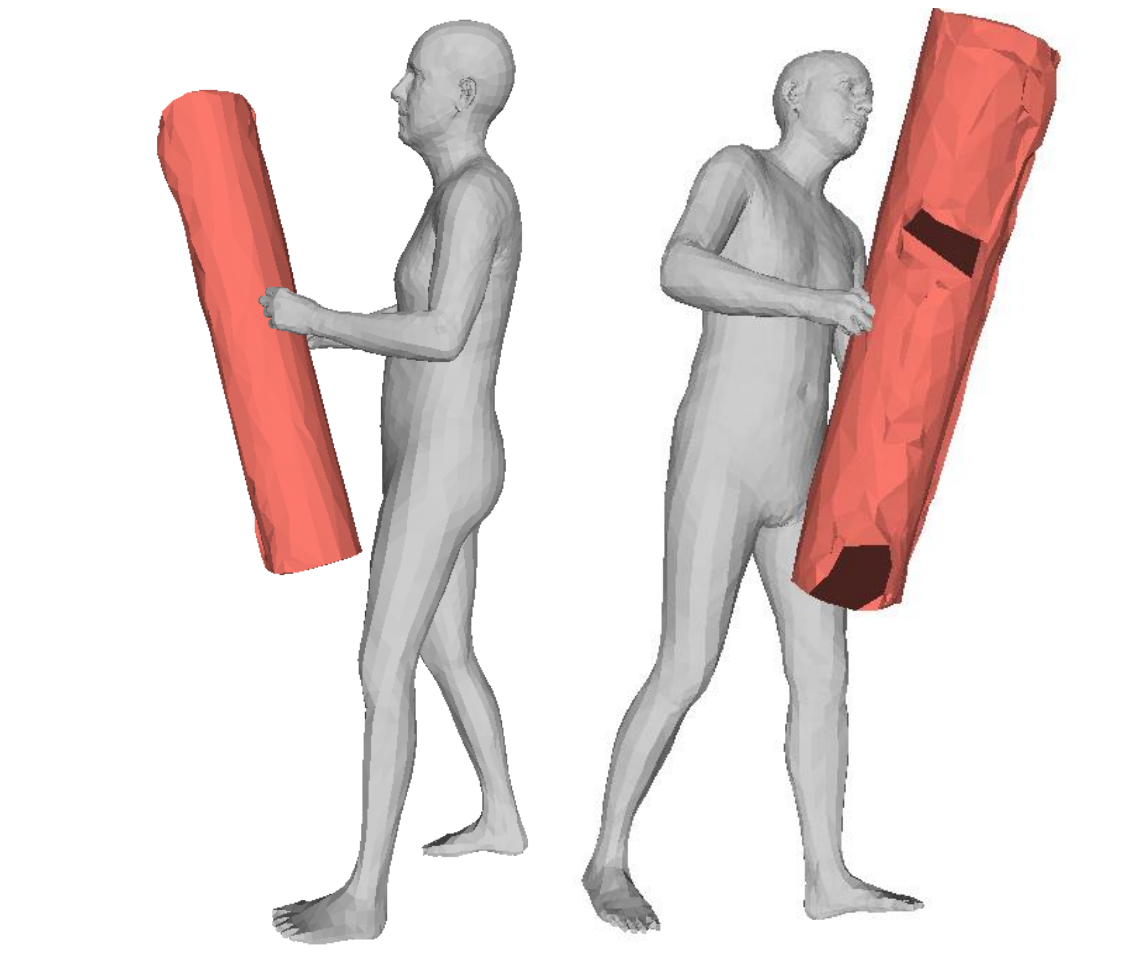} & \includegraphics[height=0.2\textwidth]{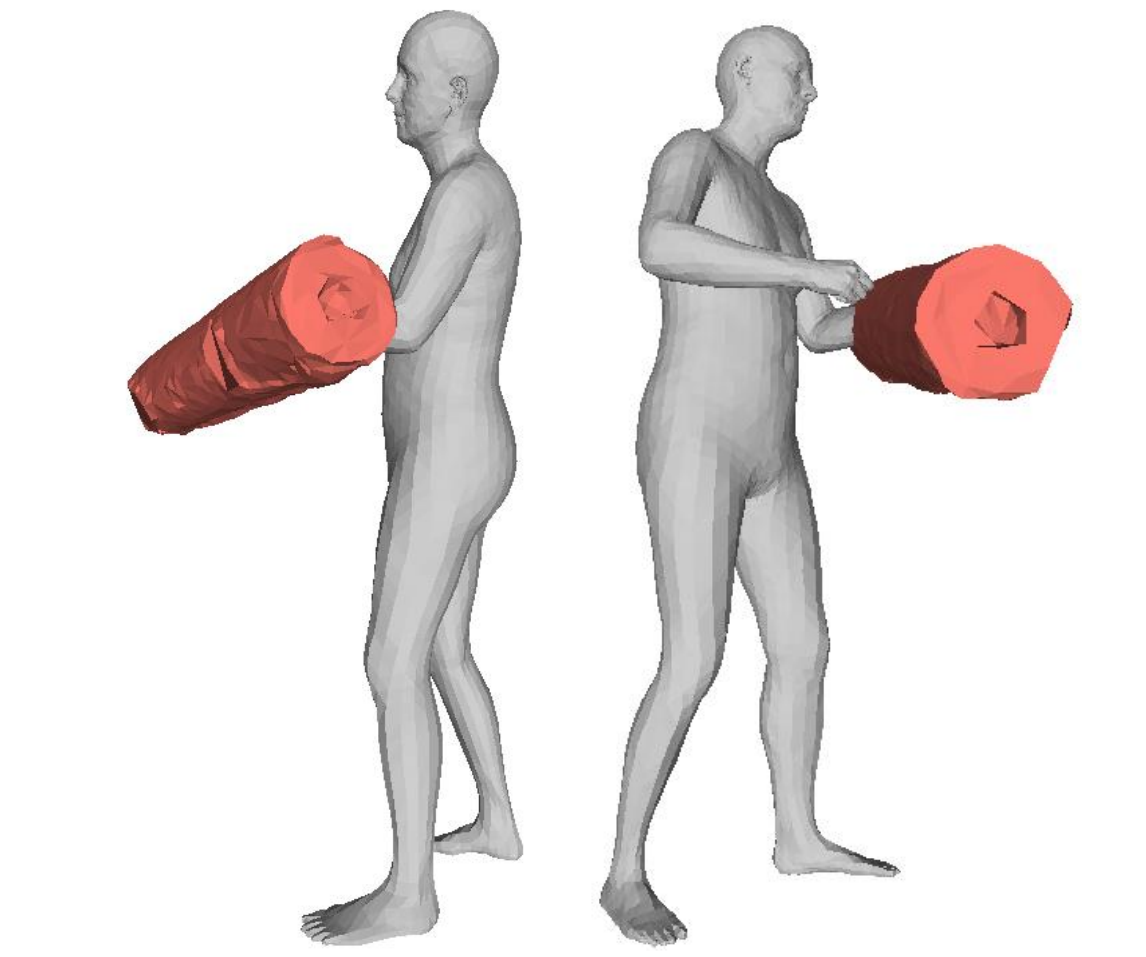} \\
    \includegraphics[height=0.2\textwidth]{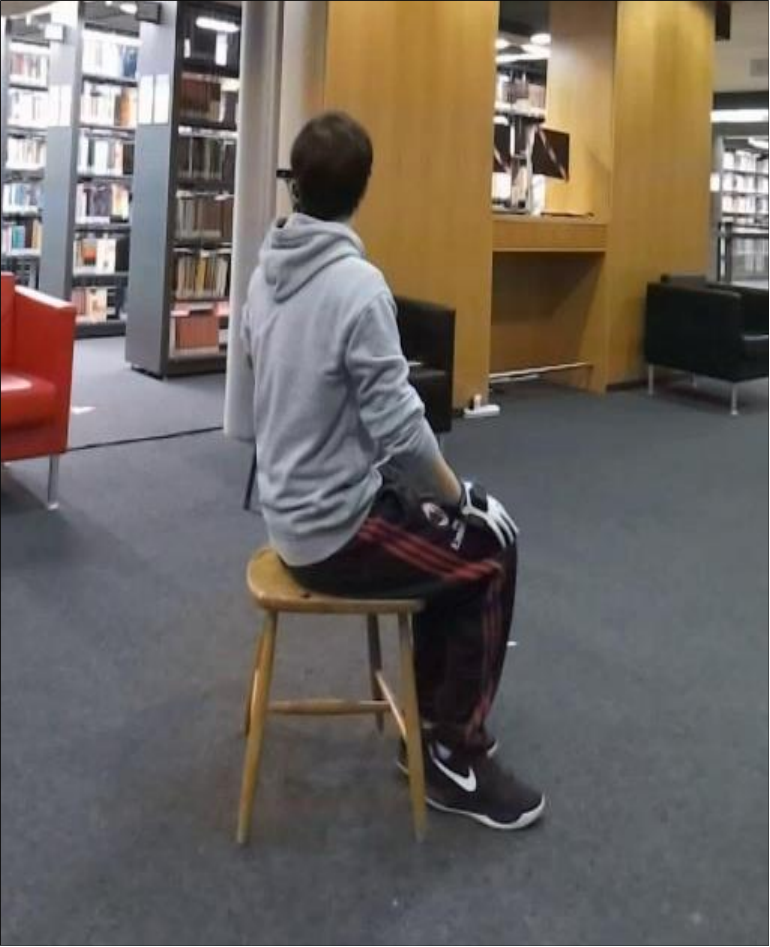}     & \includegraphics[height=0.2\textwidth]{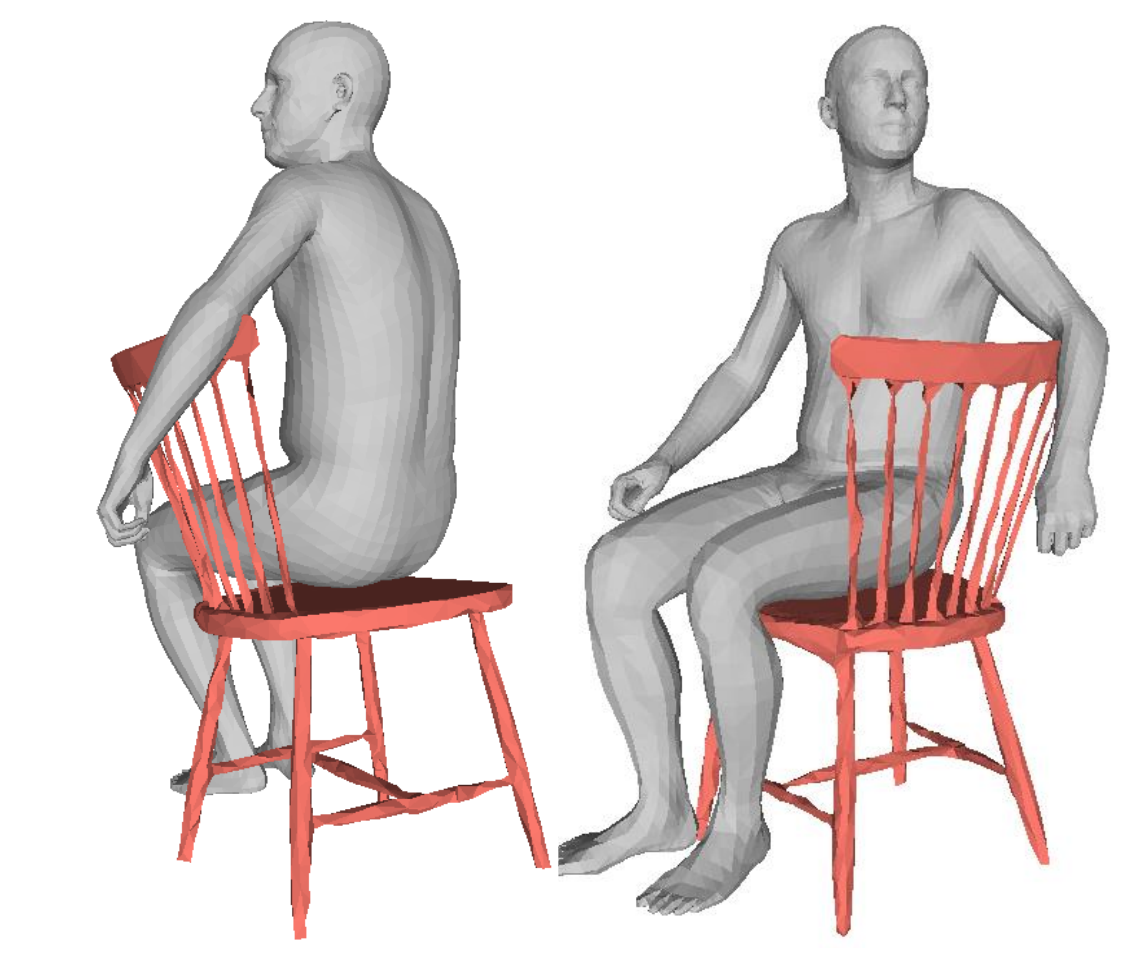} & \includegraphics[height=0.2\textwidth]{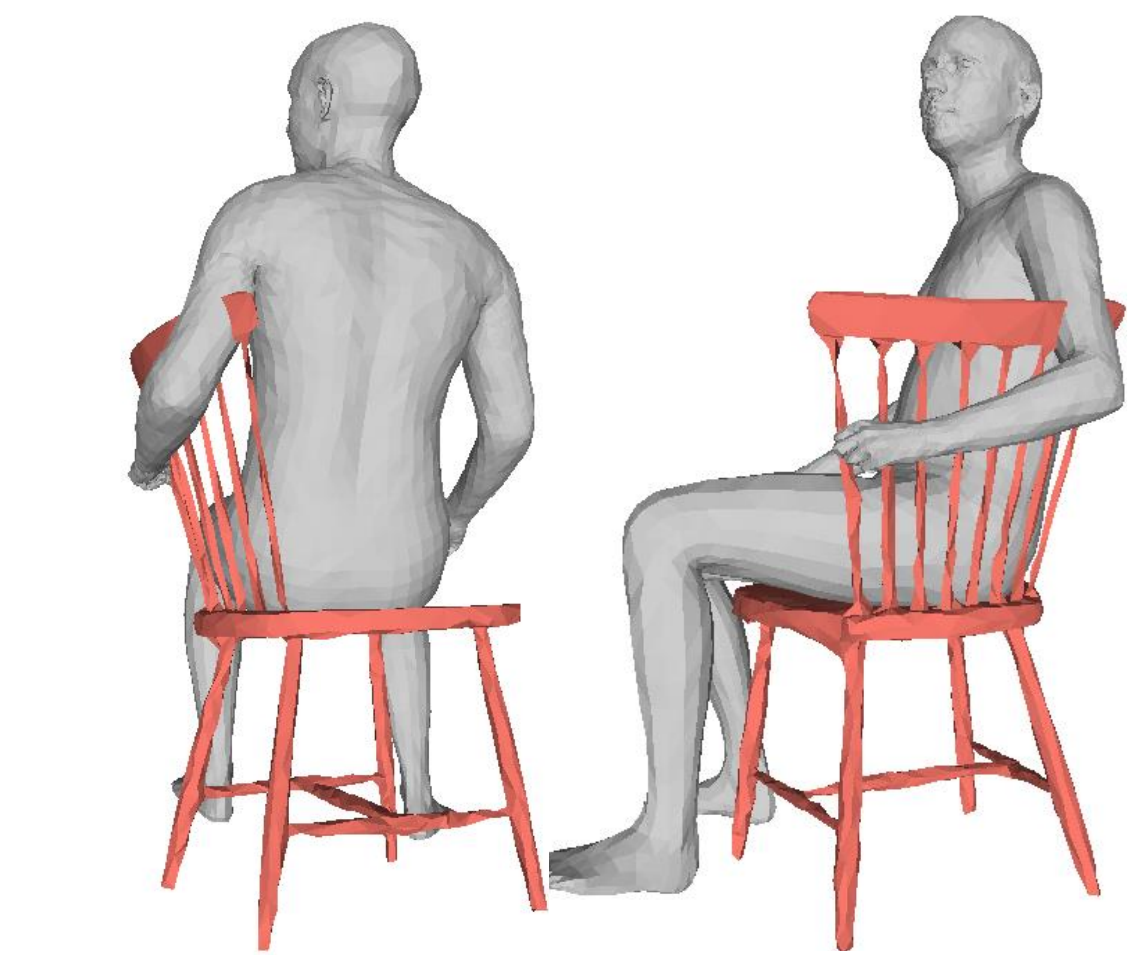} & \includegraphics[height=0.2\textwidth]{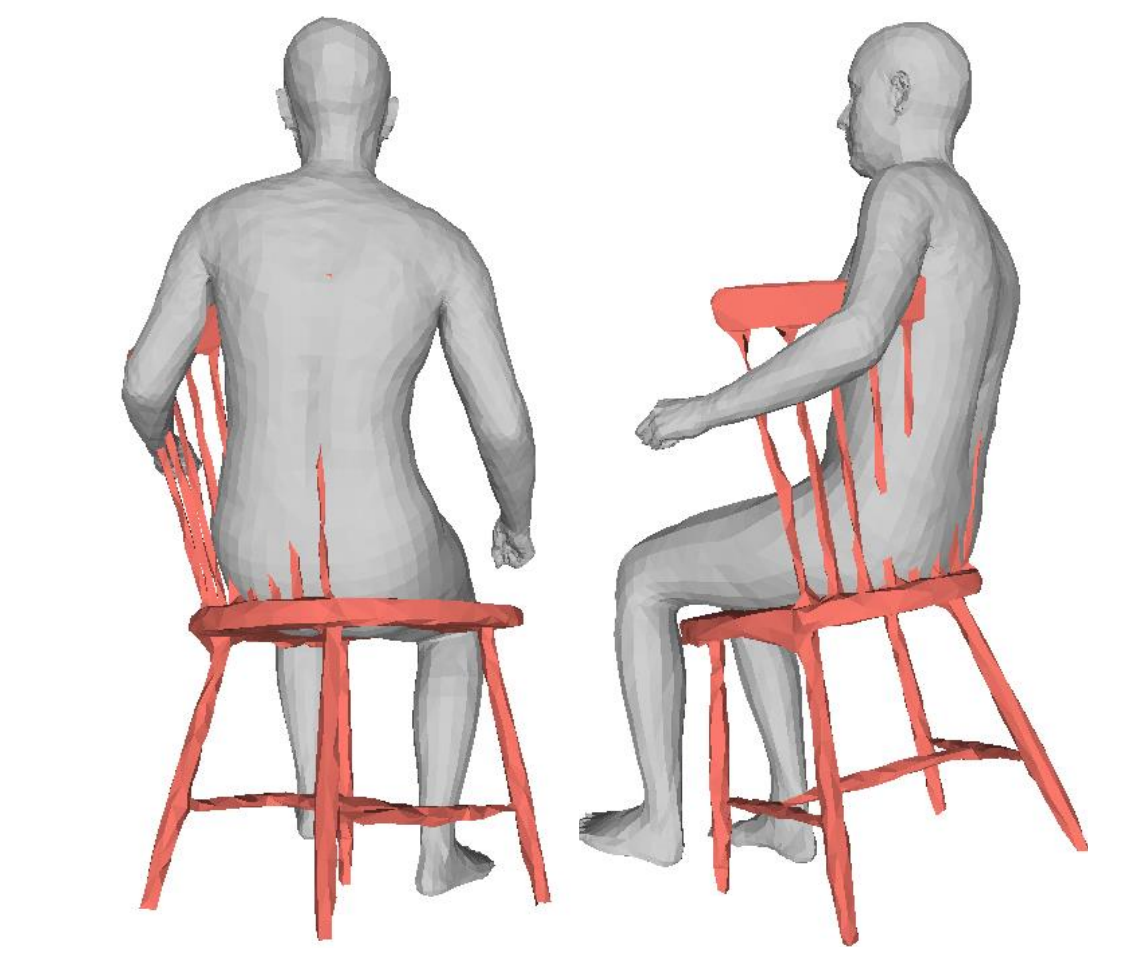} \\
    \includegraphics[height=0.2\textwidth]{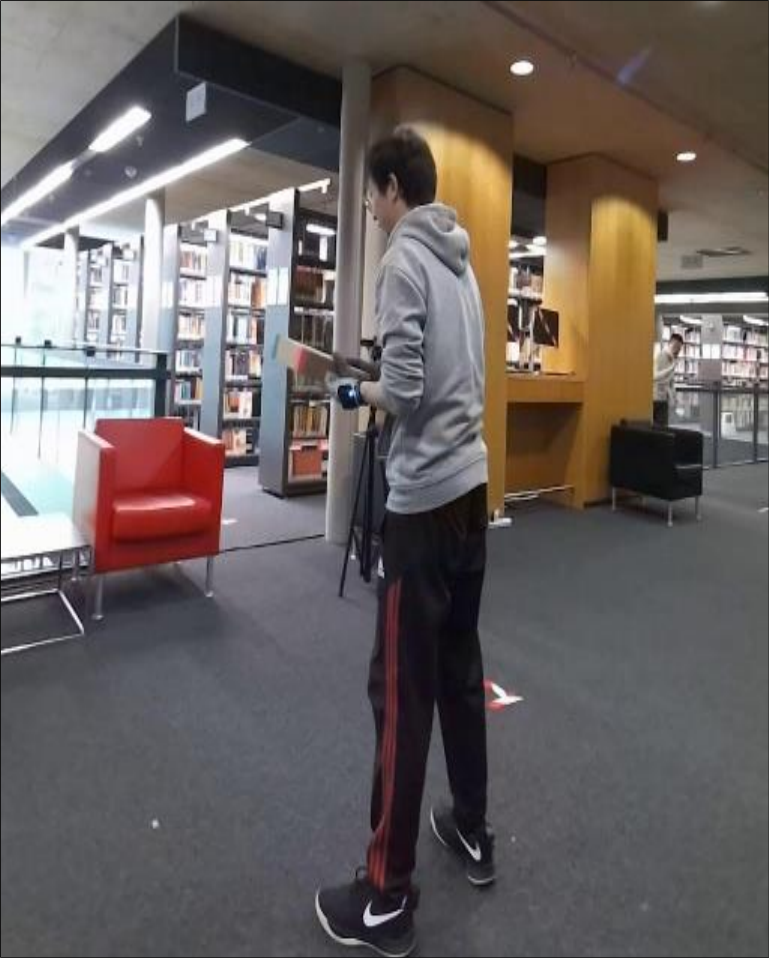}     & \includegraphics[height=0.2\textwidth]{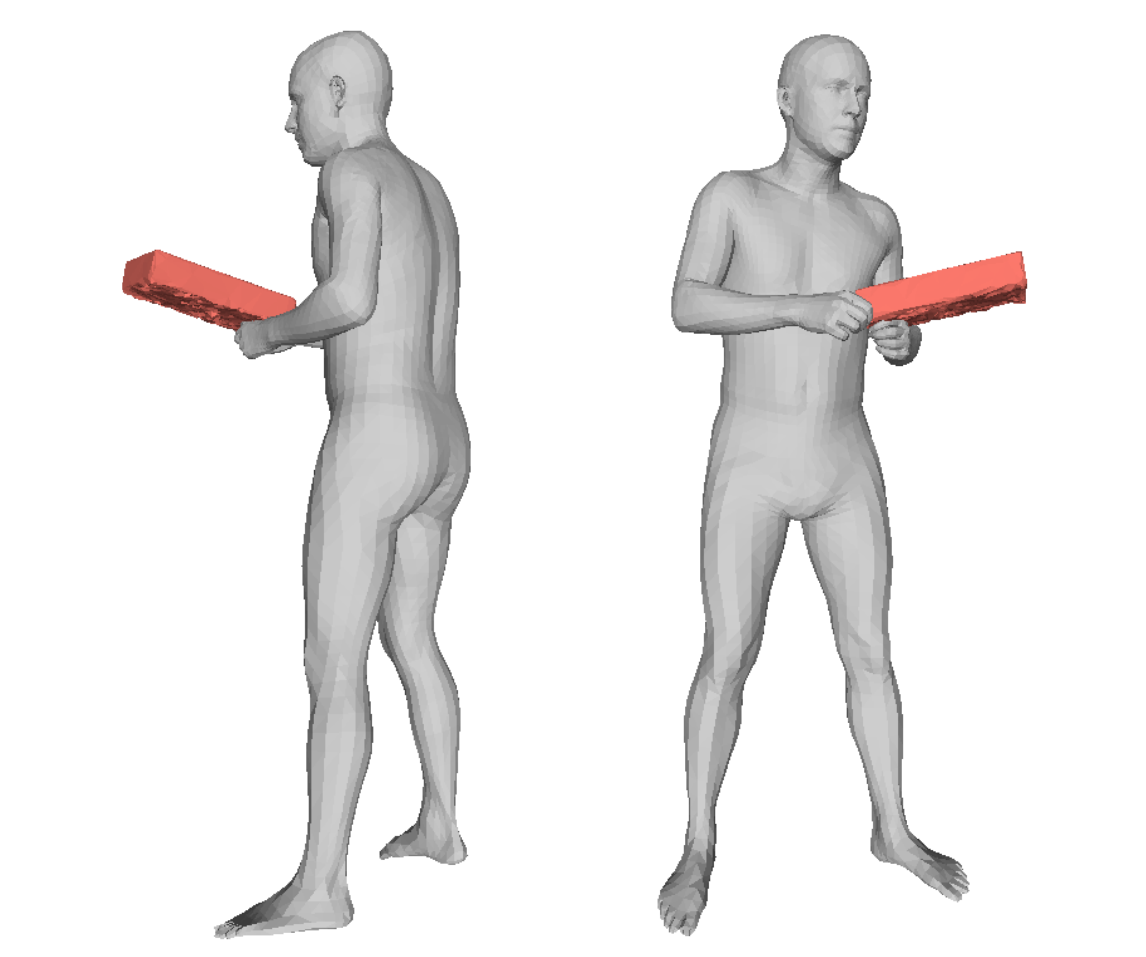} & \includegraphics[height=0.2\textwidth]{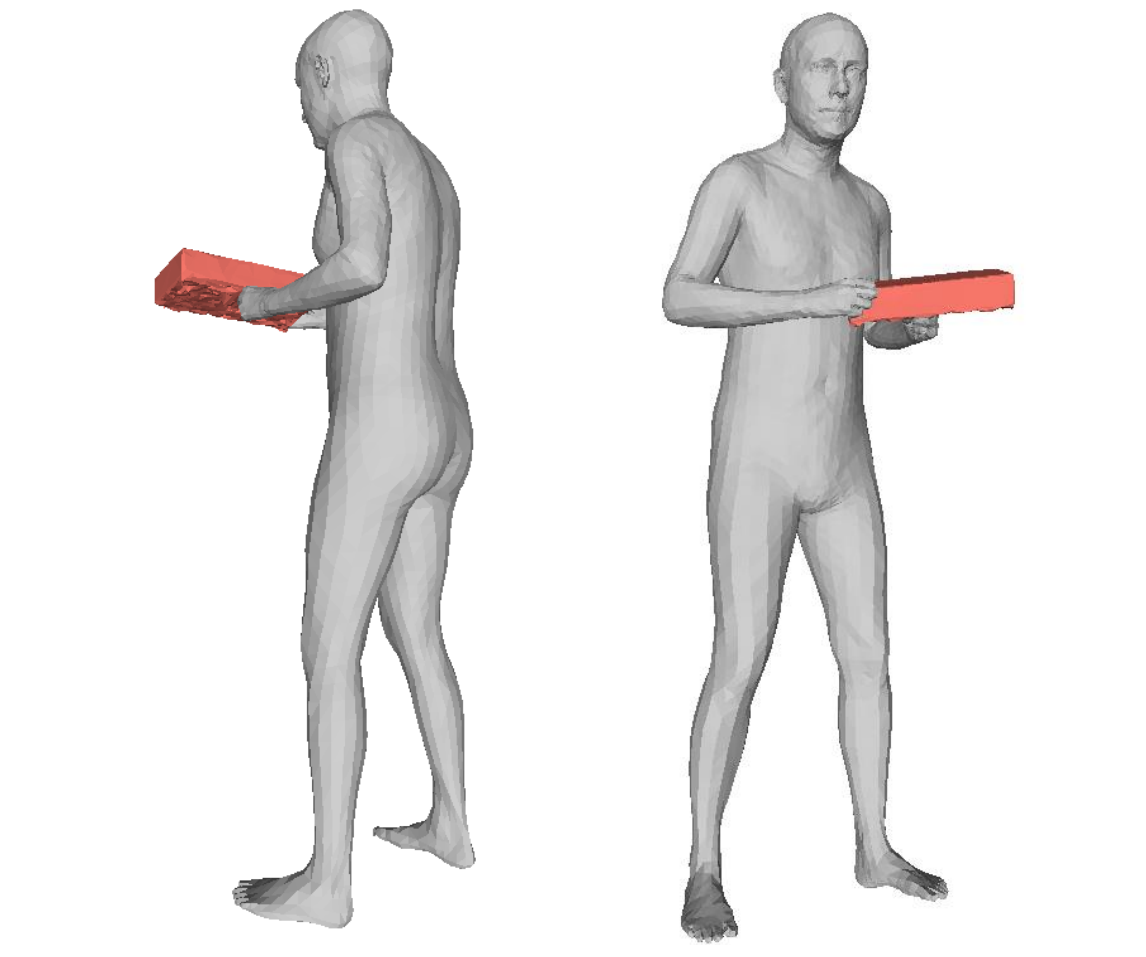} & \includegraphics[height=0.2\textwidth]{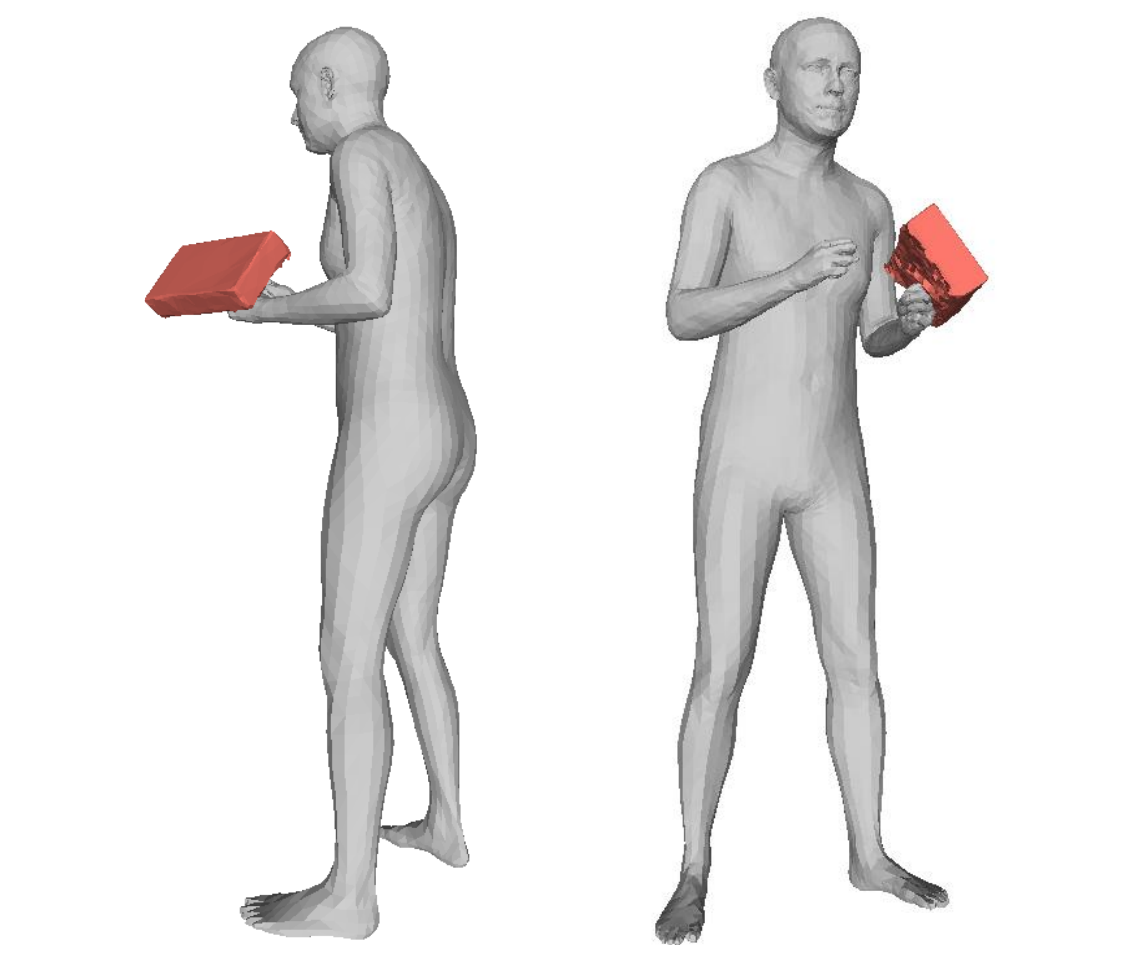} \\
     RGB &  GT & HOI-TG (Ours) & CONTHO
    \end{tabular}
    \caption{Qualitative comparison of 3D human and object reconstruction with CONTHO~\cite{joint} on BEHAVE~\cite{behave} dataset.}
    \label{fig:sullpybehave}
\end{figure*}

\begin{figure*}
    \centering
    \begin{tabular}{cccc}
    \includegraphics[height=0.2\textwidth]{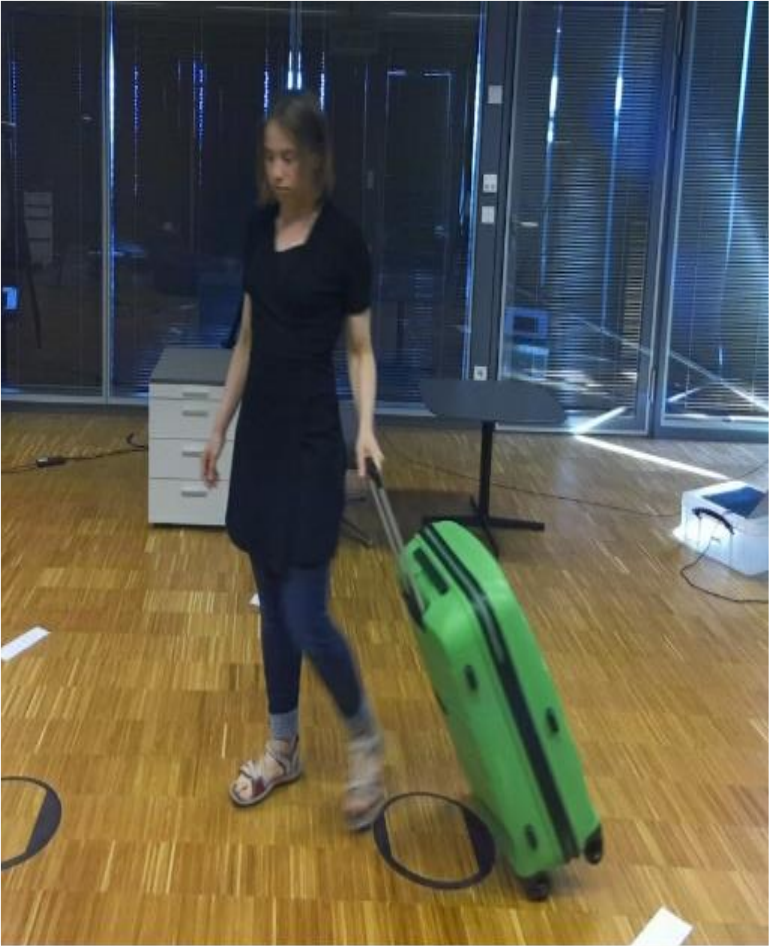}     & \includegraphics[height=0.2\textwidth]{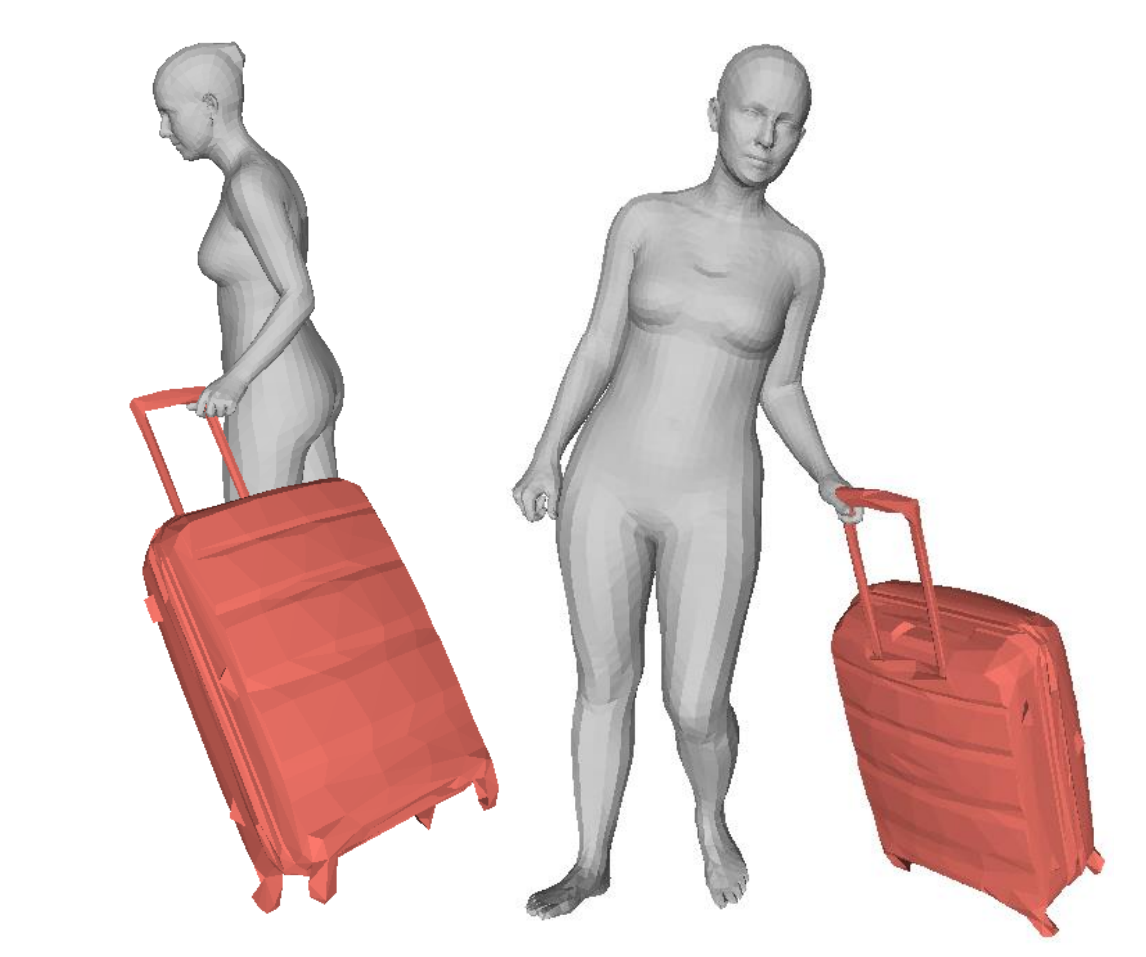} & \includegraphics[height=0.2\textwidth]{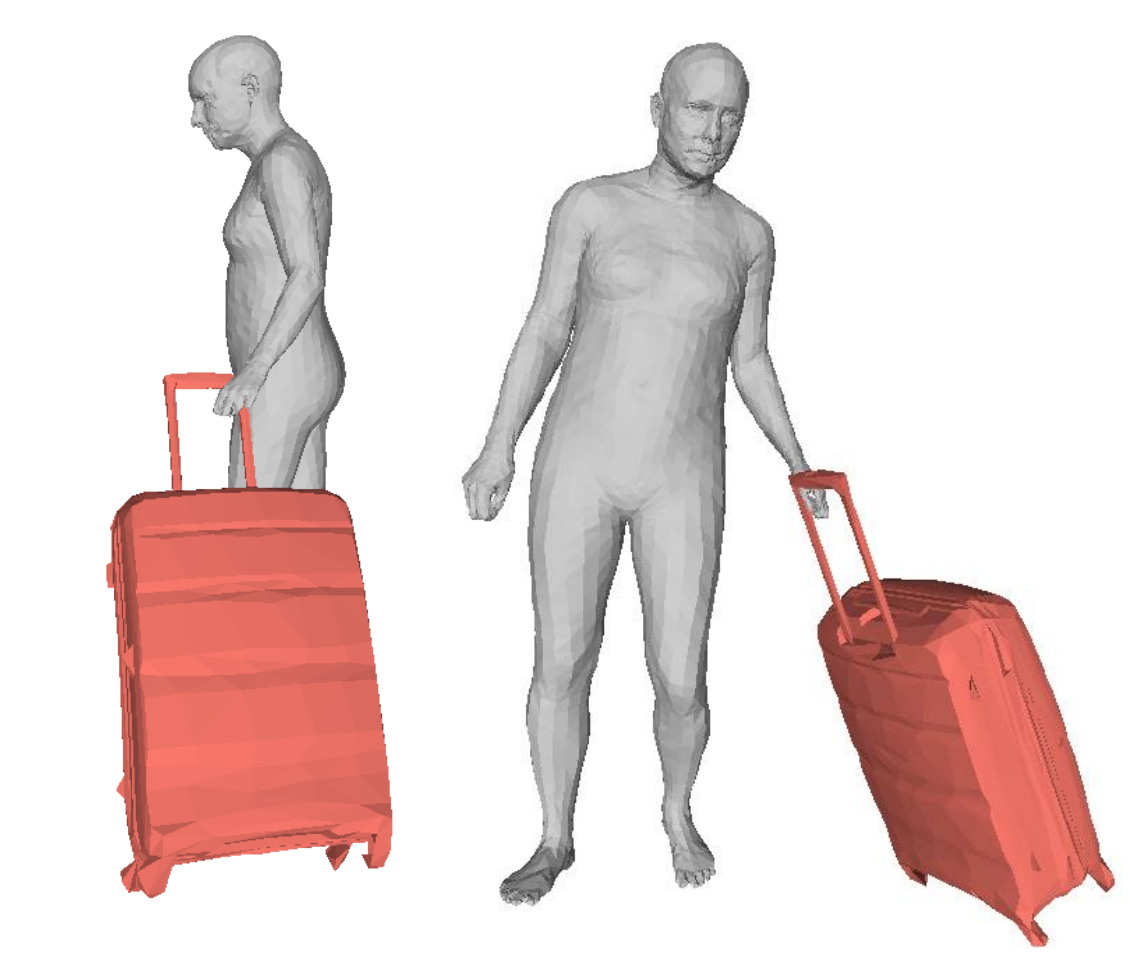} & \includegraphics[height=0.2\textwidth]{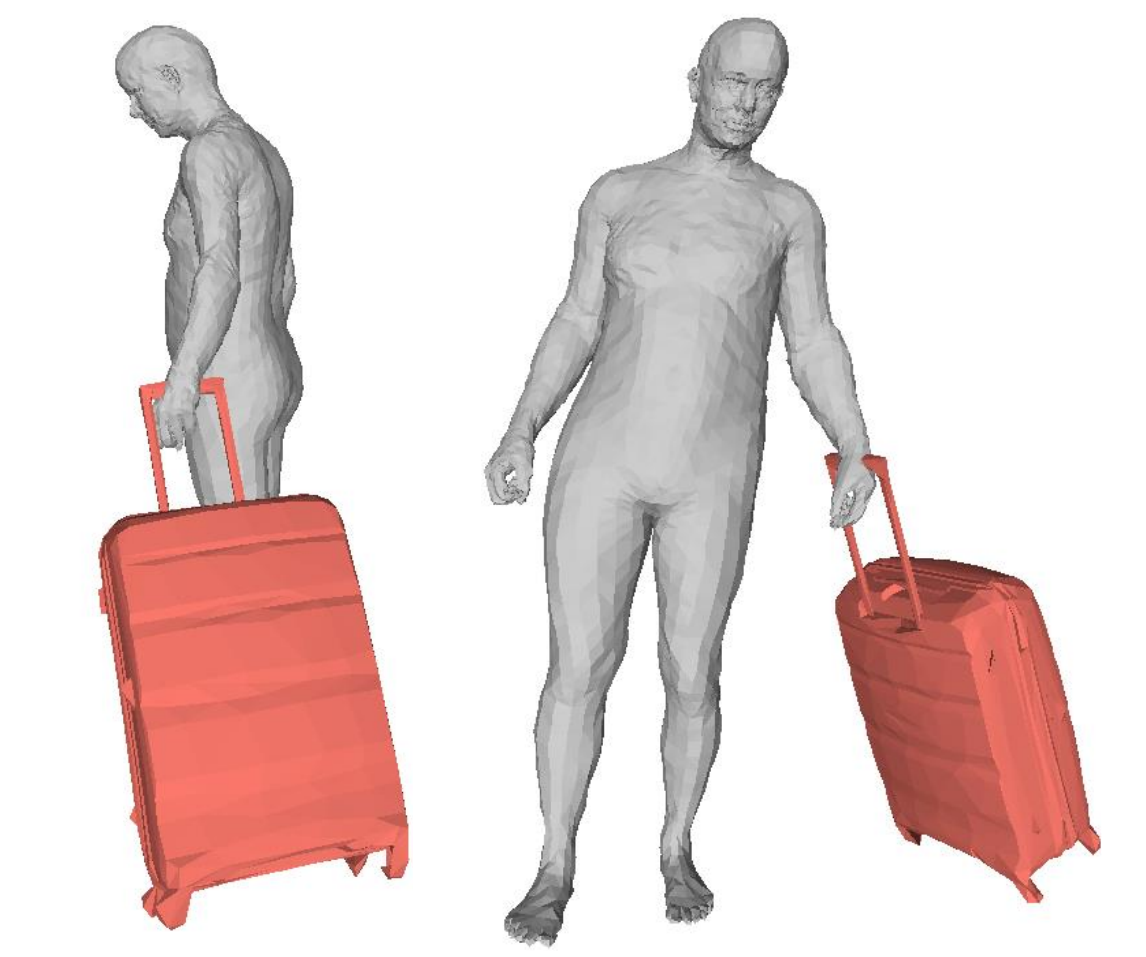} \\
    \includegraphics[height=0.2\textwidth]{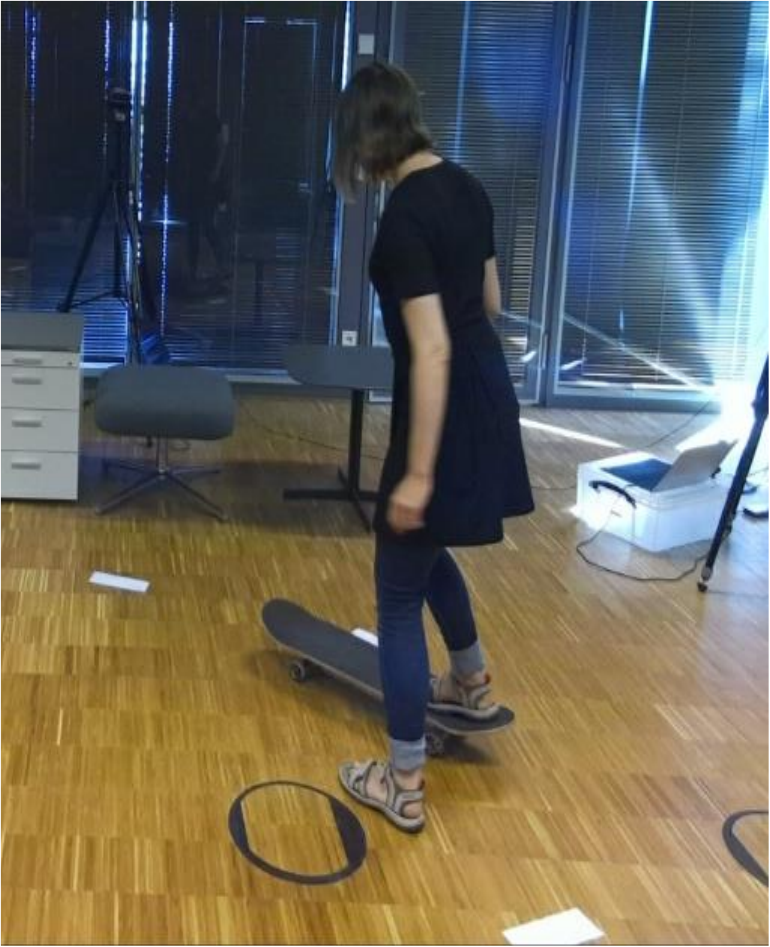}     & \includegraphics[height=0.2\textwidth]{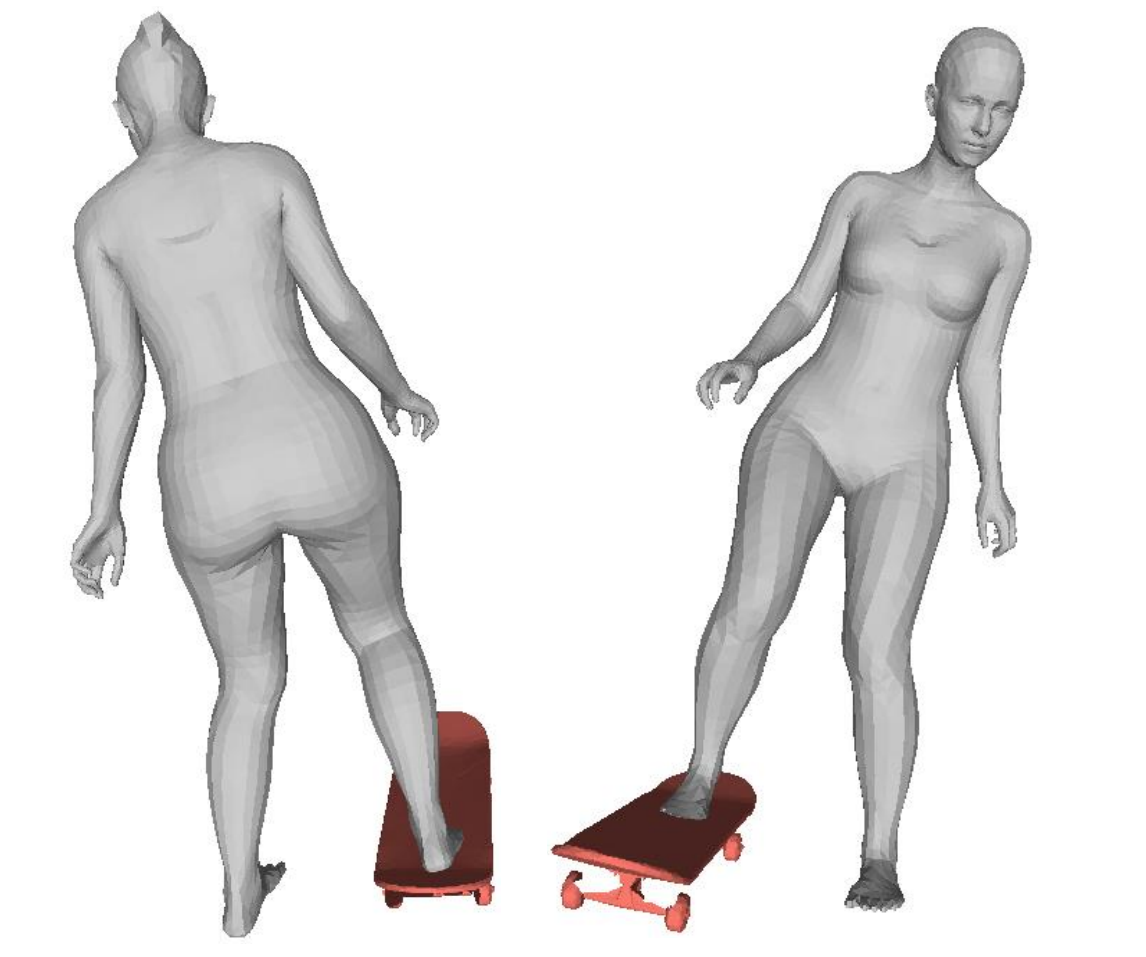} & \includegraphics[height=0.2\textwidth]{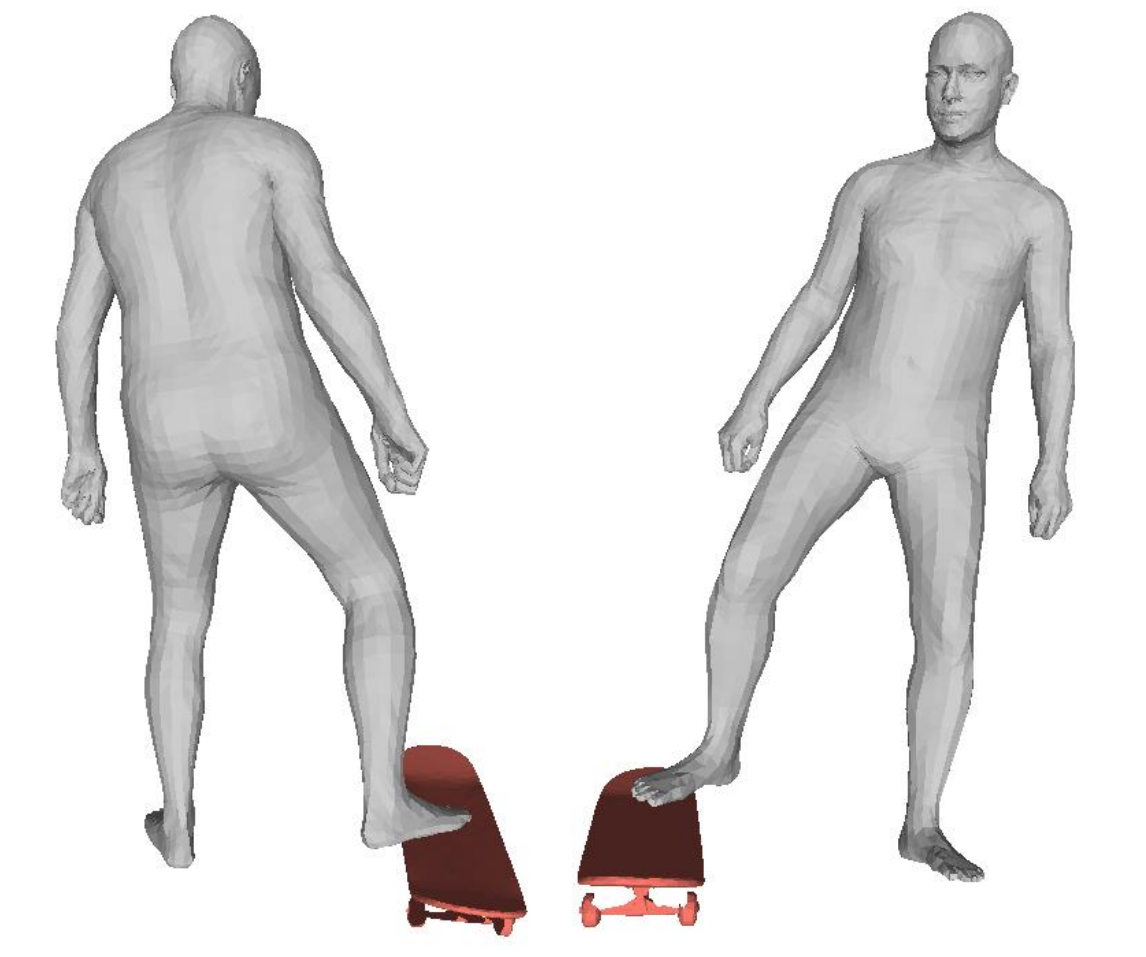} & \includegraphics[height=0.2\textwidth]{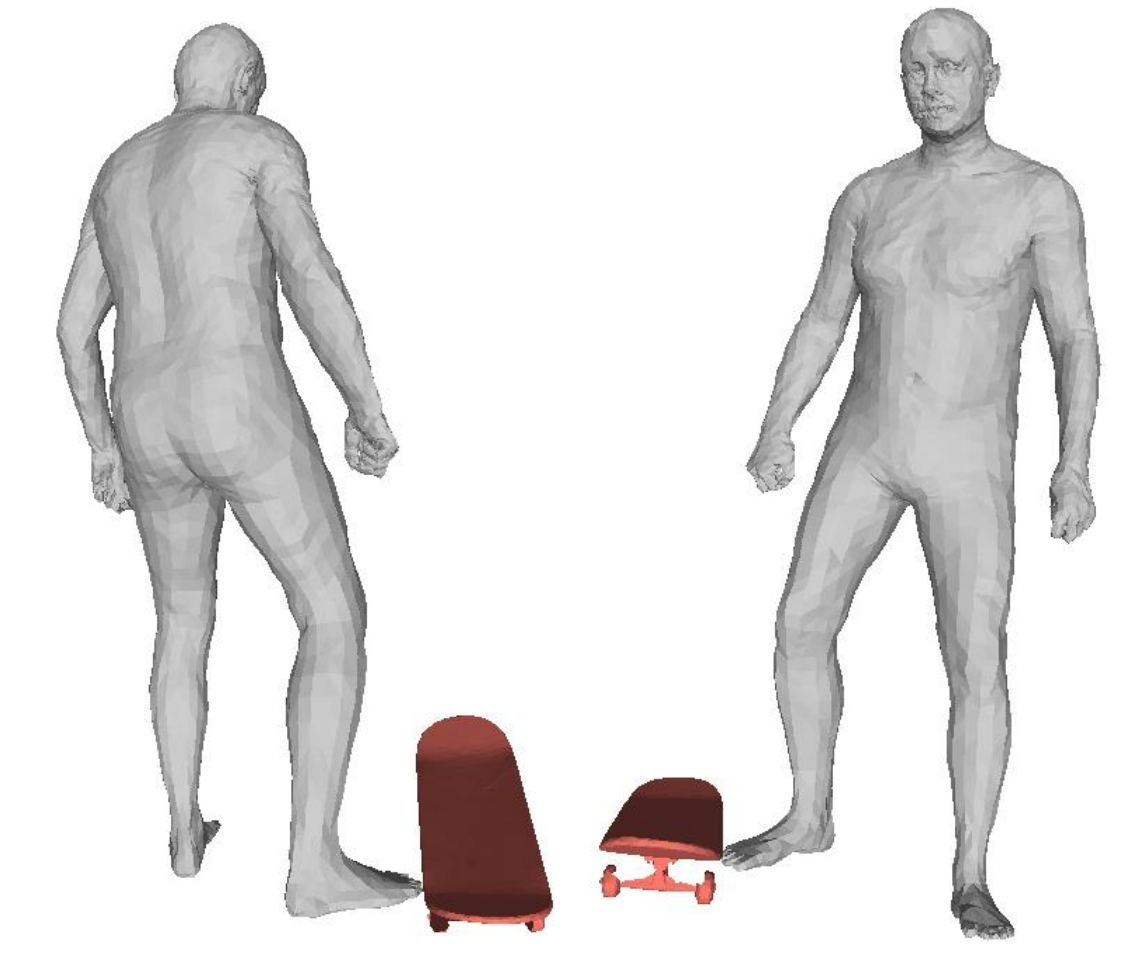} \\
    \includegraphics[height=0.2\textwidth]{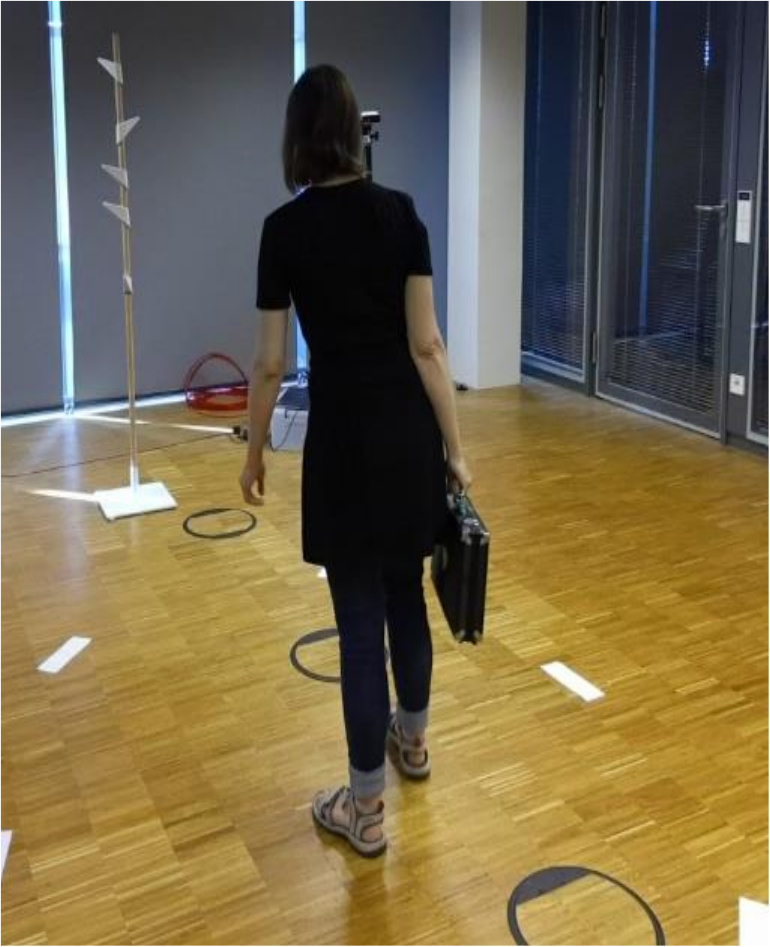}     & \includegraphics[height=0.2\textwidth]{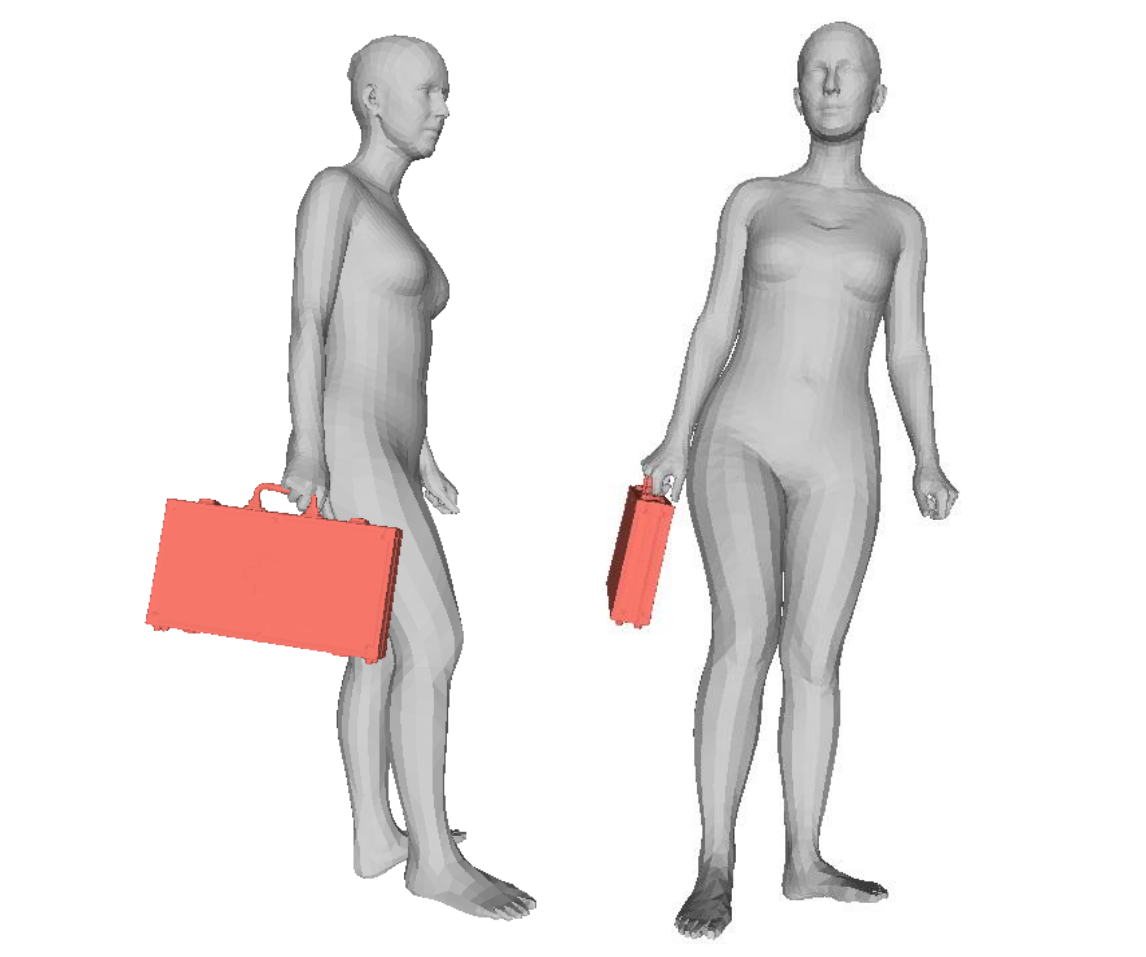} & \includegraphics[height=0.2\textwidth]{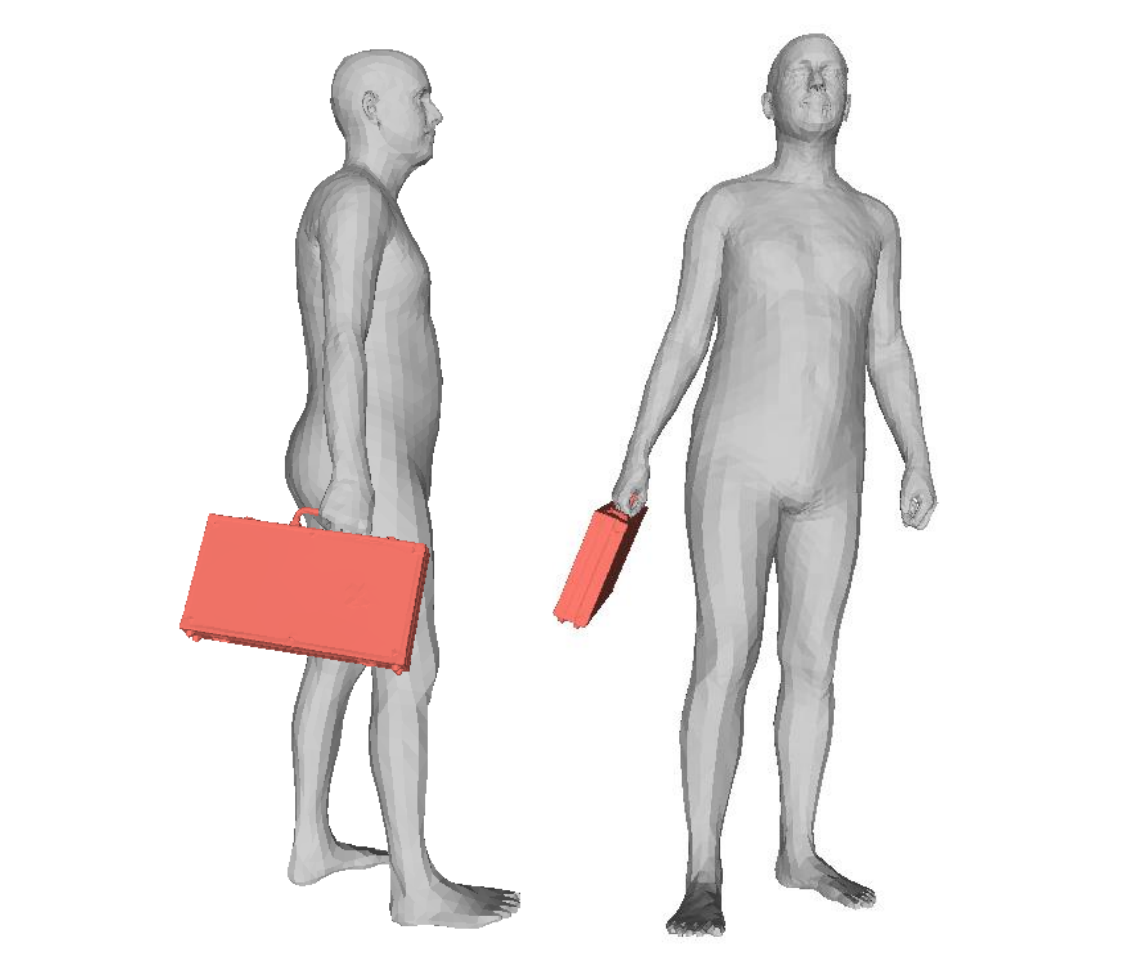} & \includegraphics[height=0.2\textwidth]{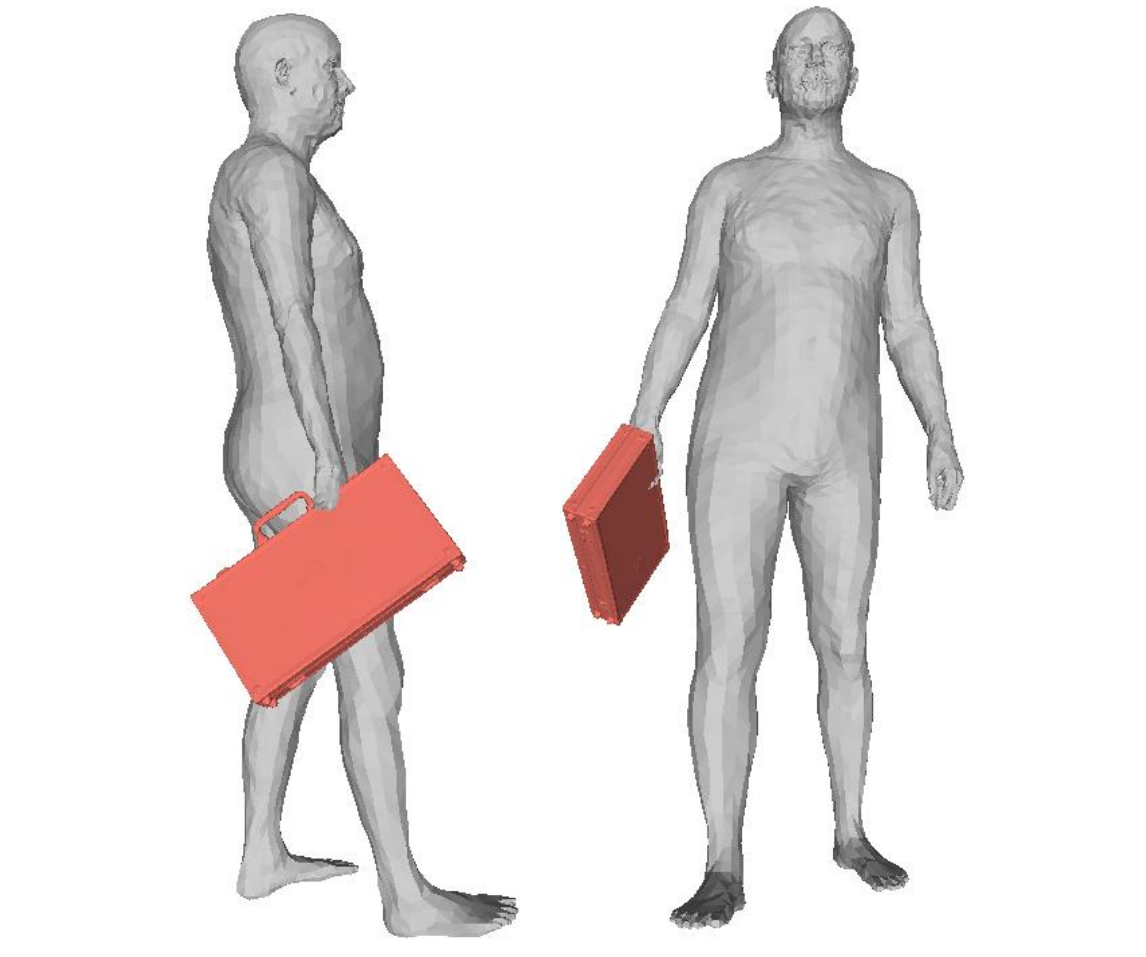} \\
    \includegraphics[height=0.2\textwidth]{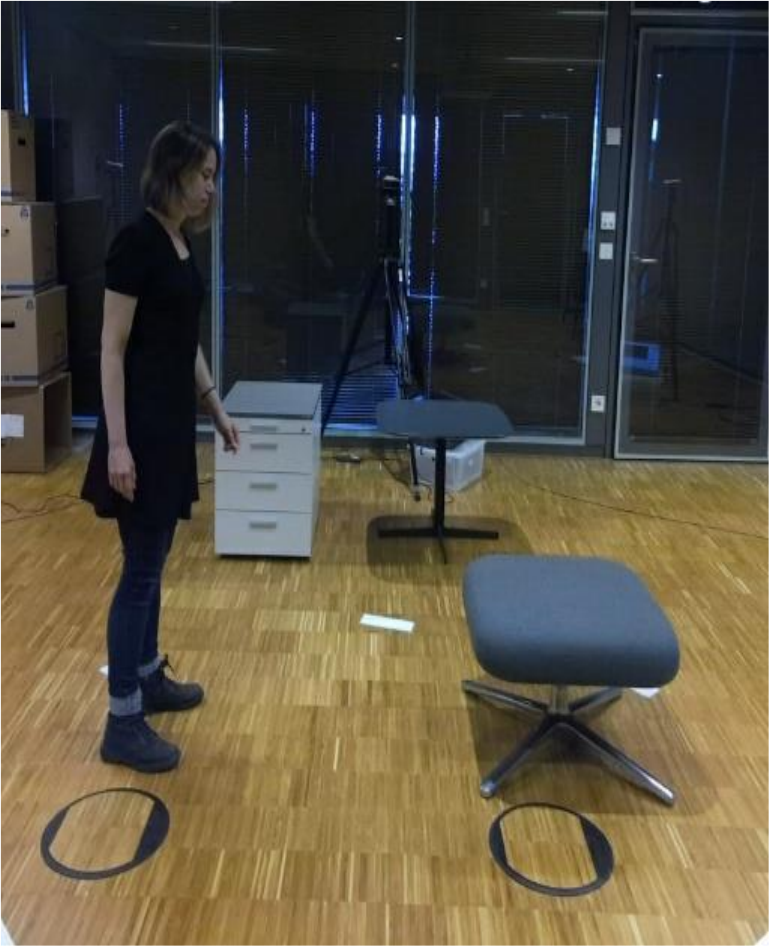}     & \includegraphics[height=0.2\textwidth]{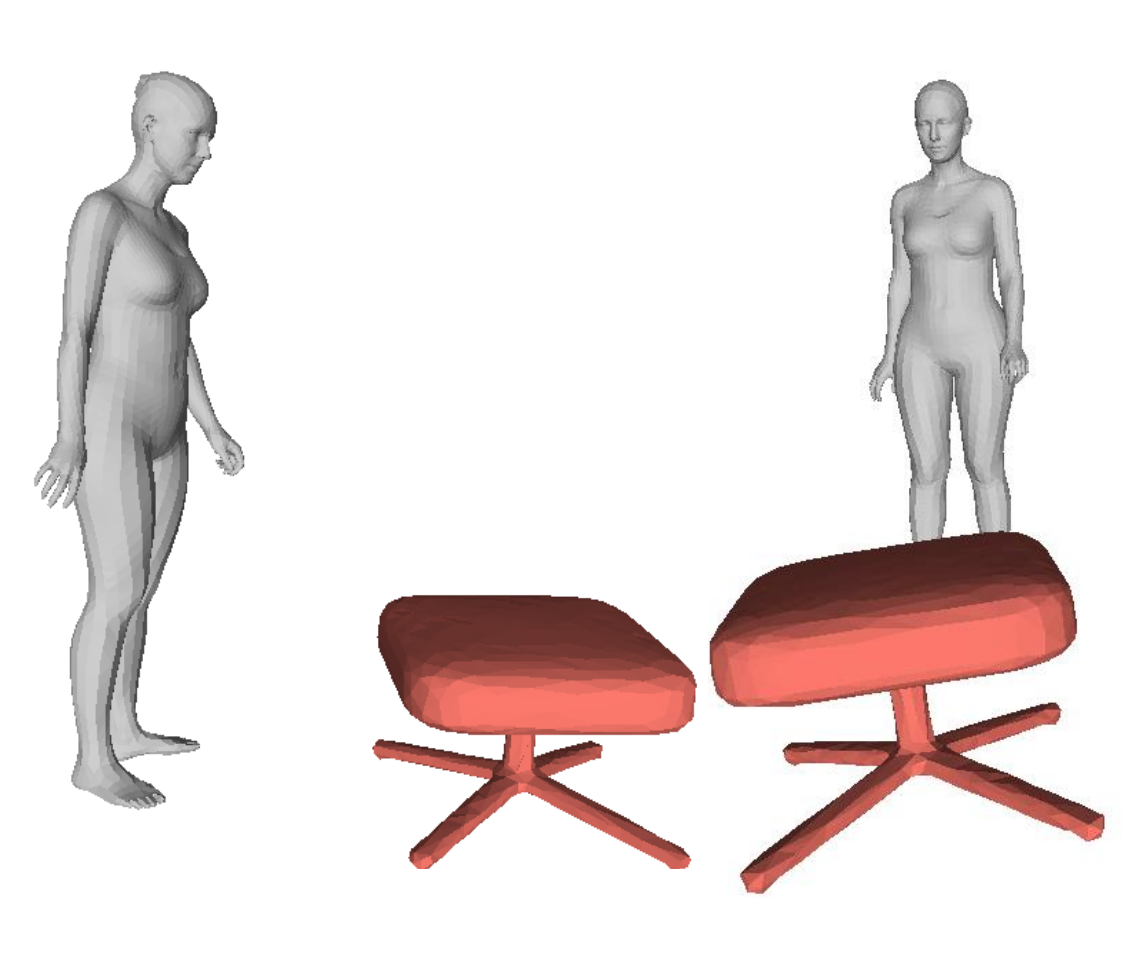} & \includegraphics[height=0.2\textwidth]{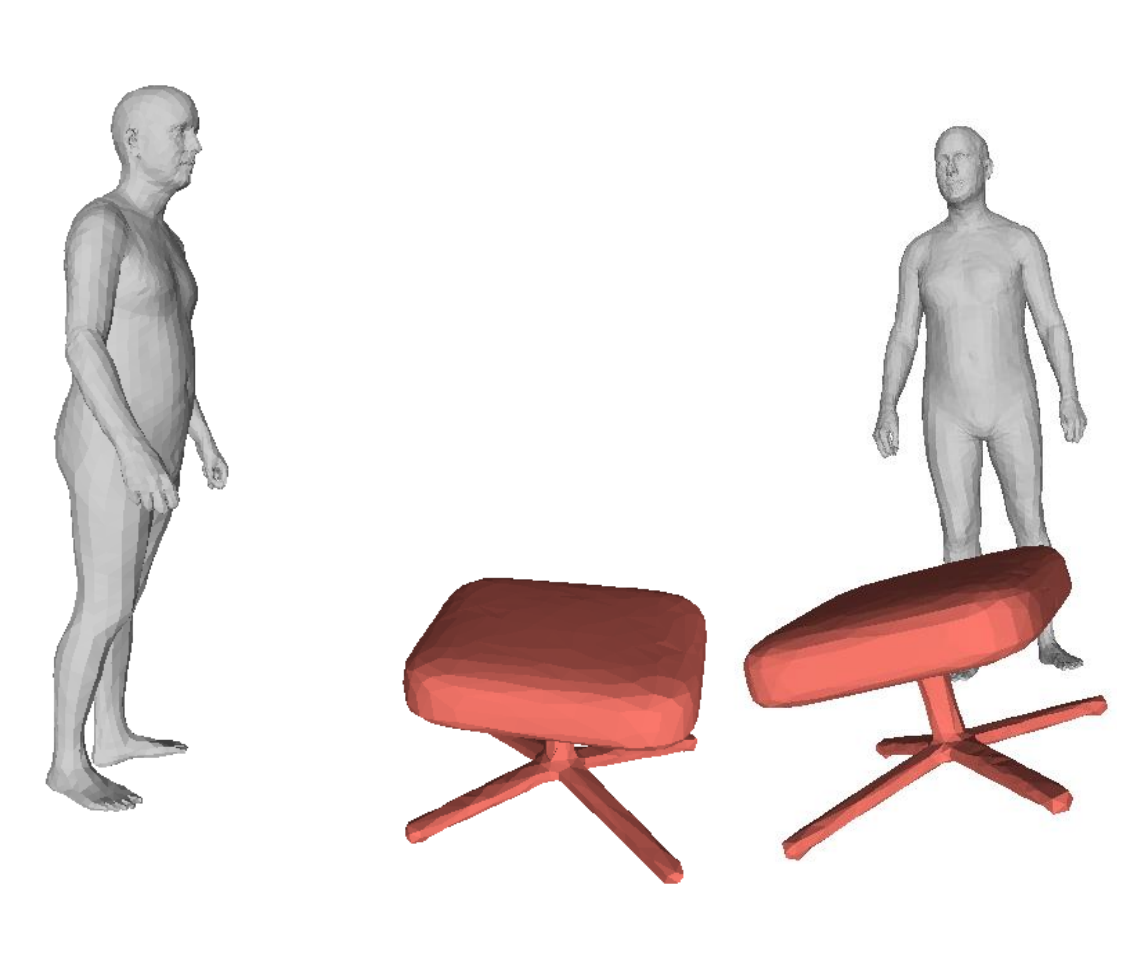} & \includegraphics[height=0.2\textwidth]{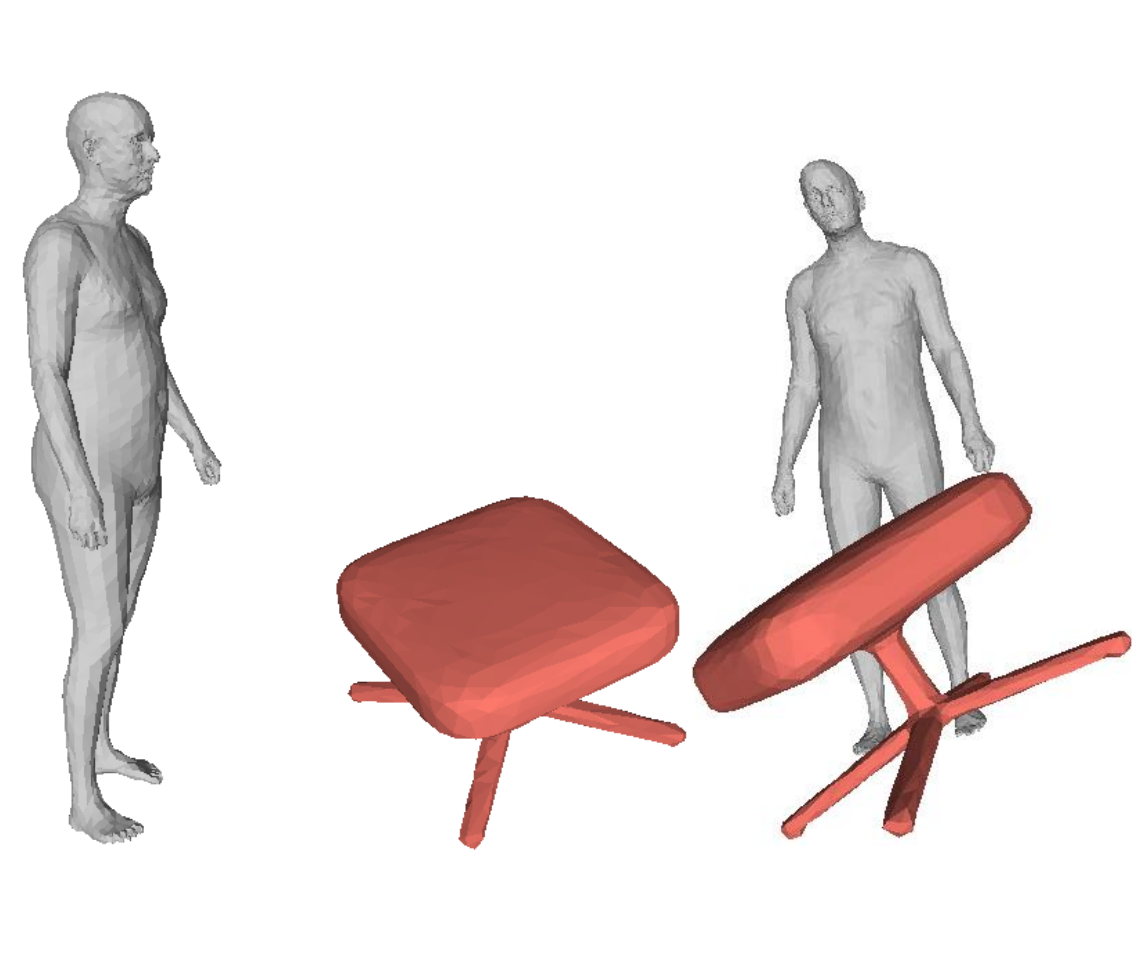} \\
    \includegraphics[height=0.2\textwidth]{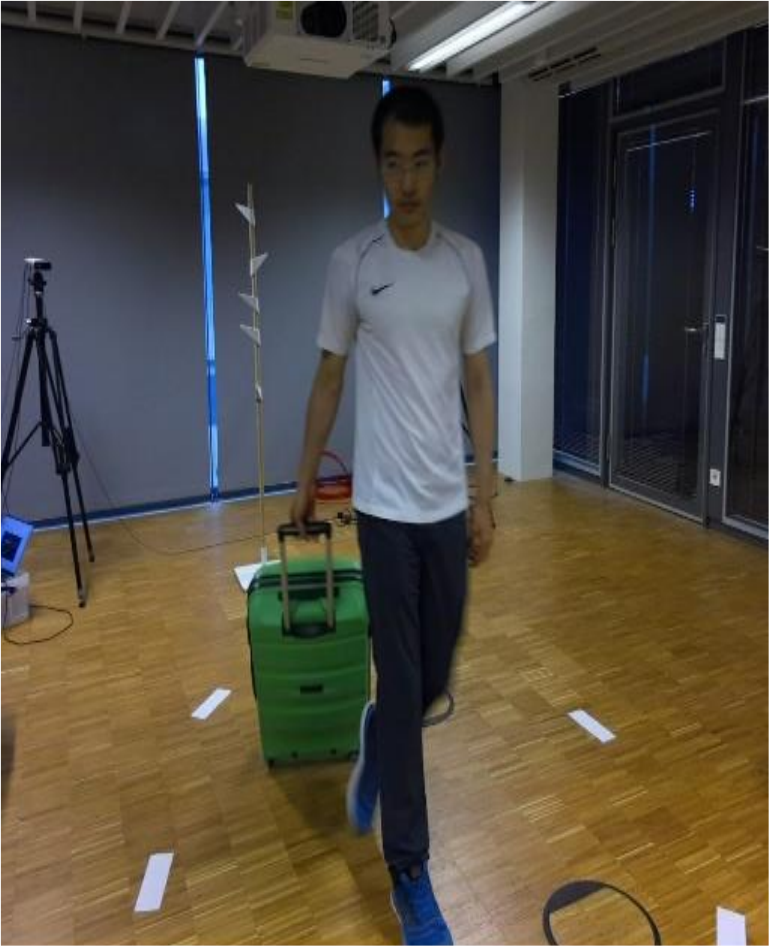}     & \includegraphics[height=0.2\textwidth]{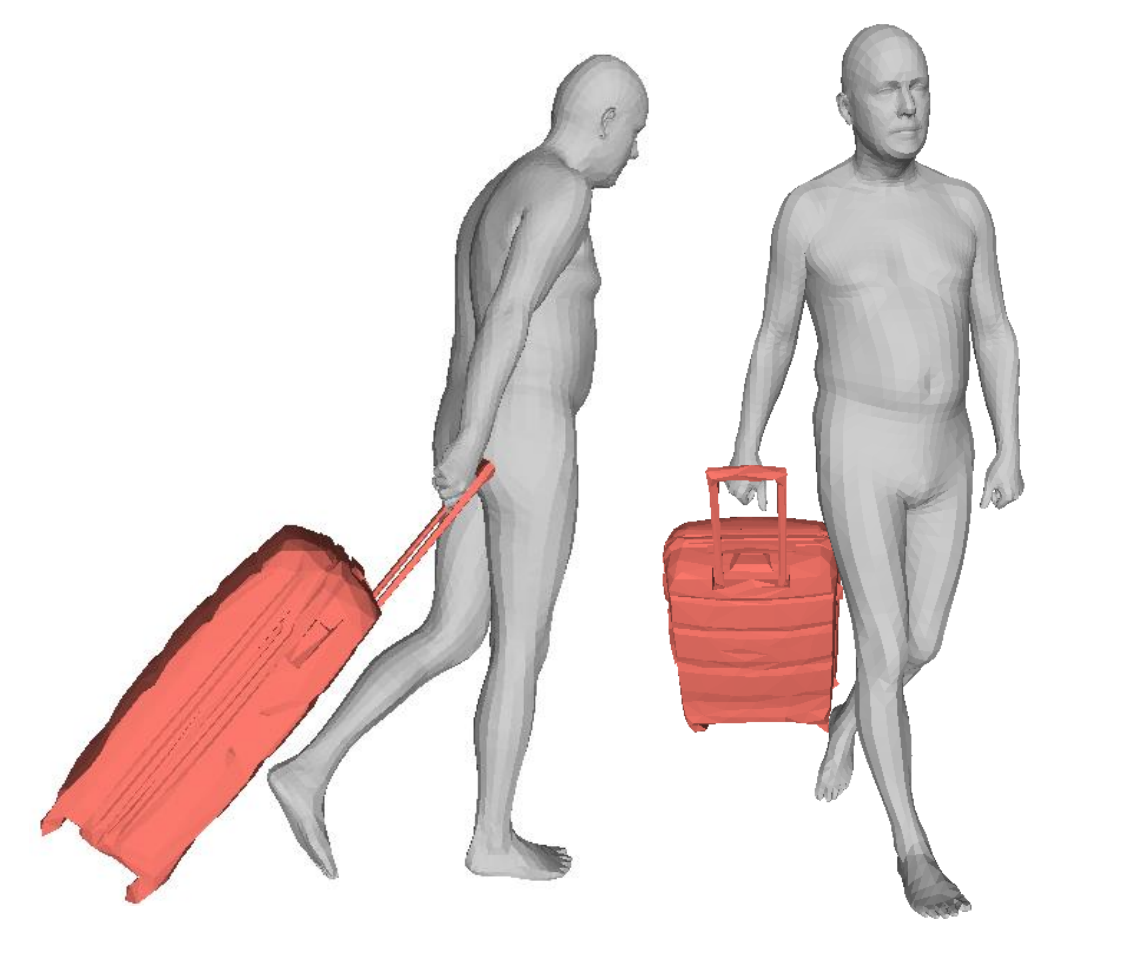} & \includegraphics[height=0.2\textwidth]{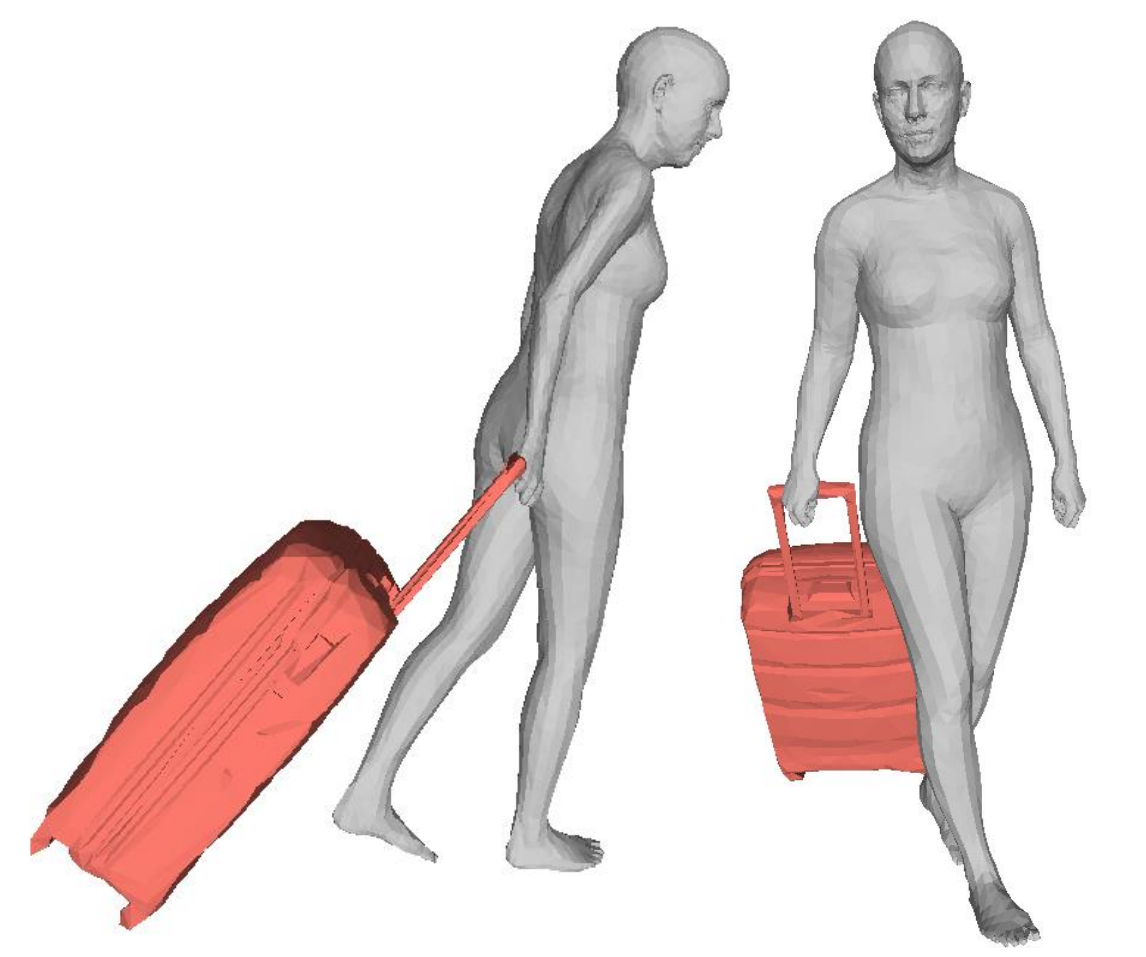} & \includegraphics[height=0.2\textwidth]{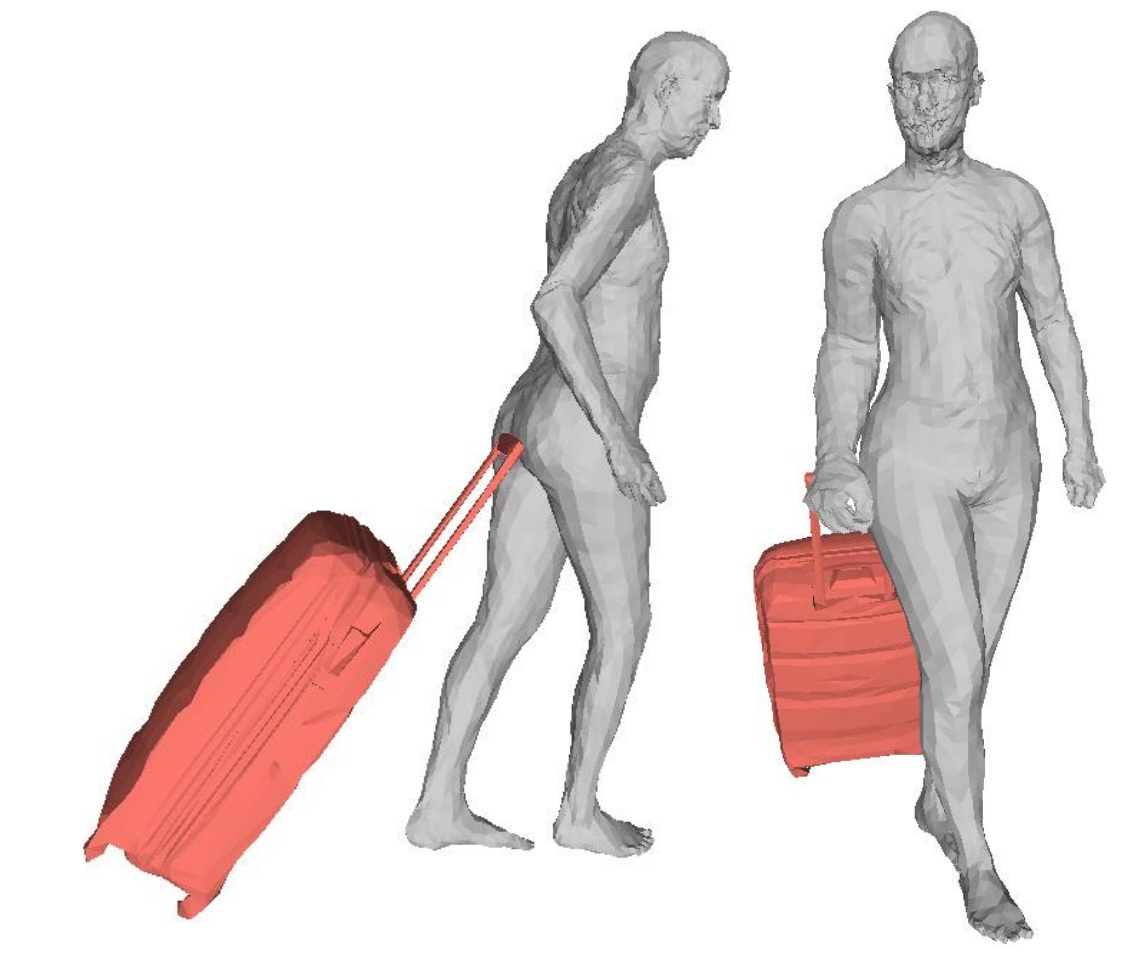} \\
    \includegraphics[height=0.2\textwidth]{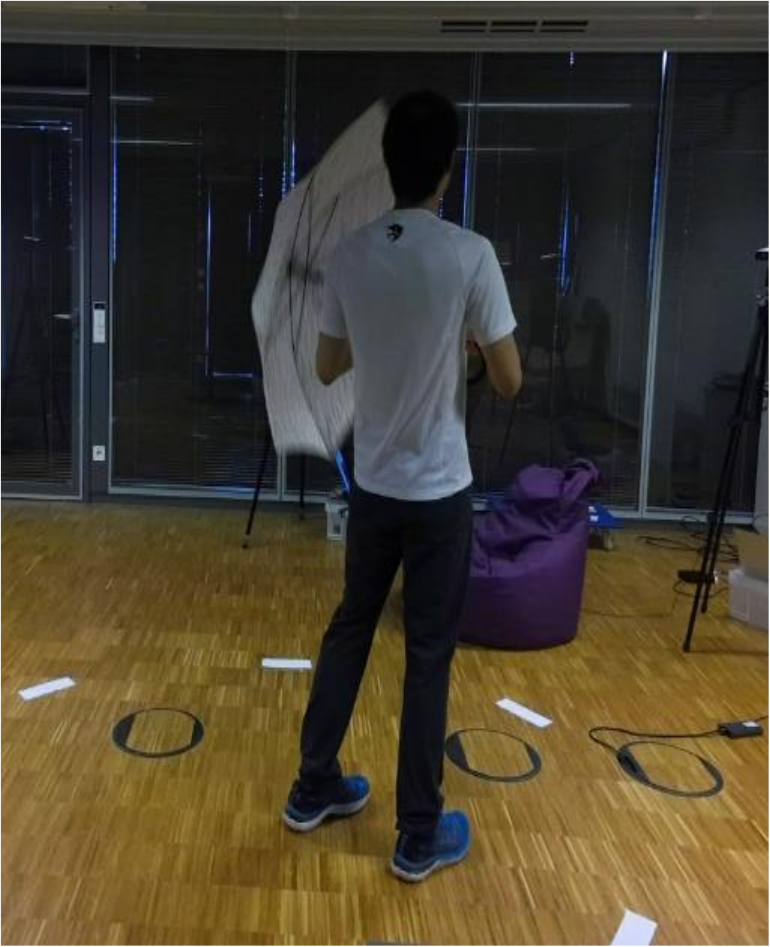}     & \includegraphics[height=0.2\textwidth]{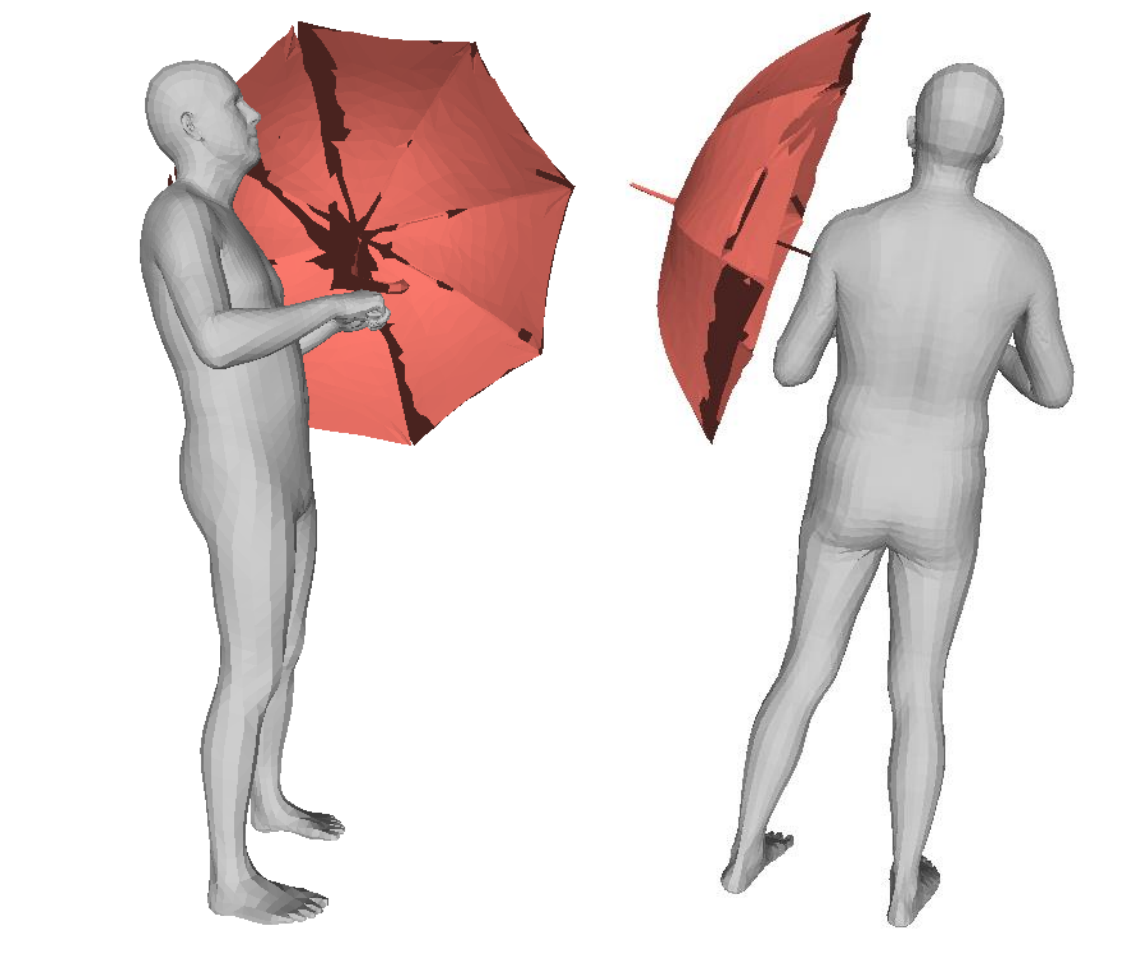} & \includegraphics[height=0.2\textwidth]{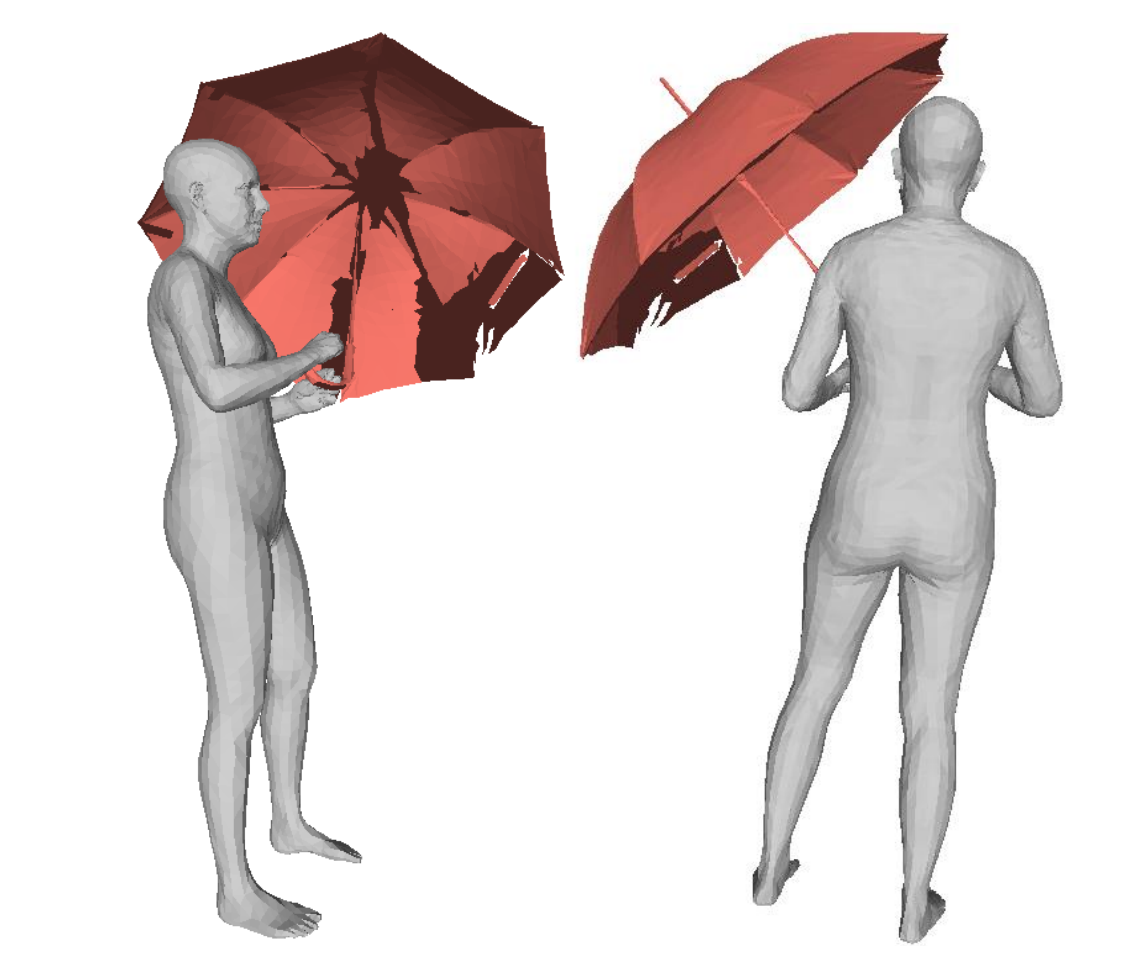} & \includegraphics[height=0.2\textwidth]{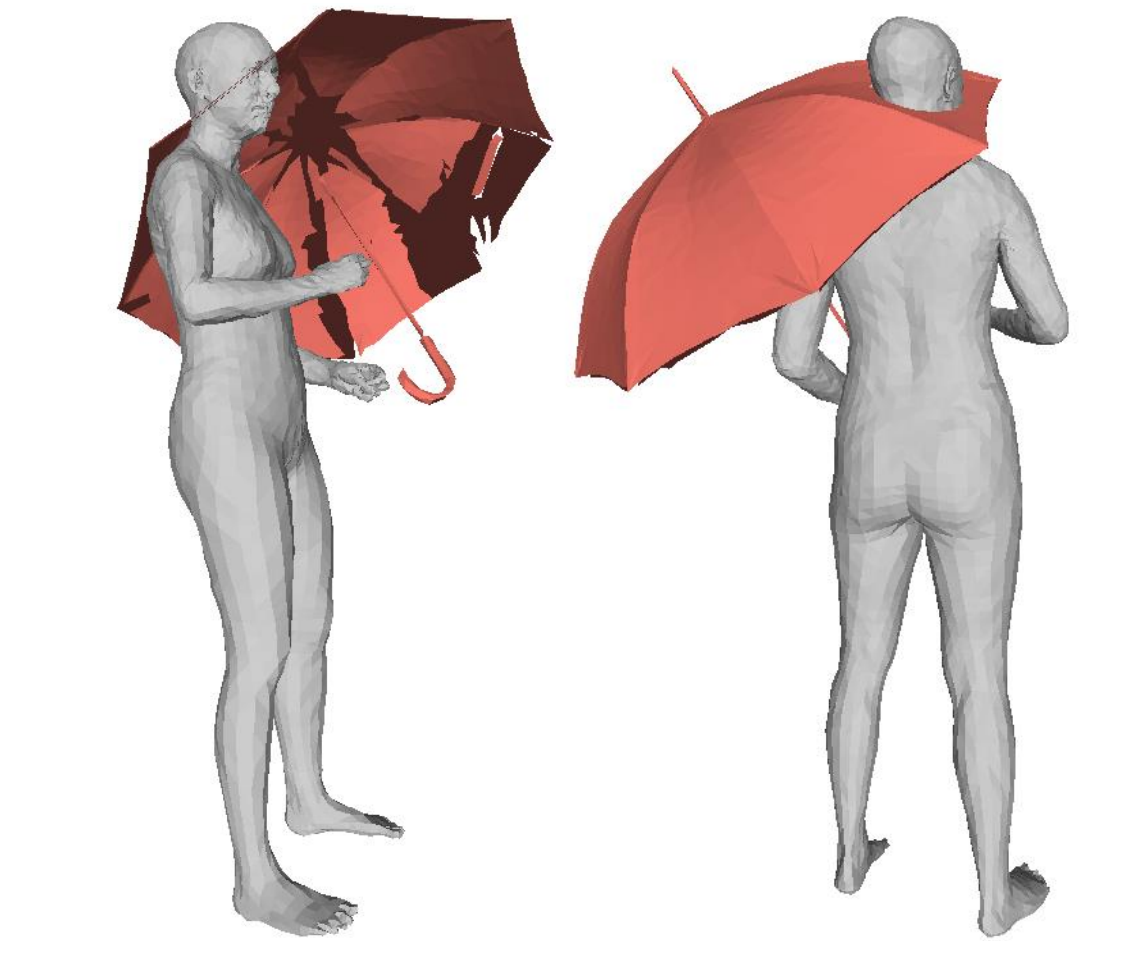} \\
     RGB &  GT & HOI-TG (Ours) & CONTHO
    \end{tabular}
    \caption{Qualitative comparison of 3D human and object reconstruction with CONTHO~\cite{joint} on InterCap~\cite{intercap} dataset.}
    \label{fig:sullpyintercap}
\end{figure*}

We provide more HOI reconstruction results in the BEHAVE~\cite{behave} and InterCap~\cite{intercap} datasets in Figs.~\ref{fig:sullpybehave} and \ref{fig:sullpyintercap}. Regarding complex interactive actions, HOI-TG surpasses CONTHO~\cite{joint} in both human mesh reconstruction and object posture estimation. Our model demonstrates an advantage in dealing with mesh penetration and inaccurate object posture. It also achieves higher reconstruction accuracy for human-object interactions without physical contact.

\setlength{\tabcolsep}{10pt}

\setlength{\tabcolsep}{5pt}

\end{document}